\documentclass{article}

\usepackage[main,final]{neurips_2026}

\usepackage[utf8]{inputenc} 
\usepackage[T1]{fontenc}    
\usepackage{hyperref}       
\usepackage{url}            
\usepackage{booktabs}       
\usepackage{amsfonts}       
\usepackage{nicefrac}       
\usepackage{microtype}      
\usepackage{xcolor}         
\usepackage{graphicx} 
\usepackage{amsfonts}
\usepackage{tablefootnote}
\usepackage{amsmath}
\usepackage{amssymb}\usepackage{tikz}\let\authsavedBox\Box\let\authsavedDiamond\Diamond\let\authsaveddiamond\diamond\let\authsavedleadsto\leadsto\let\Box\relax\let\Diamond\relax\let\diamond\relax\let\leadsto\relax\usepackage[nointegrals]{wasysym}\let\Box\authsavedBox\let\Diamond\authsavedDiamond\let\diamond\authsaveddiamond\let\leadsto\authsavedleadsto\newcommand{\authsaturn}{\text{\tikz[baseline=-0.5ex, x=1pt, y=1pt, line width=0.5pt]{\begin{scope}[rotate=-22]\draw (-4.4, 0) arc (180:360:4.4 and 1.5);\end{scope}\fill (0,0) circle (2.4);\begin{scope}[rotate=-22]\draw[white, line width=1.0pt] (-4.4, 0) arc (180:0:4.4 and 1.5);\draw[line width=0.5pt] (-4.4, 0) arc (180:0:4.4 and 1.5);\end{scope}}}}\newcommand{\authmoon}{\text{\raisebox{-0.1ex}{\scalebox{1.05}{\rotatebox[origin=c]{35}{$\leftmoon$}}}}}
\usepackage{algorithm}
\usepackage{algpseudocode}
\usepackage{colortbl}
\usepackage{array}
\usepackage{wrapfig}

\usepackage{tabularx}
\usepackage{multirow}
\usepackage{subcaption}
\usepackage{cleveref}
\usepackage{tocloft}

\definecolor{fst}{rgb}{0.810, 0.329, 0.427}
\definecolor{sec}{rgb}{0.882, 0.561, 0.616}
\definecolor{thd}{rgb}{0.961, 0.837, 0.855}

\definecolor{groundcolor}{RGB}{120, 50, 180}   
\definecolor{spacecolor}{RGB}{225, 60, 140}    
\newcommand{\colG}{\textcolor{groundcolor}{\textbf{\makebox[0.7em][c]{G}}}}
\newcommand{\colS}{\textcolor{spacecolor}{\textbf{\makebox[0.7em][c]{S}}}}

\newcommand\blfootnote[1]{%
  \begingroup
  \renewcommand\thefootnote{}\footnote{#1}%
  \addtocounter{footnote}{-1}%
  \endgroup
}

\newcommand{\fst}{\cellcolor{fst}}
\newcommand{\sed}{\cellcolor{sec}}
\newcommand{\thd}{\cellcolor{thd}}

\newcommand{\cmark}{{\color{red!80!black}\checkmark}}
\newcommand{\xmark}{{\color{black}$\times$}}

\makeatletter
\newcommand{\TOCdisable}{%
  \let\saved@addcontentsline\addcontentsline
  \renewcommand{\addcontentsline}[3]{}%
}
\newcommand{\TOCenable}{%
  \let\addcontentsline\saved@addcontentsline
}
\makeatother

\definecolor{citecolor}{HTML}{BF0040}
\definecolor{linkcolor}{HTML}{F54747}
\hypersetup{colorlinks=true,citecolor=citecolor, linkcolor=linkcolor, urlcolor=linkcolor}

\title{FluxFlow: Conservative Flow-Matching for Astronomical Image Super-Resolution}

\author{%
  \textbf{Shuhong Liu$^{1,2,\authsaturn\authmoon}$,\;
  Xining Ge$^{2,\authsaturn}$,\; Quanfeng Xu$^{3,\authsaturn}$,\; Ziteng Cui$^{1}$,\;
  Liuzhuozheng Li$^{1}$} \\
  \textbf{Gengjia Chang$^{2}$,\; Jun Liu$^{2}$,\; Ziying Gu$^{1}$,\; Dong Li$^{2}$,\; Xuangeng Chu$^{1,2}$,\; Lin Gu$^{4}$} \\
  \textbf{Tatsuya Harada$^{1,5}$} \\[4pt]
  $^{1}$The University of Tokyo \quad
  $^{2}$I2WM \quad
  $^{3}$SHAO \quad
  $^{4}$Tohoku University \quad
  $^{5}$RIKEN
}

\RequirePackage{color}\definecolor{RED}{rgb}{1,0,0}\definecolor{BLUE}{rgb}{0,0,1} 
\DeclareOldFontCommand{\sf}{\normalfont\sffamily}{\mathsf} 
\providecommand{\DIFaddtex}[1]{{#1}} 
\providecommand{\DIFdeltex}[1]{{}} 
\providecommand{\DIFaddbegin}{} 
\providecommand{\DIFaddend}{} 
\providecommand{\DIFaddFL}[1]{\DIFadd{#1}} 
\providecommand{\DIFaddbeginFL}{} 
\providecommand{\DIFaddendFL}{} 
\providecommand{\DIFadd}[1]{\texorpdfstring{\DIFaddtex{#1}}{#1}} 

\begin{document}
\DIFaddbegin \raggedbottom %

\DIFaddend

\maketitle
\blfootnote{$^{\authsaturn}$These authors contribute equally to this work. $^{\authmoon}$Corresponding author.}

\begin{abstract}
  Ground-to-space astronomical super-resolution requires recovering space-quality images from ground-based observations that are simultaneously limited by pixel sampling resolution and atmospheric seeing, which imposes a stochastic, spatially varying PSF that cannot be resolved through upsampling alone. Existing methods rely on synthetic training pairs that fail to capture real atmospheric statistics and are prone to either over-smoothed reconstructions or hallucination sources with no physical counterpart in the observed sky. We propose FluxFlow, a conservative pixel-space flow-matching framework that incorporates observation uncertainty and source-region importance weights during training, and a training-free Wiener-regularized test-time correction to suppress hallucination sources while preserving recovered detail. We further construct the DESI--HST Dataset, the large-scale real-world benchmark comprising 19,500 real co-registered ground-to-space image pairs with real atmospheric PSF variation.  \DIFadd{Experiments demonstrate that FluxFlow consistently outperforms existing baseline methods in both photometric and scientific accuracy.}  \DIFaddbegin \DIFadd{Project webpage: }\url{https://i2wm.github.io/fluxflow/}\DIFadd{.
}\DIFaddend \end{abstract}

\begin{figure}[!ht]
    \centering
    \includegraphics[width=\linewidth]{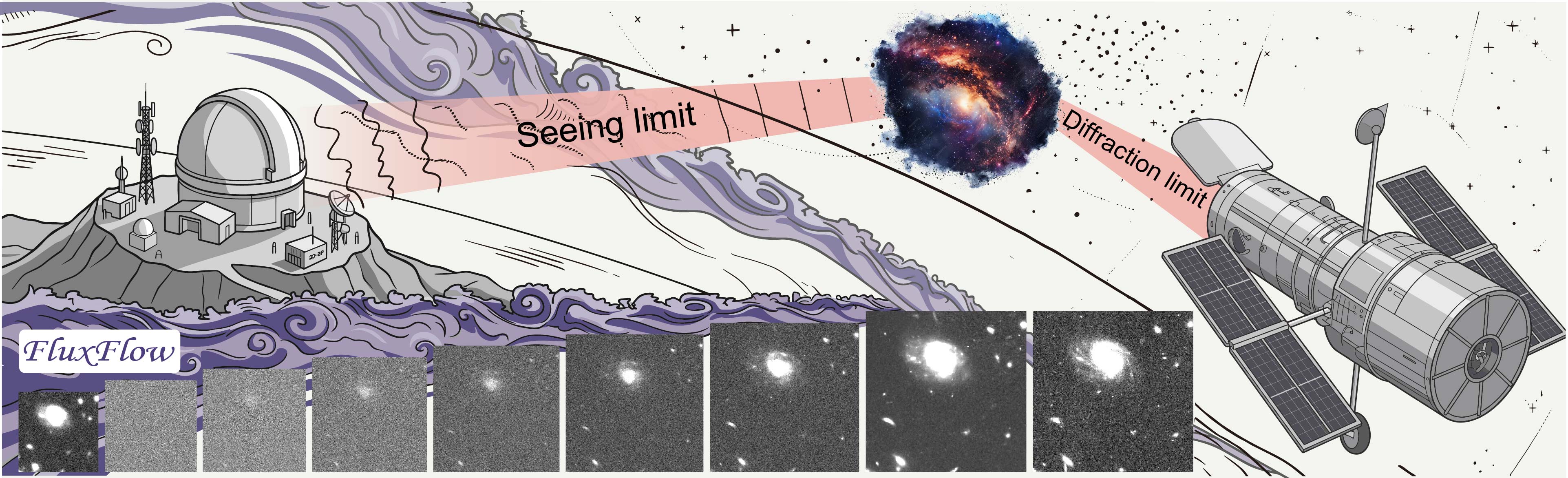}
    \caption{FluxFlow reconstructs HST-quality high-resolution images from seeing-limited low-resolution ground-based telescope observations.}
    \label{fig:teaser}
\end{figure}

\section{Introduction}

Image super-resolution (SR) has made substantial advances on natural images~\cite{liang2021swinir,zamir2022restormer,lipman2022flow,chen2025hat,ren2026esr}, where the degradation model is dominated by limited sampling density on the image plane. In astronomical imaging, however, effective resolution is jointly governed by pixel sampling and angular resolution, where the latter is determined by the optical point spread function (PSF) and further degraded by atmospheric turbulence~\cite{sirianni2005photometric,dey2019overview}. Consequently, super-resolution that merely upsamples a PSF-blurred observation cannot recover the underlying finer physical structure.

 \DIFaddbegin \begin{figure}[!t]
    \DIFaddendFL \centering
     \DIFaddbeginFL \includegraphics[width=\linewidth]{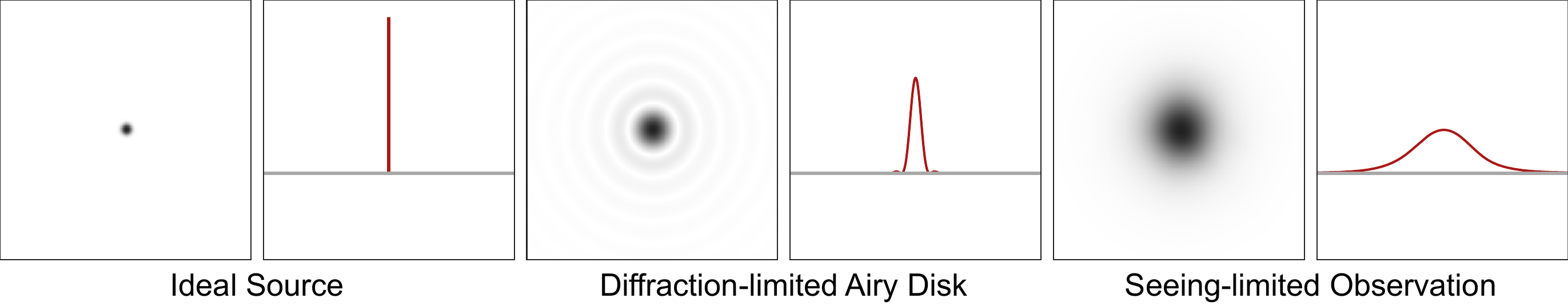}
    \DIFaddendFL \caption{Illustration of point source degradation from ideal (left) to diffraction-limited (center) to seeing-limited (right), with 2D intensity images and 1D cross-sectional profiles. }
    \label{fig:psf_example}
\end{figure}

This distinction is especially important in ground-to-space image super-resolution. As illustrated in \Cref{fig:psf_example}, Ground-based observations are typically seeing-limited, where atmospheric turbulence broadens the PSF and imposes an angular-resolution floor that often dominates over the detector sampling scale. In contrast, space-based telescopes such as the Hubble Space Telescope (HST) are free from atmospheric seeing and can approach the diffraction limit of their optics, yielding substantially sharper images. However, its small field of view and heavily oversubscribed observing time restrict such image quality to a small fraction of the sky, creating a persistent gap between wide-field ground-based surveys and high-resolution space-based observations.

Bridging this gap would enable high-resolution analysis over survey scales that are inaccessible to space telescopes alone. Wide-field surveys such as DESI~\cite{dey2019overview} image hundreds of millions of galaxies across thousands of square degrees, yet their seeing-limited image quality restricts measurements that depend on resolved structure, including weak-lensing shear estimation \cite{mandelbaum2015great3}, galaxy morphology \cite{conselice2014evolution}, and accurate PSF-aware photometry. A reliable mapping from ground-based observations to space-quality images would extend the reach of such analyses without requiring comparable space-based coverage.

Despite this potential, existing astronomical SR methods \cite{miao2024astrosr,shan2025galaxy, wu2025star, zhang2026bridge} are typically trained on synthetically degraded image pairs that do not fully reproduce the statistics of real observations, including atmospheric PSF variation, instrument-specific noise~\cite{liu2026denoising}, and spatially
correlated artifacts introduced by frame co-addition \cite{bertin2010swarp,gonzaga2012drizzlepac}. This mismatch leads to a substantial domain gap when models are applied to genuine cross-instrument data. Regression-based methods further tend
to produce overly smooth reconstructions that suppress fine morphological structure, while generative models, although better at recovering perceptual detail \cite{saharia2022image, wang2024sinsr}, can introduce compact hallucination sources that can directly bias downstream scientific products, including source catalogs \cite{bertin1996sextractor}, flux measurements \cite{conselice2014evolution}, photometric redshifts \cite{salvato2019many}, and lensing analyses \cite{bernstein2002shapes}.

To address these challenges, we introduce FluxFlow, a flow-matching framework for ground-to-space astronomical super-resolution. Our method improves physical fidelity at both training and inference time. During training, it uses spatial reliability and source-importance maps to focus learning on  \DIFadd{scientifically focused regions}  while suppressing unstable gradients from artifacts. During inference, it enforces measurement consistency with the observed low-resolution image by back-projecting residuals through a Wiener-regularized inverse of the forward operator. Together, these designs suppress the generation of hallucination sources while preserving the fine structure. To overcome the scarcity of real observational data for ground-to-space SR, we further construct the DESI--HST Dataset, the large-scale astronomical SR dataset comprising real-world co-registered ground-to-space image pairs with authentic atmospheric PSF variation and observation uncertainty maps.

Our contributions are summarized as follows:
\begin{itemize}
    \item We propose FluxFlow, a conservative pixel-space flow-matching framework with a physically informed training objective and a training-free Wiener-regularized inference-time correction that suppresses hallucination sources while preserving restored detail.
    \item We construct the DESI--HST Dataset, a real paired ground-to-space super-resolution benchmark containing 19,500 paired images at $\times 2$ and $\times 4$ scales, with real atmospheric PSF variation and per-pixel uncertainty weight maps.
    \item \DIFadd{Experiments on our curated dataset demonstrate that FluxFlow achieves superior photometric fidelity and scientific accuracy over both recent regression-based and generative approaches.}
\end{itemize}

\section{Related Work}

\subsection{Regression-Based Image Restoration}
Deep learning has driven substantial progress in single-image super-resolution and image restoration. CNN backbones such as MPRNet~\cite{zamir2021multi} and NAFNet~\cite{chen2022simple}, Transformer variants including SwinIR~\cite{liang2021swinir}, Restormer~\cite{zamir2022restormer}, HAT~\cite{chen2023activating,chen2025hat}, SRFormer~\cite{zhou2023srformer}, DAT~\cite{chen2023dual}, and HiT-SR~\cite{zhang2024hit}, and state-space models such as MambaIR~\cite{guo2024mambair} and MaIR~\cite{li2025mair} have steadily raised restoration quality through larger receptive fields and richer feature interaction. A parallel line pursues all-in-one and condition-aware restoration via degradation-adaptive~\cite{liang2022efficient}, prompt-based~\cite{potlapalli2023promptir,wang2023promptrestorer,conde2024instructir}, and scale-adaptive~\cite{he2025universal} designs, alongside efficiency-oriented work on quantization~\cite{qin2023quantsr}, Fourier shifting~\cite{zhou2024improving}, and computational rebalancing~\cite{yu2024rethinking}. In astronomy, supervised pipelines address noise calibration~\cite{liu2026denoising}, X-ray data~\cite{sweere2022deep}, galaxies~\cite{miao2024astrosr}, star fields~\cite{wu2025star}, and cross-survey enhancement~\cite{luo2025cross,li2022self}. All these methods yield a deterministic output from the degraded input and cannot capture the ambiguity of severely ill-posed one-to-many inverse problems.

\subsection{Diffusion Models for Conditional Generation}
Diffusion models~\cite{song2020score} serve as generative priors for restoration, with classifier~\cite{dhariwal2021diffusion}, classifier-free~\cite{ho2022classifier}, latent~\cite{rombach2022high}, and cascaded~\cite{ho2022cascaded} variants enabling controllable synthesis~\cite{zheng2023guided}. Restoration is cast as iterative refinement~\cite{saharia2022image,saharia2022palette}, posterior sampling with measurement operators~\cite{chung2022diffusion,song2023pseudoinverse}, or spatially conditioned generation~\cite{zhang2023adding}, and is accelerated or constrained through residual shifting~\cite{yue2023resshift}, single-step generation~\cite{wang2024sinsr,wu2024one,li2025one,dong2025tsd}, consistency distillation~\cite{he2024consistency,gong2024ir}, diffusion inversion~\cite{yue2025arbitrary}, faithfulness constraints~\cite{matsunaga2022fine,zhang2025uncertainty,chen2025faithdiff,polowczyk2026diamond}, and  \DIFaddbegin \DIFadd{Schr}{\DIFadd{\"o}}\DIFadd{dinger }\DIFaddend or Brownian bridge transport~\cite{liu20232,li2023bbdm,zhou2023denoising}. In astronomy, diffusion-based SR~\cite{ruan2025investigation,guo2026computational} targets gravitational lensing~\cite{reddy2024difflense}, solar imaging~\cite{song2024improving}, galaxy reconstruction~\cite{shan2025galaxy}, intensity interferometry~\cite{rai2025generative}, and cross-survey generation~\cite{zhang2026bridge}.

\section{DESI-HST Cross-Survey Dataset}
\label{sec:dataset}

\paragraph{Instruments and Data Collection.}

\begin{figure}[t]
    \centering
    \includegraphics[width=\linewidth]{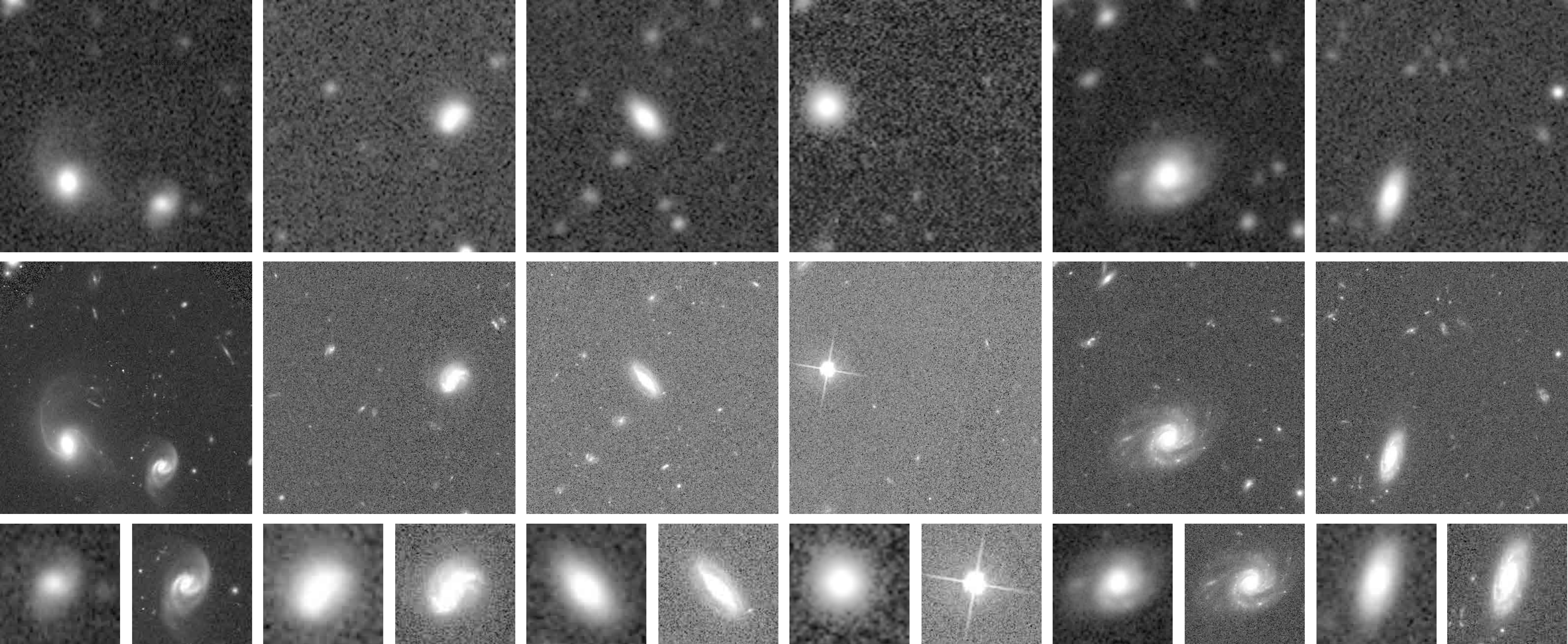}
    \caption{Paired examples from our DESI--HST dataset. Top: DESI cutouts at $128\times128$ pixels; Bottom: HST observations at $512\times512$ pixels.  \DIFaddbeginFL \DIFaddFL{The image }\DIFaddendFL pairs are  \DIFaddbeginFL \DIFaddFL{spatially aligned using World Coordinate System metadata}\DIFaddendFL . The DESI images are visibly blurred due to coarser angular resolution and atmospheric turbulence.}
    \label{fig:dataset}
\end{figure}

Unlike previous astronomical SR datasets that either synthesize LR images via degradation kernels~\cite{wu2025star} or pair two ground-based surveys~\cite{miao2024astrosr, luo2025cross}, our dataset comprises  \DIFadd{genuinely observed}  ground-to-space image pairs from two independent instruments  whose resolution gap is dominated by atmospheric turbulence. The LR images are drawn from the DESI Legacy Imaging Surveys~\cite{dey2019overview}, a ground-based program providing $i$-band imaging (${\sim}7600$--$8730$\,\AA) at $0.262''$/pixel \DIFaddbegin \footnote{\DIFadd{The symbols $\prime$ and $\prime\prime$ denote arcminutes and arcseconds, respectively. One degree equals 60 arcminutes or 3600 arcseconds.}}\DIFaddend which offers the closest spectral match to the HR targets. The HR images are obtained from HST ACS/WFC in the F814W filter (${\sim}7000$--$9500$\,\AA), delivering diffraction-limited imaging at $0.03''$/pixel after drizzle stacking. \DIFaddbegin \footnote{\DIFadd{For details of drizzle stacking, see~\citet{fruchter2002drizzle} and \citet{gonzaga2012drizzlepac}. We select the DESI $i$ band and HST F814W because they are the closest available bandpasses between the two observing programs.}
}\DIFaddend 
Details are provided in Appendix~\ref{sec:appendixa}.

Our dataset is collected from three extragalactic deep fields with overlapping HST and DESI coverage: COSMOS, GOODS-S, and UDS. Image pairs are spatially aligned using World Coordinate System metadata, with DESI coadds retained at native resolution to preserve their noise properties and PSF. The corresponding HST images are reprojected and flux-conservingly downsampled to produce aligned HR targets at $\times2$ and $\times4$ scale factors (HR at $256\times256$ and $512\times512$). The resulting LR--HR pairs are cropped into patches and split into 17{,}800 training and 1{,}700 test pairs per scale. Visualizations are shown in~\Cref{fig:dataset} and alignment procedures are detailed in Appendix~\ref{sec:appendixb}.

\paragraph{Atmospheric Turbulence.}

\begin{wraptable}{r}{0.5\textwidth}\vspace{-0.5\baselineskip}
\centering
\small
\refstepcounter{footnote}\caption{\DIFadd{Comparison of astronomical }SR datasets. \colG{} \DIFadd{and} \colS{} \DIFadd{denote ground and space observations.}\protect\hyperlink{tabone-columns}{\protect\textsuperscript{\thefootnote}}}
\setlength{\tabcolsep}{3pt}
\label{tab:astro_sr_datasets}
\begin{tabular}{lrccc}
\toprule
Dataset & Size & Real & Flux & SR-Type \\
\midrule
\cite{miao2024astrosr} & 2,000 & \cmark & \xmark & \colG$\rightarrow$\colG \\
\cite{shan2025galaxy} & 9,383 & \cmark & \xmark & \colG$\rightarrow$\colG \\
\cite{song2024improving} & 1,597 & \cmark & \xmark & \colS$\rightarrow$\colG \\
\cite{reddy2024difflense}\footnotemark & 2,880 & \cmark & \xmark & \colG$\rightarrow$\colS \\
\cite{luo2025cross} & 17,000 & \cmark & \xmark & \colG$\rightarrow$\colG \\
\cite{li2022self} & 14,604 & \cmark & \xmark & \colG$\rightarrow$\colG \\
\cite{wu2025star} & 54,738 & \cmark & \cmark & \colS$\rightarrow$\colS \\
\cite{zhang2026bridge} & 120,000 & \xmark & \cmark & \colS$\rightarrow$\colG \\
\textbf{Ours} & 19,500 & \cmark & \cmark & \colG$\rightarrow$\colS \\
\bottomrule
\end{tabular}
\end{wraptable}
 \footnotetext[\numexpr\value{footnote}-1\relax]{\hypertarget{tabone-columns}{}\DIFadd{Size counts image pairs. Real denotes observed pairs. Flux indicates whether linear photometric flux is preserved.}} \footnotetext{Difflense~\cite{reddy2024difflense}'s dataset was presented in the sRGB domain and not publicly available.}

Beyond pixel scale, the resolution gap between DESI and HST is dominated by atmospheric turbulence. Ground-based DESI observations are seeing-limited, with stochastic refractive-index fluctuations imposing a time-varying, spatially extended PSF of roughly $1.0''$--$1.3''$ FWHM in the DESI Imaging Surveys~\cite{dey2019overview}. HST operates above the atmosphere and is diffraction-limited, reaching a PSF FWHM of ${\sim}0.09''$~\cite{sirianni2005photometric}. The resulting $11$--$14\times$ ratio makes DESI-to-HST super-resolution substantially harder than previous datasets relying on pixel-scale downsampling alone.

\section{Preliminaries of Flow-Matching}
\label{sec:flow_matching}

Flow Matching~\cite{lipman2022flow} learns a continuous-time normalizing flow that transports samples from a source distribution $p_0$ to a target distribution $p_1$ via an ordinary differential equation (ODE):
\begin{equation}
    \frac{\mathrm{d}\mathbf{x}_t}{\mathrm{d}t} = v_\theta(\mathbf{x}_t, t), \quad t \in [0, 1],
    \label{eq:flow_ode}
\end{equation}
where $v_\theta$ is a time-dependent velocity field parameterized by a neural network. Given an initial sample $\mathbf{x}_0 \sim p_0$, integrating \Cref{eq:flow_ode} from $t{=}0$ to $t{=}1$ yields a sample $\mathbf{x}_1 \sim p_1$.

The central idea of Flow Matching is to define a conditional probability path $p_t(\mathbf{x} \mid \mathbf{x}_0,\mathbf{x}_1)$ that interpolates between the source and target distributions, together with its generating conditional velocity field $u_t(\mathbf{x} \mid \mathbf{x}_0,\mathbf{x}_1)$. We adopt the optimal-transport (OT) conditional path:
\begin{equation}
    p_t(\mathbf{x} \mid \mathbf{x}_0,\mathbf{x}_1) = \mathcal{N}\bigl(\mathbf{x} \;\big|\; t\,\mathbf{x}_1 + (1 - t)\,\mathbf{x}_0,\; \sigma_{\min}^2 \mathbf{I}\bigr),
    \label{eq:cond_path}
\end{equation}
where $\sigma_{\min}$ is a small constant. The conditional velocity field that generates this path takes the form:
\begin{equation}
    u_t(\mathbf{x} \mid \mathbf{x}_0,\mathbf{x}_1) = \mathbf{x}_1 - \mathbf{x}_0.
    \label{eq:cond_velocity}
\end{equation}

The network $v_\theta$ is trained by minimizing the Conditional Flow Matching (CFM) objective~\cite{lipman2022flow, tong2023improving}:
\begin{equation}
    \mathcal{L}_{\mathrm{CFM}} = \mathbb{E}_{t \sim \mathcal{U}[0,1],\, \mathbf{x}_0 \sim p_0,\, \mathbf{x}_1 \sim p_1} \left\| v_\theta(\mathbf{x}_t, t) - u_t(\mathbf{x}_t \mid \mathbf{x}_0,\mathbf{x}_1) \right\|^2,
    \label{eq:cfm_loss}
\end{equation}
where $\mathbf{x}_t = t\,\mathbf{x}_1 + (1 - t)\,\mathbf{x}_0$ is the linearly interpolated sample. The CFM loss regresses against the conditional velocity $u_t$ rather than the intractable marginal velocity field, yet provably yields the same gradient in expectation~\cite{lipman2022flow}.

At inference, samples are generated by solving \Cref{eq:flow_ode} with a numerical ODE solver, i.e., Euler or Heun \cite{karras2022elucidating}, using $N$ discretization steps from $t{=}0$ to $t{=}1$.

\section{FluxFlow}

\paragraph{Pixel-Domain Flow Matching.}
Modern generative models for natural images typically operate in the latent space of a VAE~\cite{rombach2022high}, which is ill-suited to scientific imaging. Astronomical frames are single-band and span a high dynamic range, whereas standard VAEs are trained on three-channel 8 bit photographs and cannot be transferred off-the-shelf. Retraining a tokenizer on astronomical data is also unappealing, since the compression itself degrades high-frequency structure of the targets and corrupts flux measurements. Motivated by recent studies on pixel-space diffusion that achieve high-resolution generation without a latent tokenizer~\cite{hoogeboom2023simple,hoogeboom2025simpler,li2025back}, we learn the velocity field $v_\theta$ directly on pixel inputs.

\subsection{Physically-Informed Velocity Learning}
\label{sec:conservation_losses}

While the conditional flow-matching objective in \Cref{eq:cfm_loss} encourages perceptual realism, it does not guarantee that the generated images are physically consistent. Two quantities drive most downstream analysis. The integrated flux of a source encodes its luminosity and underpins stellar-mass and star-formation-rate estimates~\cite{conselice2014evolution}, and the PSF shape determines apparent size for weak lensing and morphological classification~\cite{bernstein2002shapes,mandelbaum2015great3}. We enforce both through spatially varying weights in the velocity objective that separately encode data reliability and scientific importance.

\paragraph{Inverse-Variance Weighting.}
\label{eq:wht_compress}

\begin{wrapfigure}[15]{r}{0.36\linewidth}\vspace{-0.5\baselineskip}
    \centering
    \includegraphics[width=\linewidth]{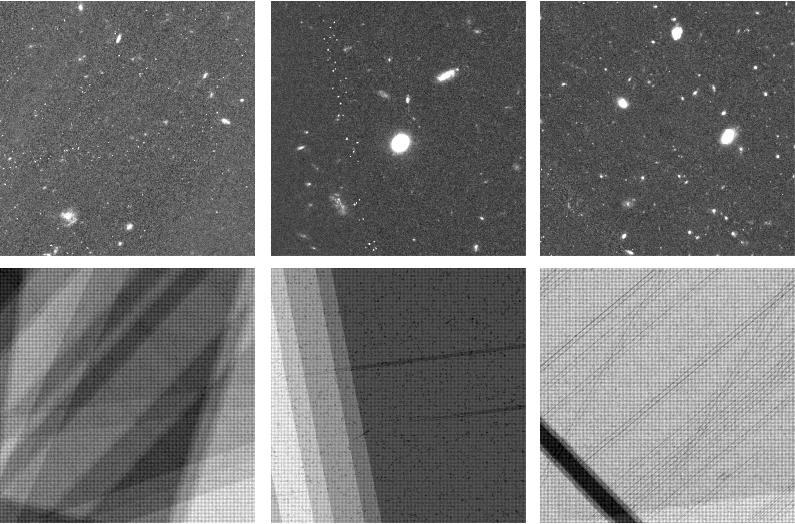}
    \caption{Visualization of HST science images (top) and WHT weight maps (bottom), showing structures from the drizzle stacking and local artifacts such as cosmic rays.}
    \label{fig:wht}
\end{wrapfigure}

The drizzled HST mosaic is accompanied by a per-pixel weight map (WHT) $\mathbf{W} \in \mathbb{R}^{H_{\mathrm{hr}} \times W_{\mathrm{hr}}}$ encoding the inverse variance at each pixel. It reflects spatially varying exposure depth from the dither pattern together with statistical noise and residual instrumental artifacts such as chip gaps and cosmic-ray trails, as shown in \Cref{fig:wht}. Pixels affected by these signatures receive lower weight. We compress the dynamic range with a square-root transform $\widetilde{\mathbf{W}} = \sqrt{\mathbf{W}}$ to suppress dominance by a few pixels.

\paragraph{Source-Region Weighting.}

The inverse-variance map addresses data quality. Orthogonally, we introduce a scientific importance weight that flags which pixels matter most for analysis. Given a binary source mask $\mathbf{M}$ obtained from SExtractor~\cite{bertin1996sextractor} run on the HST \DIFadd{images}, we construct $\mathbf{S} = 1 + \mathbf{M}$, which doubles the loss weight on source pixels relative to the background. We weight all pixels within a source uniformly because morphological and PSF-shape information resides primarily in the faint outer profile, which an intensity-dependent scheme would suppress.

\paragraph{Overall Objective.}

Multiplying the two maps yields a single spatially varying weight that encodes both reliability and importance, and is applied as a per-image normalized objective:
\begin{equation}
    \mathcal{L}_{\mathrm{WFM}} \;=\; \mathbb{E}_{t,\,\mathbf{x}_0,\,\mathbf{x}_1}\!\left[\,
        \frac{\sum_{p}\bigl(\widetilde{\mathbf{W}}\odot\mathbf{S}\bigr)_{p}\,\bigl\|v_\theta(\mathbf{x}_t,t,\mathbf{c}) - u_t\bigr\|_{p}^{2}}{\sum_{p}\bigl(\widetilde{\mathbf{W}}\odot\mathbf{S}\bigr)_{p}}
    \,\right],
    \label{eq:wfm_loss}
\end{equation}
where $\mathbf{c}$ denotes the conditioning upsampled DESI observation specified in Appendix~\ref{sup:model}, $\odot$ is element-wise multiplication, $p$ indexes spatial pixel locations, and the denominator $\sum_{p}(\widetilde{\mathbf{W}}\odot\mathbf{S})_p$ is recomputed per training pair so that each pair contributes a gradient signal of comparable magnitude irrespective of the absolute scale of $\widetilde{\mathbf{W}}\odot\mathbf{S}$. In OT-CFM the velocity target $u_t = \mathbf{x}_1 - \mathbf{x}_0$ is independent of $t$ and the interpolant $\mathbf{x}_t = (1-t)\,\mathbf{x}_0 + t\,\mathbf{x}_1$ is a pixelwise convex combination, so the mask $\mathbf{M}$ computed on $\mathbf{x}_1$ remains spatially aligned with source content at every $t$.


\subsection{Measurement-Consistent Flow Sampling}
\label{sec:mcfm}

The physics-informed weights improve velocity-field fidelity on source regions but leave the background unregulated. Unlike regression-based models~\cite{dong2015image,lim2017enhanced} that converge to the conditional mean, generative samplers can produce hallucinated sources with no counterpart in $\mathbf{y}$. We analyze their formation and propose a training-free, inference-time correction that enforces consistency with the observation.

\paragraph{Hallucination Sources.}

\begin{figure}[t]
    \centering
    \includegraphics[width=\linewidth]{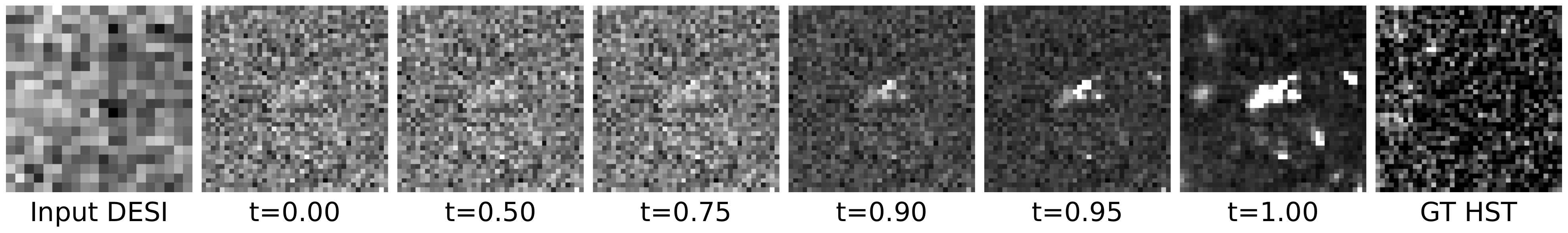}
    \caption{Sampling trajectory of a representative hallucination source across ODE integration steps. The spurious structure is already latent in the initial noise and progressively sharpened by the flow into a compact source-like morphology.}
    \label{fig:hall_flow_traj}
\end{figure}

\begin{wrapfigure}[18]{r}{0.42\linewidth}
    \centering
    \includegraphics[width=\linewidth]{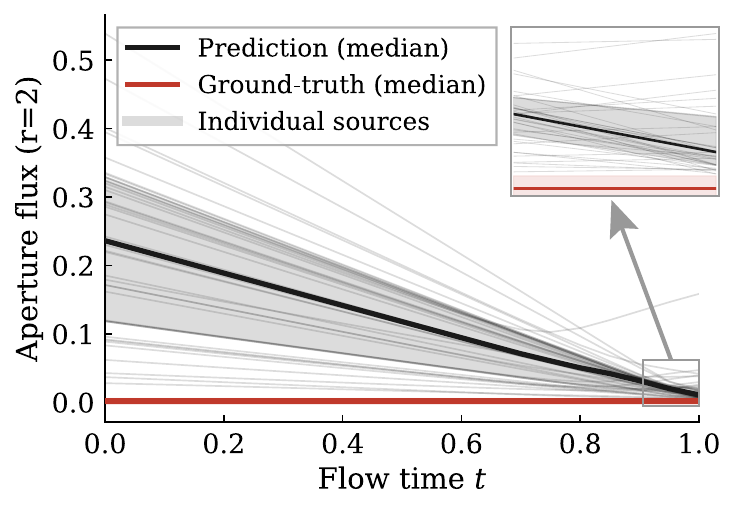}
    \caption{Aperture flux (sum) evolution of bright hallucinated sources along the ODE sampling trajectory. It starts from large-variance noise and converges to brightness levels exceeding the HST (red) at $t{=}1$.}
    \label{fig:hall_pixel_traj}
\end{wrapfigure}

We track hallucinations along the full ODE trajectory. As shown in \Cref{fig:hall_flow_traj}, a hallucination is already present in the initial noise as a local intensity fluctuation, which the velocity field sharpens at late steps into a compact source-like morphology with surrounding ring artifacts. \Cref{fig:hall_pixel_traj} plots the mean brightness within a 2-pixel radius of 50 hallucination centroids as a function of $t$, nearly converging at $t{=}1$ that consistently exceed the HST . This behavior resists standard remedies such as classifier-free guidance~\cite{ho2022classifier} and dense guidance~\cite{zheng2023layoutdiffusion,zhang2023adding}. The low SNR of ground-based observations makes hallucinations nearly indistinguishable from faint real sources in intermediate states $\mathbf{x}_t$\footnote{We trained a U-Net to predict source masks from $\mathbf{x}_t$ on the DESI-HST dataset, obtaining an mIoU of only 11.3\%.}, and once a hallucination forms, the velocity field self-reinforces it at every subsequent step.

\begin{algorithm}[t]
\DIFaddend \caption{MC-FS: Measurement-Consistent Flow Sampling}
\label{alg:mcfm}
\small
\begin{algorithmic}[1]
\Require $v_\theta$, $\mathbf{y}$, $h$, $N$, $\eta_0$, SNR
\State $\mathbf{x}^{(0)} \sim \mathcal{N}(\mathbf{0}, \mathbf{I})$
\State Precompute $W(\mathbf{k})$ via \Cref{eq:wiener_kernel}
\For{$i = 0, \ldots, N{-}1$}
    \State $t_i \gets i/N$,\ \ $\Delta t \gets 1/N$
    \State $\tilde{\mathbf{x}}^{(i+1)} \gets \mathbf{x}^{(i)} + v_\theta(\mathbf{x}^{(i)}, t_i, \mathbf{c})\,\Delta t$
    \State $\mathbf{r}^{(i)} \gets \mathbf{A}(\tilde{\mathbf{x}}^{(i+1)}) - \mathbf{y}$
    \State $\eta_i \gets \eta_0(1 - i/N)$
    \State $\mathbf{x}^{(i+1)} \leftarrow \tilde{\mathbf{x}}^{(i+1)} - \eta_i\,\mathcal{F}^{-1}\!\bigl(W \cdot \mathcal{F}[\mathbf{D}_s^{\!\top}\mathbf{r}^{(i)}]\bigr)$
\EndFor
\State \Return $\mathbf{x}^{(N)}$
\end{algorithmic}
\end{algorithm}\paragraph{Physical Forward-Model Consistency.}
\label{sec:mcfs}
Inspired by diffusion posterior sampling for inverse problems~\cite{chung2022diffusion,song2023pseudoinverse}, we exploit the known physical forward model of the imaging system to enforce measurement consistency at every step of the flow-matching ODE. The DESI observation $\mathbf{y}$ is related to the latent high-resolution image $\mathbf{x}$ by a degradation operator $\mathbf{A}$ consisting of convolution with the point spread function followed by spatial downsampling:
\begin{equation}
    \mathbf{y} = \mathbf{A}(\mathbf{x}) + \boldsymbol{n}, \qquad \mathbf{A} = \mathbf{D}_s \circ \mathbf{H}_{\mathrm{PSF}},
    \label{eq:forward_model}
\end{equation}
where $\mathbf{H}_{\mathrm{PSF}}$ denotes convolution with a constant Gaussian PSF kernel $h$ of width $\sigma_{\mathrm{PSF}}$, $\mathbf{D}_s$ is $s{\times}$ area downsampling, and $\boldsymbol{n}$ is observation noise. Re-projecting the prediction $\hat{\mathbf{x}}$ through $\mathbf{A}$ therefore  \DIFaddbegin \DIFadd{helps suppress }\DIFaddend hallucinated sources  \DIFaddbegin \DIFadd{by reducing }\DIFaddend residuals against $\mathbf{y}$ . Concretely, at a discrete integration step $i$ with current state $\mathbf{x}^{(i)}$, we first compute a candidate update $\tilde{\mathbf{x}}^{(i+1)}$ using the velocity field:
\begin{equation}
    \tilde{\mathbf{x}}^{(i+1)} = \mathbf{x}^{(i)} + v_\theta\!\left(\mathbf{x}^{(i)}, t_i, \mathbf{c}\right) \Delta t,
    \label{eq:euler_candidate}
\end{equation}
and then measure the discrepancy between the candidate and the observation in the low-resolution pixel space:
\begin{equation}
    \mathbf{r}^{(i)} = \mathbf{A}\!\left(\tilde{\mathbf{x}}^{(i+1)}\right) - \mathbf{y}.
    \label{eq:lr_residual}
\end{equation}
This residual is then projected back to the high-resolution image via the adjoint operator $\mathbf{A}^{\!\top}$ and used to produce the corrected state $\mathbf{x}^{(i+1)}$ as:
\begin{equation}
    \mathbf{x}^{(i+1)} = \tilde{\mathbf{x}}^{(i+1)} - \eta_i\, \mathbf{A}^{\!\top}\!\left(\mathbf{r}^{(i)}\right),
    \label{eq:adjoint_correction}
\end{equation}
where $\eta_i$ is a step-size schedule. Here, $\mathbf{D}_s^{\!\top}(\mathbf{y})[p,q] = \tfrac{1}{s^2}\,\mathbf{y}[\lfloor p/s\rfloor, \lfloor q/s\rfloor]$ is the strict inner-product adjoint of $s\!\times\!s$ area downsampling, and $\mathbf{H}_{\mathrm{PSF}}^{\!\top} = \mathbf{H}_{\mathrm{PSF}}$ because the Gaussian PSF is centro-symmetric. Both convolutions are realized as circular convolutions via FFT. However, naively applying \Cref{eq:adjoint_correction} in ODE sampling would accumulate PSF-scale smoothing across integration steps. This is because the adjoint $\mathbf{A}^{\!\top} = \mathbf{H}_{\mathrm{PSF}}^{\!\top} \circ \mathbf{D}_s^{\!\top}$ contains the same PSF convolution as the forward operator $\mathbf{A}$, so every correction step convolves the residual with the PSF kernel before adding it back to $\mathbf{x}^{(i+1)}$.  \DIFaddbegin \DIFadd{With a finite number of sampling }\DIFaddend steps, \DIFaddbegin \DIFadd{slow recovery of high-frequency components can broaden }\DIFaddend the  effective PSF of  reconstructed sources. An example toy model is provided in Appendix~\ref{sup:mcfs}.

\paragraph{Wiener-Deconvolved Correction.}
To preserve the resolution of the high-resolution estimate while still enforcing measurement consistency, we instead back-project the residual through a  \DIFaddbegin \DIFadd{Wiener correction for }\DIFaddend the rank-deficient $\mathbf{A}$, replacing the PSF adjoint in $\mathbf{A}^{\top}$ with a frequency-dependent reweighting. The correction signal is computed in the frequency domain using the 2D DFT $\mathcal{F}$ on the HR grid with circular boundary conditions, consistent with the FFT-based realization of $\mathbf{H}_{\mathrm{PSF}}$ above, and $\mathbf{k}$ indexes its discrete frequencies. Given the PSF transfer function $H(\mathbf{k}) = \mathcal{F}[h]$, we define the Wiener correction kernel:
\begin{equation}
    W(\mathbf{k}) = \frac{H^{*}(\mathbf{k})}{|H(\mathbf{k})|^2 + \lambda_{\mathrm{SNR}}^{-1}},
    \label{eq:wiener_kernel}
\end{equation}
where $H^{*}$ is the complex conjugate and $\lambda_{\mathrm{SNR}}$ is a regularization parameter controlling the aggressiveness of the deconvolution. When $\lambda_{\mathrm{SNR}} \to \infty$ the filter approaches the exact inverse $H^{-1}$; when $\lambda_{\mathrm{SNR}} \to 0$ it vanishes in magnitude but stays proportional to the matched filter $H^{*}(\mathbf{k})$ of \Cref{eq:adjoint_correction}. The Wiener-corrected update then takes the form:
\begin{equation}
    \mathbf{x}^{(i+1)} = \tilde{\mathbf{x}}^{(i+1)} - \eta_i\, \mathcal{F}^{-1}\!\left(W(\mathbf{k})\cdot\mathcal{F}\!\left[\mathbf{D}_s^{\!\top}\mathbf{r}^{(i)}\right]\right),
    \label{eq:wiener_correction}
\end{equation}

\DIFaddbegin

 \DIFaddbegin

\DIFaddend where $W$ is defined on the HR grid and acts on the HR-lifted residual $\mathbf{D}_s^{\!\top}\mathbf{r}^{(i)}$ through the same transfer function $H(\mathbf{k})$. With these strict adjoints, \Cref{eq:adjoint_correction} becomes the exact gradient step on $\tfrac{1}{2}\|\mathbf{A}(\mathbf{x})-\mathbf{y}\|^2$, and \Cref{eq:wiener_correction} replaces $\mathbf{H}_{\mathrm{PSF}}^{\!\top}$ in this step with the Wiener kernel $W$. This \DIFaddbegin \DIFadd{helps reduce }\DIFaddend PSF \DIFaddbegin \DIFadd{broadening with a finite number of sampling steps while suppressing hallucinated sources}\DIFaddend . Structures in $\tilde{\mathbf{x}}^{(i+1)}$ that are inconsistent with $\mathbf{y}$ still produce a non-zero residual in \Cref{eq:lr_residual} and are therefore penalized. The step size $\eta_i$ follows a linear decay schedule $\eta_i = \eta_0 (1 - i/N)$, applying stronger corrections at early steps when the trajectory is far from convergence and relaxing the constraint at late steps to allow the flow to refine fine details. The full inference procedure is summarized in \Cref{alg:mcfm}.


\section{Experiment}
\label{sec:experiment}

\paragraph{Implementation Details.} We train our OT-CFM using the UNet backbone outlined in Appendix~\ref{sup:model}. Optimization uses AdamW (lr = 1e-4, weight decay = 1e-4) for 300 epochs at batch size 32, with linear warm-up and cosine decay, and an EMA weights decay of 0.9999. In inference, we sample 10 Euler steps with MC-FS correction. Each $\hat{x}$ is forward-projected through a Gaussian PSF ($\sigma_{\mathrm{PSF}}=2$ for both $\times2$ and $\times4$) and area-downsampling, and the residual against the DESI observation is Wiener-deconvolved ($\lambda_{\mathrm{SNR}}=50$) and subtracted with $\eta_0=0.5$. The detailed model architecture of our flow model is provided in Appendix~\ref{sup:model}. All methods are trained on a single A100-80G GPU.

\paragraph{Baseline Methods.}
We compare our method against conventional Bicubic upsampling, recent regression-based approaches, including SwinIR \cite{liang2021swinir}, HAT \cite{chen2025hat}, and FISR \cite{wu2025star}, as well as generative approaches, including cGAN \cite{rai2025generative}, GD-Net \cite{shan2025galaxy}, and AS-Bridge \cite{zhang2026bridge}. These baselines span interpolation, direct regression, and generative reconstruction, allowing us to compare different approaches to recovering high-resolution source structure.

\paragraph{Evaluation Metric.}  We adopt PSNR and SSIM to evaluate photometric fidelity. To assess scientific accuracy, we extract source ellipsoid masks from the HST GT using SExtractor and compute the averaged Flux-L1 error, which measures the absolute flux deviation within source regions. PSNR and SSIM summarize image-level reconstruction quality, while Flux-L1 focuses on photometric errors within the detected sources. Additional experiments on AstroSR~\cite{miao2024astrosr} and STAR~\cite{wu2025star} are provided in Appendix~\ref{sec:sup:experiments}.

\begin{figure}[!tp]
    \centering
    \includegraphics[width=\linewidth]{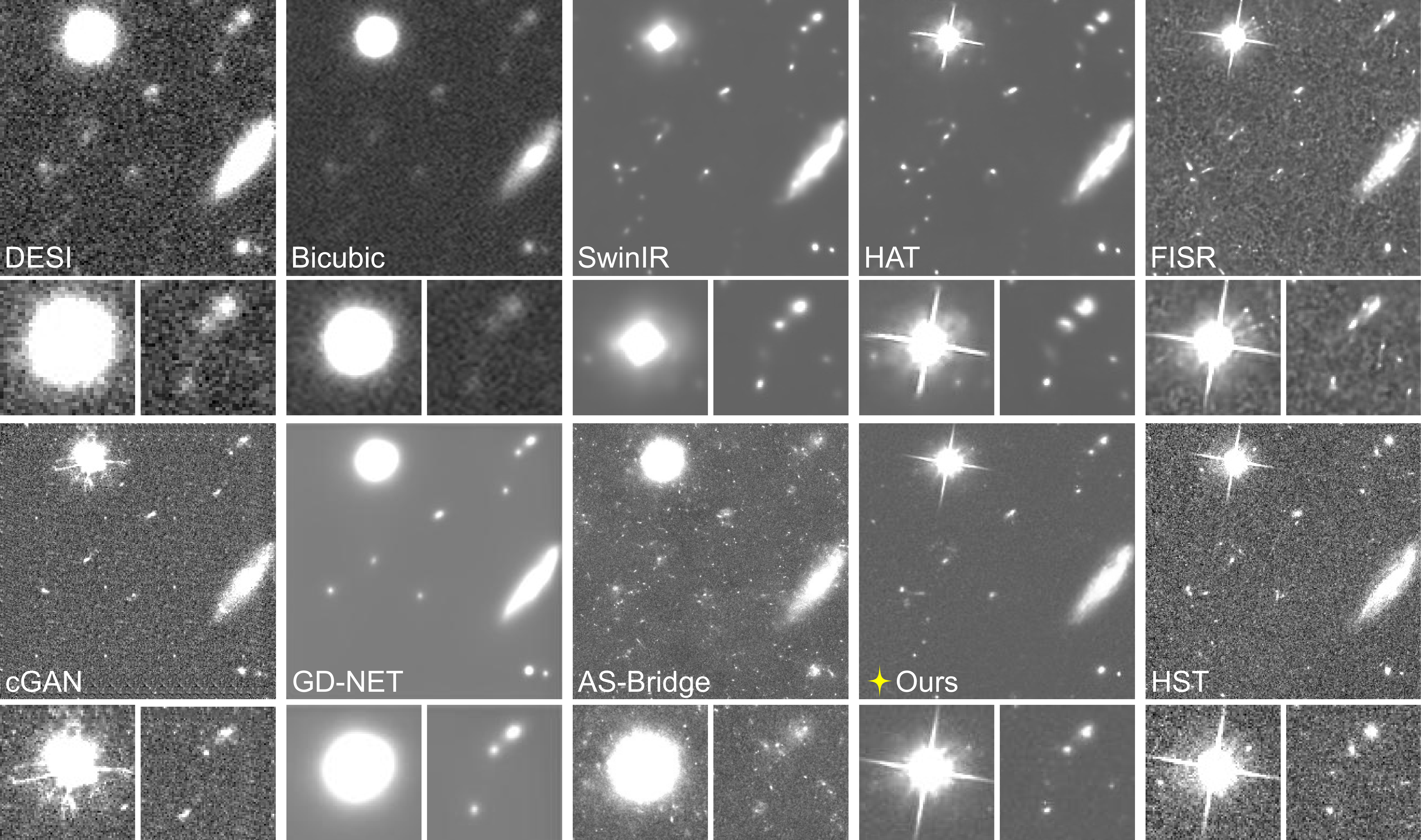}
    \caption{Qualitative comparison of baseline methods and FluxFlow on $\times2$ super-resolution.}
    \label{fig:result_x2}
\end{figure}

\begin{figure}[!tp]
    \centering
    \includegraphics[width=\linewidth]{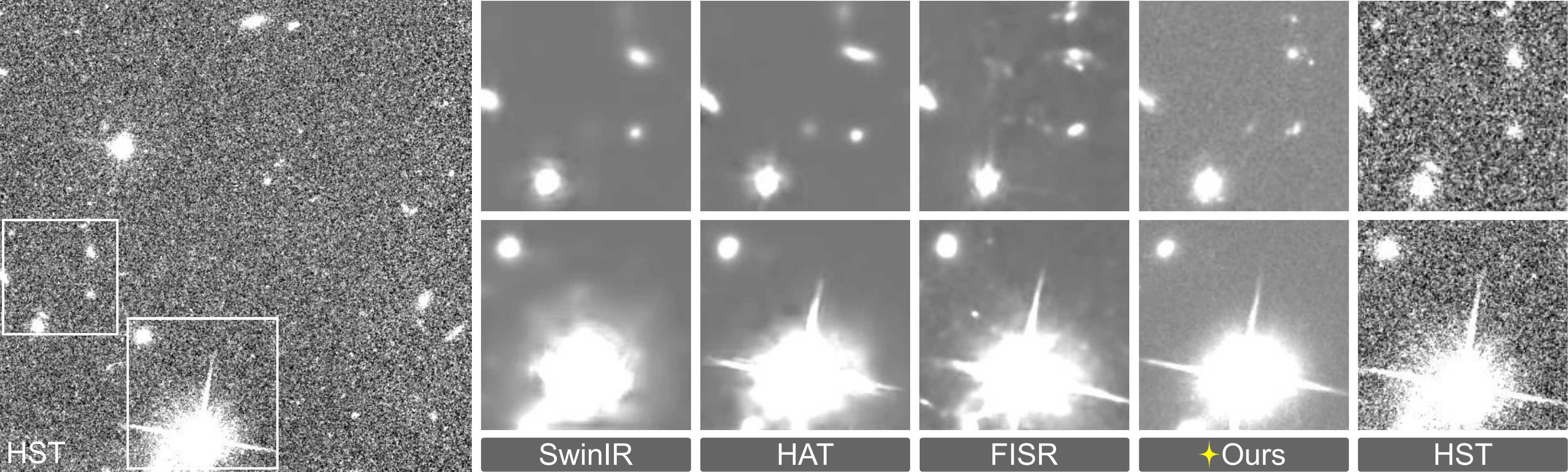}
    \caption{Qualitative comparison of baseline methods and FluxFlow on $\times4$ super-resolution.}
    \label{fig:result_x4}
\end{figure}

\begin{table}[!ht]
\centering
\footnotesize
\caption{Quantitative comparison at $\times2$ and $\times4$ on our DESI--HST super-resolution dataset.}
\label{tab:experiment}
\setlength{\tabcolsep}{3pt}
\renewcommand{\arraystretch}{1}
{\fontsize{8pt}{10pt}\selectfont
\begin{tabularx}{\textwidth}{ll|*{8}{>{\centering\arraybackslash}X}}
\hline
Scale & Metric & Bicubic & SwinIR & HAT & FISR & cGAN & GD-Net & AS-Bridge & \textbf{Ours} \\
\hline
   & PSNR$\uparrow$ & 26.69 & 31.22\thd & 31.31\fst & 30.93 & 28.88 & 28.82 & 28.38 & 31.23\sed \\
$\times2$ & SSIM$\uparrow$ & 0.594 & 0.702 & 0.711\sed & 0.704\thd & 0.579 & 0.692 & 0.626 & 0.714\fst \\
   & Flux-L1$\downarrow$ & 11.251 & 4.136 & 3.695\thd & 3.526\sed & 4.502 & 3.639 & 5.151 & 2.959\fst \\
\hline
   & PSNR$\uparrow$ & 22.74 & 29.15\sed & 29.21\fst & 29.02 & 26.48 & 24.64 & 24.72 & 29.14\thd \\
$\times4$ & SSIM$\uparrow$ & 0.415 & 0.569\thd & 0.570\fst & 0.568 & 0.477 & 0.496 & 0.514 & 0.570\fst \\
   & Flux-L1$\downarrow$ & 18.08 & 4.939 & 4.168\thd & 4.082\sed & 5.479 & 4.599 & 6.542 & 3.755\fst \\
\hline
\end{tabularx}
}
\end{table}

\subsection{Quantitative and Qualitative Evaluation}

\Cref{tab:experiment} shows quantitative results at $\times 2$ and $\times 4$ on our DESI--HST dataset. FluxFlow obtains PSNR within 0.1\,dB of the highest value at both scales, while achieving the best SSIM and Flux-L1 \DIFadd{among baseline methods}. \DIFaddbegin \DIFadd{PSNR and SSIM evaluate reconstruction fidelity, while Flux-L1 measures the absolute flux error within source regions. \Cref{fig:result_x2} presents qualitative comparisons at $\times 2$. Regression methods tend to oversmooth source cores and extended galaxy structures, whereas FluxFlow preserves these details and suppresses hallucinated sources in regions with low signal to noise.}\DIFaddend

\section{Ablation Study}

\DIFaddbegin \begin{minipage}[c]{0.48\linewidth}
    \DIFaddend \centering
    \includegraphics[width=\linewidth]{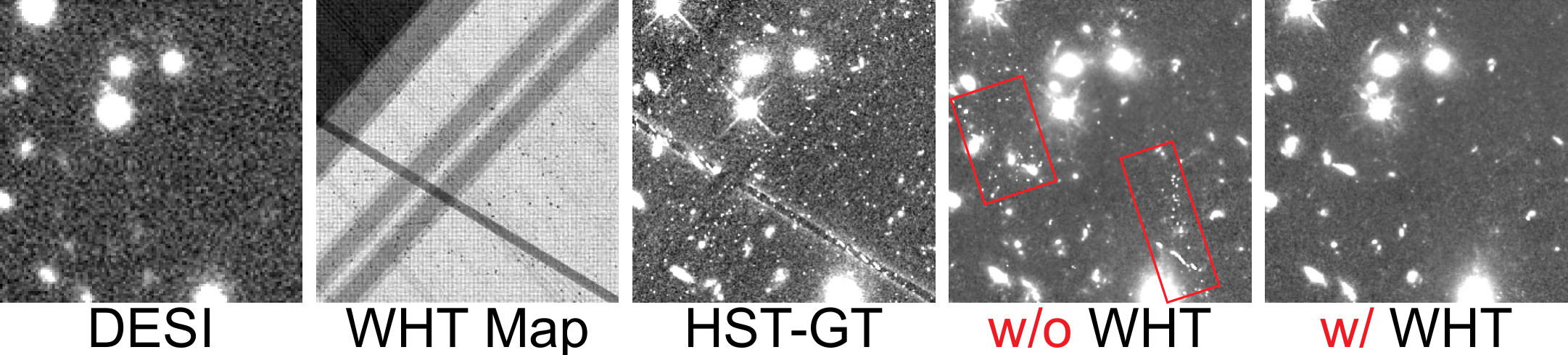}
    \captionof{figure}{WHT weighting ablation. Defects (cosmic rays, drizzle gaps) in red boxes.}
    \label{fig:ablation_wht}
    \includegraphics[width=\linewidth]{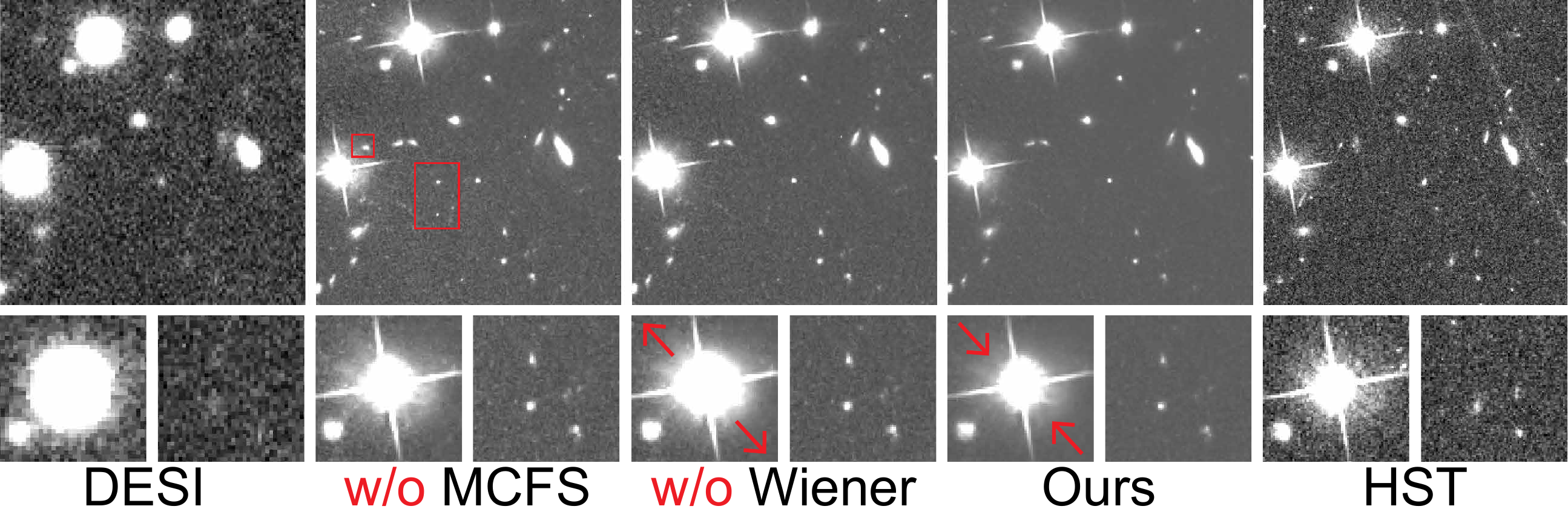}
    \captionof{figure}{MC-FS and Wiener-deconvolved correction ablation. Hallucinations in red boxes.}
    \label{fig:ablation_mcfs}
\end{minipage}%
\hfill
\DIFaddbegin \begin{minipage}[c]{0.50\linewidth}
    \DIFaddend \centering
    \captionof{table}{Ablation on photometry and detection metrics. \emph{WHT} denotes inverse-variance weighting, \emph{Src} denotes source-weighted loss, and MC-FS and Wiener correction are described in \cref{sec:mcfm}.}
    \label{tab:ablation}
    \footnotesize
    \setlength{\tabcolsep}{1.8pt}
    \begin{tabular}{l|ccc|ccc}
    \toprule
    & \multicolumn{3}{c|}{Photometry} & \multicolumn{3}{c}{Detection} \\
    Ablation & PSNR$\uparrow$ & SSIM$\uparrow$ & FluxL1$\downarrow$ & Prec$\uparrow$ & Rec$\uparrow$ & F1$\uparrow$ \\
    \midrule
    w/ VAE & 28.90 & 0.593 & 5.075 & \fst 0.859 & 0.211 & 0.3402 \\
    w/ Bridge & 30.77 & 0.663 & 3.206 & 0.702 & 0.393 & 0.5019 \\
    \midrule
    w/o WHT & 30.17 & 0.685 & \fst 2.814 & 0.688 & \sed 0.483 & \thd 0.5272 \\
    w/o Src & \sed 31.16 & \sed 0.711 & 3.098 & \thd 0.729 & \thd 0.464 & \sed 0.5382 \\
    w/o MCFS  & 30.78 & 0.698 & \thd 3.025 & 0.602 & \fst 0.553 & 0.5256 \\
    w/o Wiener & \thd 30.79 & \thd 0.699 & 3.031 & 0.708 & 0.448 & 0.5066 \\
    \midrule
    \textbf{Ours} & \fst 31.23 & \fst 0.714 & \sed 2.959 & \sed 0.738 & 0.458 & \fst 0.5412 \\
    \bottomrule
    \end{tabular}
\end{minipage}

\paragraph{Inverse-Variance and Source-Region Weighting.}
\Cref{fig:ablation_wht} visualizes a region containing a cosmic-ray trail and a drizzle chip-gap in the HST . Without the WHT term defined in \Cref{eq:wfm_loss}, the velocity field treats these defects as valid targets and reproduces them in the reconstruction. The WHT-weighted variant suppresses their gradients and yields a clean output, and \Cref{tab:ablation} reports the corresponding drops in PSNR, F1, and precision.The marginally lower Flux-L1 of w/o WHT is an acceptable trade-off, since it arises from fitting drizzle artifacts that overlap with source apertures.The w/o source-weight variant similarly degrades, as a uniform loss is dominated by background pixels and cannot preferentially preserve the faint outer galaxy profiles that carry morphological and PSF-shape information. The two weights are complementary, with WHT filtering data quality and source-weight emphasizing scientifically informative regions.

\paragraph{Measurement-Consistent Sampling.}
We ablate MC-FS and the Wiener-regularization separately to isolate their roles. Without MC-FS, the sampler produces compact hallucination sources shown in \Cref{fig:ablation_mcfs}, and \Cref{tab:ablation} reports the highest recall together with a precision collapse, as these spurious peaks inflate false positives and pull F1 below the full model. The w/o Wiener variant replaces the Wiener correction of \Cref{eq:wiener_correction} with the raw adjoint of \Cref{eq:adjoint_correction} and visibly broadens the reconstructed PSF via the cumulative smoothing of Appendix~\ref{sup:mcfs:wiener_derivation}. The broadened response smears source flux beyond the target apertures into the background, and \Cref{tab:ablation} reports a uniform photometric degradation across PSNR, SSIM, and Flux-L1. Our full model attains the strongest photometric scores together with the best detection F1, confirming that both inference-time components are necessary. Additional ablation studies on latent-space and bridge-based OT-CFM, hyperparameters, and visualizations of detection-outcome are provided in Appendix~\ref{sec:sup:ablation} and Appendix~\ref{sup:sec:visualization_sample}.

\section{Conclusion}
We present FluxFlow, a conservative flow-matching framework for ground-to-space astronomical super-resolution. The training objective  \DIFaddbegin \DIFadd{combines inverse-variance }\DIFaddend and source-region  \DIFaddbegin \DIFadd{weighting}\DIFaddend ,  \DIFaddbegin \DIFadd{while }\DIFaddend measurement-consistent sampling  \DIFaddbegin \DIFadd{applies }\DIFaddend \DIFaddbegin \DIFadd{Wiener correction under an approximate forward model}\DIFaddend .  \DIFaddbegin \DIFadd{The }\DIFaddend DESI--HST dataset  \DIFaddbegin \DIFadd{pairs independent ground- and space-based observations}\DIFaddend .  \DIFadd{Experiments show that FluxFlow delivers superior photometric and scientific accuracy over existing baselines.}

\paragraph{Limitation.}
Hallucinations are suppressed rather than eliminated\DIFaddbegin \DIFadd{. The HST training targets contain residual noise, instrumental defects, optical diffraction patterns, and differences caused by the bandpasses, so reconstructions can inherit these features. The fixed isotropic Gaussian PSF does not capture variation across exposures, positions, or wavelengths, nor does it model anisotropy or extended wings. Future work will investigate generalization to other fields and instruments, source discovery, and calibrated uncertainty.

}

\section*{\DIFadd{Acknowledgements and disclosures}}
\DIFadd{This work was partially supported by JST Moonshot R\&D Grant Number JPMJPS2011 and by }\DIFaddend the  \DIFaddbegin \DIFadd{New Energy and Industrial Technology Development Organization (NEDO) under project JPNP26007}\DIFaddend . \DIFaddbegin \DIFadd{Ziteng Cui was supported by JSPS KAKENHI Grant Number JP26K21296. Shuhong Liu was supported by JST BOOST, Japan Grant Number JPMJBS2418.

}\DIFaddend 

\bibliographystyle{plainnat}
\bibliography{sample}


\appendix

\begin{figure}[!ht]
    \centering
    \includegraphics[width=\textwidth]{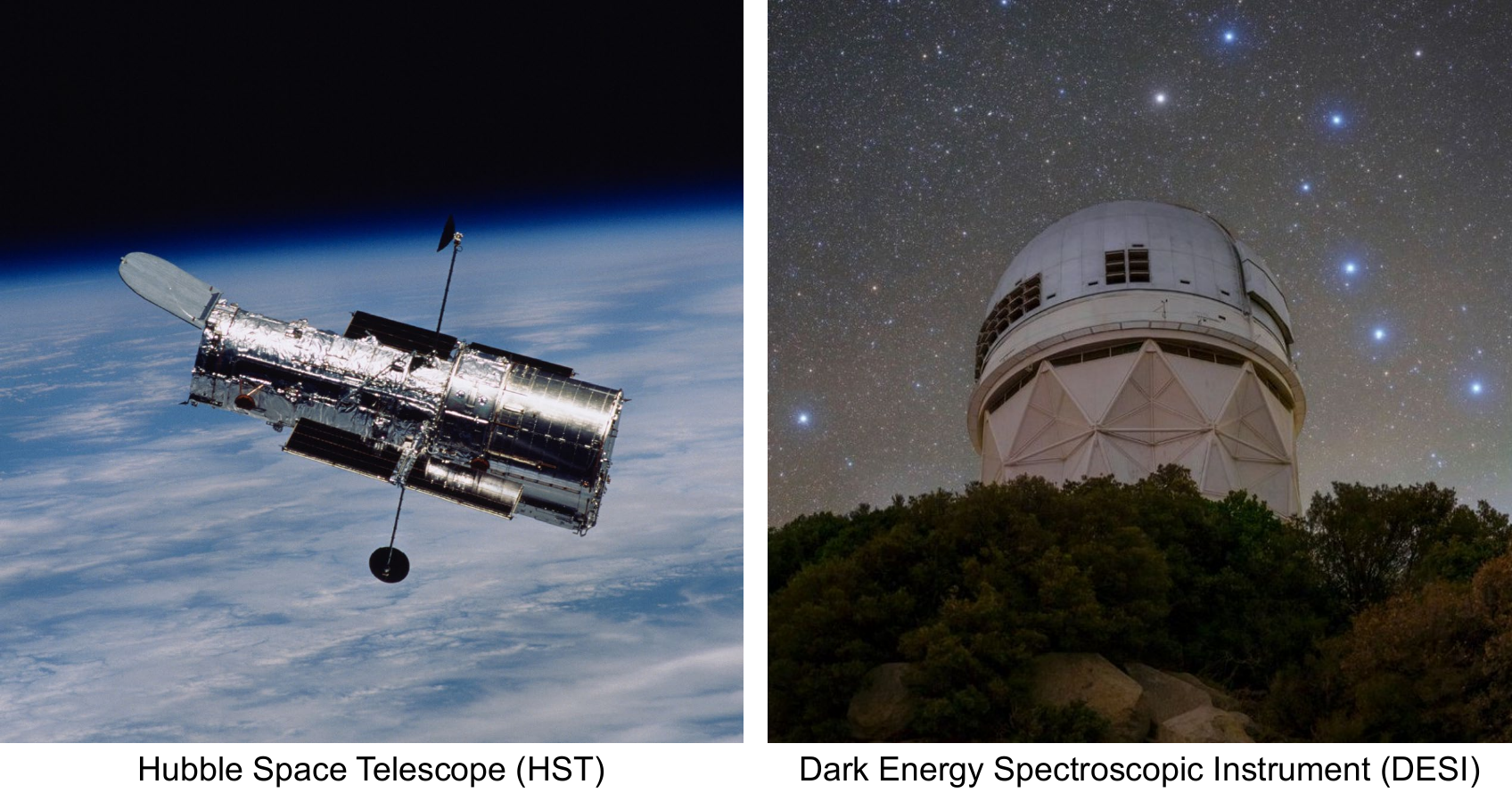}
    \caption{\DIFaddbeginFL \DIFaddFL{The left image shows }\DIFaddendFL HST\DIFaddbeginFL \DIFaddFL{, which provides the HR targets. The right image shows the Mayall telescope, which hosts the DESI spectrograph. The DECam $i$ band LR images used here were obtained with the Blanco telescope. Image credits are NASA for }\DIFaddendFL HST \DIFaddbeginFL \DIFaddFL{and KPNO/NOIRLab/NSF/AURA/T. Slovinsk\'{y} for Mayall.}\DIFaddendFL}
    \label{fig:sup_observer}
\end{figure}

\begin{figure}[!t]
    \centering
    \includegraphics[width=\textwidth]{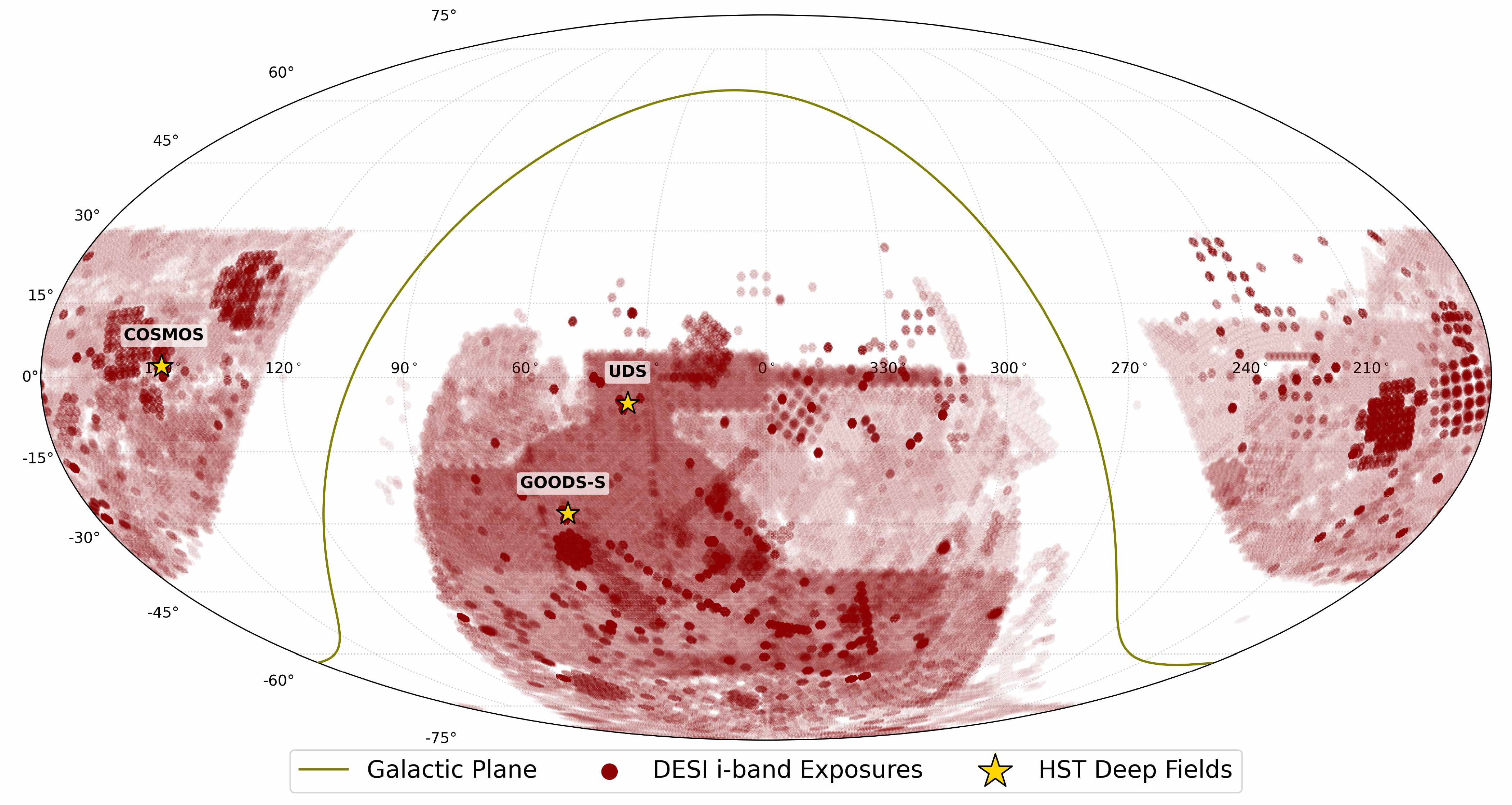}
    \caption{The footprint of the Legacy Surveys (DR10) i-band observations (dark red) overlaid with the locations of three equatorial and southern HST deep fields: COSMOS, UDS, and GOODS-S (gold stars). The olive green curve indicates the Galactic plane.}
    \label{fig:sup_footprint}
\end{figure}

\section{Survey Data and Instrument Characteristics}\label{sec:appendixa}


To fully appreciate the complexity of mapping the ground-based low-resolution (LR) observations to the space-based high-resolution (HR) targets, it is necessary to detail the disparate instrument characteristics ( Figure~\ref{fig:sup_observer}) and image co-addition pipelines of the DESI Legacy Imaging Surveys and the Hubble Space Telescope (HST), whose overlapping observational footprints are shown in Figure~\ref{fig:sup_footprint}.

\subsection{DESI Legacy Surveys: Ground-Based Seeing and Co-addition}

The LR images from the DESI Legacy Imaging Surveys are fundamentally limited by atmospheric turbulence, which blurs the astronomical signal into a seeing-limited Point Spread Function (PSF). The $i$-band data (${\sim}7600$--$8730$\,\AA) utilized in our benchmark, which offers the closest spectral match to the HR targets, is primarily acquired by the Dark Energy Camera (DECam), a 570-megapixel wide-field imager mounted at the prime focus of the Victor M. Blanco 4-meter telescope at the Cerro Tololo Inter-American Observatory (CTIO). 

Despite the substantial light-gathering power of the 4-meter primary mirror, ground-based observations are subject to atmospheric degradation. With DECam's native pixel scale of $0.262''$/pixel, the typical seeing Full Width at Half Maximum (FWHM) for these observations ranges from $1.0''$ to $1.3''$. The final LR images provided in our benchmark are co-added mosaics. The data processing pipeline maps multiple individual dithered exposures onto a common predefined sky footprint (often referred to as "bricks"). This co-addition process includes sky-background subtraction, astrometric calibration, and standard inverse-variance weighting. While this improves the signal-to-noise ratio (SNR) and depth, the resulting images still inherently suffer from the spatially varying PSF blur and elevated sky background noise intrinsic to ground-based optical astronomy.

\DIFaddbegin \DIFadd{Long exposures average short-timescale atmospheric speckles into an effective seeing PSF. Under Kolmogorov turbulence with $D/r_0\gg1$, the angular FWHM is approximately $0.98\lambda/r_0$, compared with the diffraction scale $\lambda/D$~\cite{fried1966resolution,roddier1981turbulence,tokovinin2002seeing}. Here $D$ is aperture diameter, $r_0$ the atmospheric coherence length, and $\lambda$ the observing wavelength. This asymptotic relation does not identify the fitted Gaussian width measured in pixels. Conversion depends on plate scale, and the effective coadd response also includes instrumental and registration effects.

}

\DIFaddend \subsection{HST ACS/WFC: Space-Based Optics and Astrodrizzle}

In contrast, the HR targets are obtained by the Hubble Space Telescope (HST), which orbits at an altitude of approximately 540 km. Operating entirely above the Earth's atmosphere, HST's 2.4-meter primary mirror delivers a highly stable, diffraction-limited PSF that is entirely immune to atmospheric seeing. The Advanced Camera for Surveys Wide Field Channel (ACS/WFC) in the F814W filter (${\sim}7000$--$9500$\,\AA) serves as the primary instrument for our HR data. It consists of two 2048 $\times$ 4096 pixel charge-coupled devices (CCDs) separated by a physical chip gap, delivering a native pixel scale of approximately $0.05''$/pixel. 

As introduced in the main text, raw HST exposures inherently suffer from this detector undersampling and transient cosmic-ray events. To mitigate these issues and achieve the final $0.03''$/pixel grid, the HST pipeline employs  Astrodrizzle , \DIFaddbegin \DIFadd{based on Drizzle~\cite{fruchter2002drizzle,gonzaga2012drizzlepac}, }\DIFaddend which combines multiple sub-pixel dithered exposures. Instead of a simple direct co-addition, the algorithm geometrically shrinks the footprint of the input pixels (controlled by the \textit{pixfrac} parameter) and "drizzles" their flux onto a finer output grid. This sub-pixel interlacing effectively recovers high-frequency spatial structures that approach the optical diffraction limit of the 2.4-meter aperture, successfully bridging the gap between the native $0.05''$ detector resolution and the final $0.03''$ product.

\subsection{Inverse-Variance Weight Maps for HST Observations}

Concurrently, the generation of the HST inverse-variance weight maps (\textit{wht}) is mathematically deterministic and crucial for our benchmark. Fundamentally, the value of each pixel in these maps represents the inverse variance ($1/\sigma^2$) of the measured flux. This variance ($\sigma^2$) is not an arbitrary metric but is rigorously derived from the intrinsic hardware characteristics of the ACS/WFC CCDs and the physics of photon counting. Specifically, the pipeline calculates the total noise per pixel by modeling the detector's read noise (electronic noise generated during signal readout), dark current (thermal electron accumulation), and the Poisson noise originating from the sky background.

During the Astrodrizzle process, as multiple dithered exposures are mapped onto the final output grid, these individual pixel variances are mathematically propagated. Because the final HR mosaic is constructed from these frames with varying overlap footprints, the effective signal-to-noise ratio (SNR) is intrinsically non-uniform across the field of view. 

The weight maps  \DIFaddbegin \DIFadd{encode the pipeline-estimated statistical reliability and masks, reducing the contribution of low-weight or invalid pixels to the training loss}\DIFaddend . They  \DIFaddbegin \DIFadd{do not identify every defect and do not remove stable optical features such as diffraction spikes }\DIFaddend or  \DIFaddbegin \DIFadd{flare}\DIFaddend . In  \DIFaddbegin \DIFadd{particular}\DIFaddend ,  \DIFaddbegin \DIFadd{weighting }\DIFaddend the loss  \DIFaddbegin \DIFadd{does not convert the HST target into an artifact-free sky image}\DIFaddend .

\section{Precision WCS-Based Image Registration}\label{sec:appendixb}

To learn reliable cross-device mappings, eliminating translational, rotational, and geometric distortions between the ground-based and space-based platforms is a prerequisite. We achieve absolute spatial anchoring by utilizing the World Coordinate System (WCS) metadata embedded within the Flexible Image Transport System (FITS) headers of both instruments.

The WCS framework provides a rigorous non-linear mathematical transformation mapping 2D detector coordinates $(x,y)$ to absolute celestial coordinates (Right Ascension $\alpha$ and Declination $\delta$ in the J2000 epoch), while properly accounting for telescope-specific focal plane distortions. Because our objective is to map DESI observations to higher-resolution HST counterparts at exact scaling factors ($\times 2$ and $\times 4$), we establish a common celestial   grid anchored directly to the DESI footprint.

We establish the native DESI pixel grid as the absolute spatial \DIFadd{reference}, retaining the DESI coadds at native resolution to preserve their noise properties and PSF. Based on this anchor, we define a corresponding high-resolution grid that precisely inherits the DESI tangent plane center and orientation, with its pixel scale divided by the targeted super-resolution factor. The HST exposures are then mathematically reprojected and flux-conservingly downsampled onto this customized grid via their native WCS metadata,  \DIFaddbegin \DIFadd{establishing a shared sampling grid. Residual astrometric errors may remain}\DIFaddend . 

Finally, the resulting perfectly aligned LR--HR pairs are cropped into spatially corresponding patches to produce the final dataset elements, yielding HR targets structured at $256\times256$ and $512\times512$ resolutions for the $\times 2$ and $\times 4$ scales, respectively. This strict astrometric registration eliminates spurious spatial shifts, compelling the neural network to learn genuine morphological transformations rather than compensating for coordinate misalignments.

\DIFaddbegin \section{\DIFadd{Scope and Scientific Use}}
\label{sec:reference_scope}

\DIFadd{HST is an observational target, not artifact-free truth. Its limited overlap, noise, defects, bandpass, and optical structures constrain supervision and evaluation. WHT weighting reduces the learning of low-confidence defects such as chip gaps and cosmic-ray trails, but cannot remove every stable instrumental structure, including diffraction and flare patterns. The reported metrics evaluate photometric and morphological reconstruction accuracy against the HST observations. Scientific deployment requires independent validation.

}

\DIFadd{FluxFlow cannot reliably recover a source absent from its HST supervision but present in another catalog, because such a source provides no positive HR training signal. MC-FS may preserve coarse evidence already present in the LR image, but cannot distinguish a real object from noise, transients, bandpass or depth differences, or model error. This dependence on available pairs limits discovery use. Unmatched reconstructed sources require independent catalog, multi-band, or multi-epoch validation.

}

\DIFadd{We select the DESI $i$ band and HST F814W because they are the closest available bandpasses between the two observing programs. The LR/HR band configuration is identical across training and testing. HST images, catalogs, and WHT maps are not required at inference. FluxFlow receives only the LR/DESI observation, and MC-FS compares the forward projection of the reconstruction with that same input.

}

\DIFadd{Future work will investigate metadata-conditioned anisotropic Moffat profiles, spatially varying PSF fields, and per-source adaptive PSF estimation and reprojection, as well as multi-instrument and multi-epoch  data.

}

\DIFaddend \section{Additional Related Works}
\label{sup:mcfs:related}

MC-FS draws on a line of research that injects measurement-consistency corrections into the sampling trajectories of generative models for inverse problems. We clarify its relationship to the most relevant prior approaches.

\subsection{Diffusion Posterior Sampling}

\DIFaddbegin 

\DIFaddend Diffusion Posterior Sampling (DPS)~\cite{chung2022diffusion} approximates the posterior score by adding to each reverse diffusion step a gradient of the log-likelihood evaluated at the Tweedie estimate $\hat{\mathbf{x}}_0(\mathbf{x}_t)$ using:
\begin{equation}
    \mathbf{x}_{t-1}
     \leftarrow
     \mathbf{x}_{t-1}^{\mathrm{prior}}
    -
    \zeta  \nabla_{\mathbf{x}_t}
     \DIFaddbegin \left\|
        \DIFaddend \mathbf{y}
        -
        \mathbf{A} \DIFaddbegin \bigl(\DIFaddend \hat{\mathbf{x}}_0(\mathbf{x}_t) \DIFaddbegin \bigr)
    \right\|\DIFaddend ^2.
\end{equation}
The gradient propagates through the denoiser via automatic differentiation, so DPS effectively composes the Jacobian of the score network with $\mathbf{A}^{\!\top}$. Pseudoinverse-Guided Diffusion Models ($\Pi$GDM)~\cite{song2023pseudoinverse} replace $\mathbf{A}^{\!\top}$ with the Moore--Penrose pseudoinverse $\mathbf{A}^{+}$ to obtain a more accurate posterior approximation, again applied to the Tweedie estimate.

 \DIFaddbegin \DIFadd{FluxFlow learns }\DIFaddend a  \DIFaddbegin \DIFadd{paired conditional transport toward $p_\theta(\mathbf{x}_{\mathrm{HST}}\mid\mathbf{y}_{\mathrm{DESI}})$}\DIFaddend ,  \DIFaddbegin \DIFadd{whereas conventional }\DIFaddend posterior  \DIFaddbegin \DIFadd{solvers typically start from an unconditional image prior}\DIFaddend . Under the  linear  \DIFaddbegin \DIFadd{flow parameterization}\DIFaddend , the  \DIFaddbegin \DIFadd{current }\DIFaddend velocity  \DIFaddbegin \DIFadd{already gives }\DIFaddend a  one-step \DIFaddbegin \DIFadd{destination }\DIFaddend estimate  \DIFaddbegin \DIFadd{$\hat{\mathbf{x}}_{1\mid t}=\mathbf{x}_t+(1-t)v_\theta(\mathbf{x}_t,t,\mathbf{c})$. No }\DIFaddend additional clean-image predictor is \DIFaddbegin \DIFadd{required}\DIFaddend .  \DIFaddbegin \DIFadd{MC-FS instead evaluates its residual on }\DIFaddend the  \DIFaddbegin \DIFadd{Euler proposal $\tilde{\mathbf{x}}^{(i+1)}$}\DIFaddend ,  \DIFaddbegin \DIFadd{deliberately injecting measurement information early }\DIFaddend in  \DIFaddbegin \DIFadd{the trajectory}\DIFaddend .  \DIFaddbegin \DIFadd{Its Wiener correction avoids }\DIFaddend differentiation through the velocity network\DIFaddbegin \DIFadd{. This }\DIFaddend is  \DIFaddbegin \DIFadd{a reconstruction heuristic supported by }\DIFaddend the  \DIFaddbegin \DIFadd{stated spectral model and experiments, not an exact posterior score or probability-flow ODE}\DIFaddend . \Cref{tab:mcfs_comparison}  \DIFaddbegin \DIFadd{contrasts the original diffusion formulations with MC-FS}\DIFaddend .

\begin{table}[!tp]
    \centering
    \setlength{\tabcolsep}{2.8pt}
    \footnotesize
    \caption{ \DIFaddbeginFL \DIFaddFL{Conceptual comparison }\DIFaddendFL with \DIFaddbeginFL \DIFaddFL{the original }\DIFaddendFL diffusion-based posterior sampling  \DIFaddbeginFL \DIFaddFL{formulations}\DIFaddendFL .  \DIFaddbeginFL \DIFaddFL{The residual }\DIFaddendFL is  \DIFaddbeginFL \DIFaddFL{evaluated at }\DIFaddendFL a  \DIFaddbeginFL \DIFaddFL{one-step clean-image estimate for DPS }\DIFaddendFL and  \DIFaddbeginFL \DIFaddFL{$\Pi$GDM}\DIFaddendFL , and  \DIFaddbeginFL \DIFaddFL{at the Euler proposal for MC-FS}\DIFaddendFL .}
    \label{tab:mcfs_comparison}
    \begin{tabular}{lcccc}
        \toprule
        Method
        & Generative backbone
        & Residual point
        & Back-projection
        & Autograd through network \\
        \midrule
        DPS~\cite{chung2022diffusion}
        & Diffusion
        & $\hat{\mathbf{x}}_0(\mathbf{x}_t)$
        & $\mathbf{A}^{\!\top}$ via Jacobian
        & Required \\
        $\Pi$GDM~\cite{song2023pseudoinverse}
        & Diffusion
        & $\hat{\mathbf{x}}_0(\mathbf{x}_t)$
        & Pseudoinverse $\mathbf{A}^{+}$
        & Required \\
        \DIFadd{\textbf{Ours}}
        & Flow matching
        & $\tilde{\mathbf{x}}^{(i+1)}$
        & Wiener-regularized inverse
        & Not required \\
        \bottomrule
    \end{tabular}
\end{table}

\subsection{Plug-and-Play Frameworks  \DIFaddbegin \DIFadd{for }\DIFaddend Inverse Problems}

 \DIFaddbegin 

\DIFaddend MC-FS  \DIFaddbegin \DIFadd{is related to }\DIFaddend the \DIFaddbegin \DIFadd{alternating prior and data-fidelity structure of }\DIFaddend plug-and-play  \DIFaddbegin \DIFadd{methods}\DIFaddend ~\cite{venkatakrishnan2013plug,romano2017little,zhang2021plug,hurault2022proximal,kamilov2023plug,zhu2023denoising}.  \DIFaddbegin \DIFadd{The learned conditional velocity proposes an update and the residual filtered by the Wiener kernel supplies }\DIFaddend a  \DIFaddbegin \DIFadd{soft correction for measurement consistency. The relationship is an analogy because the blur-domain Wiener filter composed with area-downsampling adjoints is not, in general, the exact proximal operator of the full blur-and-downsampling data term.

}

\DIFadd{MC-FS does not perform a hard projection onto $\mathcal{C}(\mathbf{y})$ }\DIFaddend and \DIFaddbegin \DIFadd{does not guarantee posterior samples. For }\DIFaddend a  \DIFaddbegin \DIFadd{fixed initial Gaussian draw, its updates are deterministic. Variation across initial draws is not calibrated posterior uncertainty. Its purpose here is stable conditional reconstruction. Stochastic, uncertainty-calibrated sampling is left to future work. The design targets one conservative reconstruction per observation. Refreshing noise at every }\DIFaddend step \DIFaddbegin \DIFadd{could reintroduce compact fluctuations }\DIFaddend that  \DIFaddbegin \DIFadd{the }\DIFaddend consistency \DIFaddbegin \DIFadd{correction is intended to attenuate. The initial Gaussian draw retains stochasticity, while the subsequent updates introduce no fresh noise.

}

\subsection{\DIFadd{Flow-Matching for Inverse Problems}}

\DIFadd{Efficient SR approaches such as SinSR~\cite{wang2024sinsr}, OSEDiff~\cite{wu2024one}, and TSD-SR~\cite{dong2025tsd} are relevant to reducing sampling cost. Applying their natural-image pretrained models directly to single-band flux-calibrated observations would conflate efficiency }\DIFaddend with \DIFaddbegin \DIFadd{domain and degradation mismatch. A controlled comparison requires adaptation to }\DIFaddend the  \DIFaddbegin \DIFadd{same paired data, normalization, uncertainty handling, and source-level evaluation}\DIFaddend .

 \DIFaddbegin 

\DIFadd{FlowDPS~\cite{kim2025flowdps} decomposes flow updates into destination-image and noise components}\DIFaddend , \DIFaddbegin \DIFadd{enabling likelihood guidance and stochastic noise refresh. We adapt DPS, $\Pi$GDM, FlowDPS, and FLOWER updates to a shared paired conditional FluxFlow checkpoint to isolate }\DIFaddend the  \DIFaddbegin \DIFadd{effect }\DIFaddend of  the  \DIFaddbegin \DIFadd{sampling update.

}

\paragraph{\DIFadd{Comparison of sampling methods.}}

\DIFadd{The comparison keeps }\DIFaddend the  \DIFaddbegin \DIFadd{paired conditional velocity network fixed and changes }\DIFaddend the  \DIFaddbegin \DIFadd{inference update. The DESI condition $\mathbf{c}$ remains present }\DIFaddend in \DIFaddbegin \DIFadd{every network call. These samplers share the following destination estimate and measurement residual.
}\begin{equation}
    \DIFadd{\hat{\mathbf{x}}_{1\mid t}
    =
    \mathbf{x}_t
    +
    (1-t)v_\theta(\mathbf{x}_t,t,\mathbf{c}),
    \qquad
    \mathbf{r}_t
    =
    \mathbf{A}\hat{\mathbf{x}}_{1\mid t}
    -
    \mathbf{y}.
    \label{eq:adapted_destination}
}\end{equation}

\DIFadd{Writing $\mathbf{J}_{\hat{x}}=\partial\hat{\mathbf{x}}_{1\mid t}/\partial\mathbf{x}_t$}\DIFaddend ,  \DIFaddbegin \DIFadd{DPS differentiates }\DIFaddend the  \DIFaddbegin \DIFadd{destination residual through }\DIFaddend the  \DIFaddbegin \DIFadd{conditional network}\DIFaddend .  \DIFaddbegin \DIFadd{$\Pi$GDM replaces the adjoint residual with }\DIFaddend a  \DIFaddbegin \DIFadd{regularized pseudoinverse residual before applying the destination Jacobian as follows.
}\begin{equation}
    \DIFadd{\begin{aligned}
        \mathbf{g}_{\mathrm{DPS}}
        &=
        \nabla_{\mathbf{x}_t}
        \tfrac12\|\mathbf{r}_t\|_2^2
        =
        \mathbf{J}_{\hat{x}}^{\!\top}
        \mathbf{A}^{\!\top}\mathbf{r}_t,
        \\
        \mathbf{g}_{\Pi\mathrm{GDM}}
        &=
        \mathbf{J}_{\hat{x}}^{\!\top}
        \mathbf{A}^{\dagger}_{\epsilon}\mathbf{r}_t.
    \end{aligned}
    \label{eq:adapted_guidance}
}\end{equation}

\DIFadd{Here $\mathbf{A}^{\dagger}_{\epsilon}$ denotes the regularized pseudoinverse used }\DIFaddend in  the  \DIFaddbegin \DIFadd{adaptation. Its regularization }\DIFaddend is  \DIFaddbegin \DIFadd{distinct from MC-FS's Wiener parameter $\lambda_{\mathrm{SNR}}$.

}

\DIFadd{FlowDPS refines the destination using the pseudoinverse }\DIFaddend and \DIFaddbegin \DIFadd{recombines it with a refreshed source-noise endpoint according to }\DIFaddend the \DIFaddbegin \DIFadd{flow schedule. FLOWER applies a }\DIFaddend data-fidelity proximal  \DIFaddbegin \DIFadd{refinement to }\DIFaddend the  \DIFaddbegin \DIFadd{destination and reprojects }\DIFaddend the  \DIFaddbegin \DIFadd{refined estimate along the flow path}\DIFaddend .  \DIFaddbegin \DIFadd{

}

\DIFaddend MC-FS  \DIFaddbegin \DIFadd{instead evaluates $\mathbf{A}\tilde{\mathbf{x}}^{(i+1)}-\mathbf{y}$ at the conditional Euler proposal }\DIFaddend and  \DIFaddbegin \DIFadd{applies }\DIFaddend the  \DIFaddbegin \DIFadd{Wiener correction in \Cref{eq:wiener_correction}}\DIFaddend . \DIFaddbegin \DIFadd{Thus it corrects the current trajectory directly, without network-Jacobian backpropagation or an iterative pseudoinverse/proximal solve. }\DIFaddend The  \DIFaddbegin \DIFadd{linear MMSE and finite-step spectral analysis motivate this choice, but do not establish an exact }\DIFaddend posterior \DIFaddbegin \DIFadd{law for the corrected nonlinear trajectory.}\DIFaddend

 A growing body of work studies flow matching directly in pixel space with various weighting and parameterization choices~\cite{chen2025provably,gagneux2026training}, and develops flow-matching solvers and priors specifically tailored to linear inverse problems~\cite{pourya2025flower,zhang2024flow}. WienerFlow~\cite{zeng2025wienerflow} introduces Wiener-adaptive flow matching in the related setting of low-light image enhancement and is closest in spirit to MC-FS,  \DIFaddbegin \DIFadd{but it changes }\DIFaddend the  \DIFaddbegin \DIFadd{training formulation }\DIFaddend and  \DIFaddbegin \DIFadd{is not a drop-in sampling update for }\DIFaddend the  \DIFaddbegin \DIFadd{same fixed conditional checkpoint. Our }\DIFaddend correction \DIFaddbegin \DIFadd{instead targets an approximate ground-to-space super-resolution operator}\DIFaddend . In astronomy, score-based generative priors have been applied to multi-band source separation~\cite{sampson2024score} and broader astrophysical applications~\cite{ting2025deep}\DIFadd{.} FluxFlow complements these by focusing on the photometrically conservative ground-to-space mapping under realistic atmospheric PSFs.

\paragraph{\DIFadd{Quantitative sampler comparison.}}
\DIFadd{We compare these sampling methods at $\times 2$ using the same FluxFlow checkpoint, loss, training strategy, 1,701-image test split, 10-step protocol, and calibrated forward model with $\sigma_{\mathrm{PSF}}=2$. \Cref{tab:conditional_sampler_comparison} reports the results. FLOWER achieves the highest PSNR and F1, while MC-FS achieves the best SSIM, Flux-L1, and precision and remains competitive in PSNR and F1. These results isolate the sampling update under a shared learned conditional distribution.}

\begin{table}[!htbp]
\centering
\small
\caption{\DIFadd{Conditional sampler comparison on DESI--HST at $\times 2$ with a shared checkpoint and 10 sampling steps.}}
\label{tab:conditional_sampler_comparison}
\begin{tabular*}{\linewidth}{@{\extracolsep{\fill}}lrrrrrr}
\toprule
\DIFadd{Method} & \DIFadd{PSNR $\uparrow$} & \DIFadd{SSIM $\uparrow$} & \DIFadd{Flux-L1 $\downarrow$} & \DIFadd{Precision $\uparrow$} & \DIFadd{Recall $\uparrow$} & \DIFadd{F1 $\uparrow$} \\
\midrule
\DIFadd{DPS} & \DIFadd{30.56} & \DIFadd{0.653} & \DIFadd{3.197} & \DIFadd{0.621} & \fst \DIFadd{\textbf{0.503}} & \DIFadd{0.5230} \\
\DIFadd{$\Pi$GDM} & \DIFadd{30.75} & \DIFadd{0.686} & \DIFadd{3.060} & \sed \DIFadd{0.711} & \DIFadd{0.456} & \DIFadd{0.5252} \\
\DIFadd{FlowDPS} & \sed \DIFadd{31.26} & \thd \DIFadd{0.705} & \thd \DIFadd{2.986} & \DIFadd{0.654} & \sed \DIFadd{0.490} & \thd \DIFadd{0.5323} \\
\DIFadd{FLOWER} & \fst \DIFadd{\textbf{31.32}} & \sed \DIFadd{0.709} & \sed \DIFadd{2.977} & \thd \DIFadd{0.708} & \thd \DIFadd{0.469} & \fst \DIFadd{\textbf{0.5449}} \\
\DIFadd{\textbf{Ours}} & \thd \DIFadd{31.23} & \fst \DIFadd{\textbf{0.714}} & \fst \DIFadd{\textbf{2.959}} & \fst \DIFadd{\textbf{0.738}} & \DIFadd{0.458} & \sed \DIFadd{0.5412} \\
\bottomrule
\end{tabular*}
\end{table}

\section{Implementation Details}
\label{sup:implementation}

\paragraph{Source  \DIFaddbegin \DIFadd{Labels }\DIFaddend on  \DIFaddbegin \DIFadd{the HST}\DIFaddend .}
 \DIFaddbegin \DIFadd{HST }\DIFaddend instance labels are produced by running SEP~\cite{bertin1996sextractor,barbary2016sep} on each HST cutout, using the accompanying weight map to construct a validity mask in which pixels with non-positive weight are excluded from both the background statistics and the detection step. We estimate the local sky on a $32\times32$ mesh with no additional median filtering, subtract the resulting background plane from the science image, and obtain a per-pixel noise standard deviation from the same background model. Sources are then extracted by thresholding the background-subtracted image at $2\times$ the local RMS and retaining connected components of at least 5 pixels. SEP returns a segmentation map together with a structured catalog. We store the segmentation map as a 32-bit integer array in which the value $0$ denotes background and each positive integer is a unique source identifier starting from one. The catalog records the centroid coordinates, the Kron ellipse parameters of semi-major axis, semi-minor axis, and position angle, the pixel count of each source, and the SEP extraction flag. Source identifiers in the catalog are aligned by construction with the labels of the segmentation map, so that selecting all pixels with a given label directly recovers the spatial support of the corresponding catalog entry. This labelling defines the  set used by the detection metric in the main text and by the per-source flux comparison reported in the supplementary tables.

\paragraph{Flux-L1 Metric.}
Using the segmentation map and Kron-ellipse parameters produced above, we form an elliptical aperture mask $M_k$ for each of the $K$  \DIFaddbegin \DIFadd{HST }\DIFaddend sources and integrate the aperture flux on both the prediction and the HR  \DIFaddbegin \DIFadd{HST }\DIFaddend with the SEP sum estimator. The per-image error is the absolute deviation of integrated flux per source, summed over all $K$  \DIFaddbegin \DIFadd{HST }\DIFaddend sources, and aggregating across the test set then yields the macro-averaged Flux-L1 reported in \Cref{tab:experiment}:
\begin{equation}
    \mathrm{Flux\text{-}L1}
    \;=\; \sum_{k=1}^{K}\Bigl|\!\sum_{p\in M_k}\hat{x}_{p} \;-\; \sum_{p\in M_k}x^{\ast}_{p}\Bigr|,
    \label{eq:flux_l1_appendix}
\end{equation}
where $\hat{x}$ denotes the prediction and $x^{\ast}$ the HR HST \DIFadd{image}.

\paragraph{Source Detection on Prediction.}
To quantify hallucination in the predicted images, we run the same detection pipeline independently on the prediction and on the HR  \DIFaddbegin \DIFadd{HST}\DIFaddend , match the two catalogues, and report the per-image macro-averaged precision, recall, and F1 in the detection columns of \Cref{tab:ablation}. We assess source recovery through a three-stage pipeline that compares each prediction against a  \DIFaddbegin \DIFadd{HST }\DIFaddend label mask. For every predicted high-resolution image, we first replace invalid pixels with zeros and estimate a local sky background with SEP on a $64\times64$ grid. After subtracting this background plane we extract sources using a detection threshold of $2\times$ the local background RMS, requiring at least $5$ connected pixels per candidate. Deblending operates over $32$ contour levels with a flux-contrast threshold of $0.02$, splitting overlapping blobs whose sub-peaks rise above this fraction of the parent flux. The reported sub-pixel centroids of the surviving detections form the predicted catalogue. On the  \DIFaddbegin \DIFadd{HST }\DIFaddend side, the integer label mask assigns each positive value to one source. We discard labels covering fewer than $10$ pixels to suppress fragmentary annotations, and the geometric centroid of every remaining label is computed as the unweighted mean of its pixel coordinates, with no flux weighting applied.

Predicted and  \DIFaddbegin \DIFadd{HST }\DIFaddend centroids are then aligned through greedy nearest-first one-to-one matching. We build the pairwise squared-distance matrix between the two sets, retain only candidate pairs whose separation falls within a small tolerance of 10 pixels, and sort the survivors by ascending distance. Walking through this sorted list, the first pair whose prediction and \DIFadd{HST source} are both still unclaimed is registered as a true positive, after which both endpoints are marked used so that any later pair reusing either of them is skipped. Predictions left without a partner contribute false positives, and unmatched  \DIFaddbegin \DIFadd{HST }\DIFaddend sources contribute false negatives. We compute precision, recall, and F1 per image and aggregate over all test samples by averaging the per-image scores, giving every image equal weight regardless of how many sources it contains and preventing dense scenes from masking failures on sparse ones.

\begin{figure}[!tp]
    \centering
    \includegraphics[width=\textwidth]{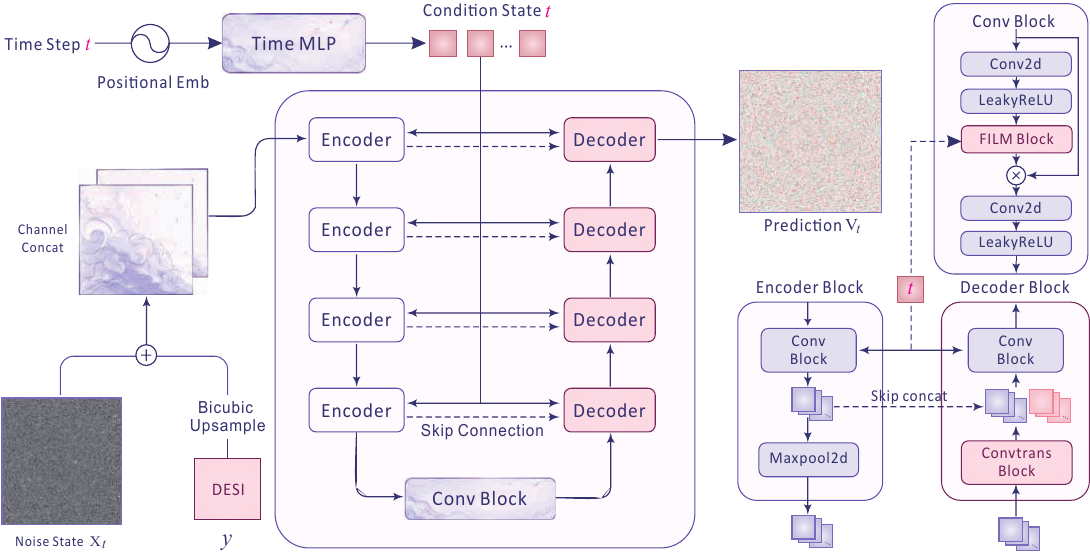}
    \caption{Architecture of the FluxFlow velocity network $v_\theta$. A conditional U-Net takes the channel-wise concatenation of the noisy state $\mathbf{x}_t$ and the bicubic-upsampled DESI observation $\mathbf{y}$, and predicts the OT-CFM velocity $v_\theta$ on the HR grid\DIFaddFL{. T}ime conditioning enters every Conv Block via FiLM modulation.}
    \label{fig:sup_model_arch}
\end{figure}

\paragraph{Model Architecture.}
\label{sup:model}
The velocity network $v_\theta$ is a conditional U-Net operating directly in pixel space, illustrated in \Cref{fig:sup_model_arch}. The DESI observation $\mathbf{y}$ is first lifted to the HR grid via bicubic upsampling and concatenated channel-wise with the noisy state $\mathbf{x}_t$ to form the network input. The encoder is a stack of \emph{Encoder Blocks}, each applying a Conv Block followed by $2{\times}2$ max-pooling that halves the spatial resolution and doubles the channel count from \texttt{base\_dim}{=}48\DIFadd{. A} bottleneck Conv Block sits at the lowest resolution. The symmetric decoder mirrors this pyramid with \emph{Decoder Blocks} that upsample via a transposed-convolution layer (\emph{ConvTrans Block}) and fuse skip connections from the matching encoder level, producing the OT-CFM velocity prediction $v_\theta$ on the HR grid. Each Conv Block is a Conv2d--LeakyReLU--FiLM--Conv2d--LeakyReLU stack in which the FiLM layer~\cite{perez2018film} applies a feature-wise affine modulation conditioned on the time-conditioning state $\tau$, so that the time signal is injected at every resolution. The time step $t \in [0, 1]$ is mapped to $\tau$ by a sinusoidal positional embedding followed by a two-layer MLP with output dimension $256$. The two scale settings ($\times 2$ and $\times 4$) share this architecture exactly and differ only in the spatial resolution of $\mathbf{x}_t$ and the upsampled condition.

\paragraph{Pixel Normalization.}
The DESI and HST cutouts have a wide intrinsic dynamic range in their native flux units (nanomaggies). Before training, we compute global intensity clip thresholds at the $0.01\%$ tail percentiles of all pixels in the training split to suppress extreme values caused by cosmic rays and hot pixels~\cite{liu2026denoising}, which would otherwise dominate the normalization range. One threshold pair is computed for the LR set and another for the HR set, both estimated once on the training split and frozen across all experiments. The clipped pixels are then linearly min--max rescaled to $[0,1]$ using the global per-modality flux extrema. Both training and inference operate in this normalized domain, and PSNR/SSIM are reported in the same domain so that the maximal value is exactly $1$ and the metrics are directly comparable across methods. For the Flux-L1 metric, we denormalize the predictions back to nanomaggies before computing the per-source aperture flux deviation defined in \Cref{eq:flux_l1_appendix}, so that the reported flux errors carry their original physical units.

\begin{figure}[!tp]
    \centering
    \includegraphics[width=\textwidth]{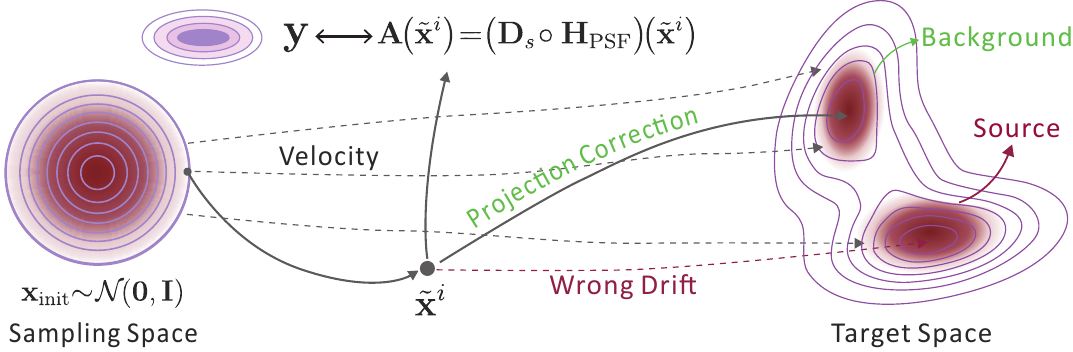}
    \caption{Conceptual illustration of MC-FS. The flow transports $\mathbf{x}_{\mathrm{init}}\!\sim\!\mathcal{N}(\mathbf{0},\mathbf{I})$ toward the target space, but the learned velocity may drift the candidate $\tilde{\mathbf{x}}^{i}$ toward a hallucinated source (red). MC-FS compares $\mathbf{y}$ with $\mathbf{A}(\tilde{\mathbf{x}}^{i})$ (\Cref{sec:mcfs}) to expose this inconsistency and induce a projection correction (green) that pulls the trajectory back toward modes consistent with the observation.}
    \label{fig:sup_mcfs}
\end{figure}

\DIFaddbegin \paragraph{\DIFadd{Adaptive optics and reconstruction.}}
\DIFadd{Adaptive optics corrects wavefront distortions during acquisition~\cite{babcock1953seeing,beckers1993adaptive}; recent work combines controlled adaptive-optics distortions with computational super-resolution~\cite{swanson2025super}. Its performance depends on the instrument, wavelength, guide-star geometry, and field of view. It also requires specialized hardware, real-time control, and demanding observing conditions. FluxFlow instead operates on existing seeing-limited survey coadds without adaptive-optics measurements or hardware control.

}

\DIFaddend \paragraph{Forward-Model PSF Calibration.}
The Gaussian PSF width $\sigma_{\mathrm{PSF}}=2$ used in the MC-FS forward operator defined in \Cref{eq:forward_model} is calibrated directly on the paired DESI--HST training set. For each training pair we forward-project the HST high-resolution image through a candidate Gaussian-blur followed by $s\!\times\! s$ area-downsampling, and compute the validity-masked mean-squared error \DIFaddbegin\DIFadd{over detected source regions }\DIFaddend against the corresponding DESI observation. Sweeping $\sigma$ on a fine grid and taking the value that minimizes this error averaged across the training set yields the data-driven optimum $\sigma^{\ast}=1.92$. We round this to the integer value $\sigma_{\mathrm{PSF}}=2$ used throughout inference.  \DIFaddbegin \DIFadd{Here $\sigma$ }\DIFaddend is  \DIFaddbegin \DIFadd{measured }\DIFaddend in  \DIFaddbegin \DIFadd{HR-grid pixels}\DIFaddend . \DIFaddbegin \DIFadd{This is one dataset-level effective kernel, not a per-exposure or per-source PSF estimate, and it can absorb instrumental response, bandpass differences, and registration error as well as atmospheric blur. It should not be identified directly with a measured atmospheric coherence length.

}\DIFaddend 

\paragraph{Training Details.}
We train the velocity network $v_\theta$ of \Cref{fig:sup_model_arch} with the OT-CFM objective. For each sample we draw $x_0 \sim \mathcal{N}(0, I)$ and $t \sim \mathcal{U}(0, 1)$, form the linear interpolant $x_t = (1-t)\,x_0 + t\,x_1$ with target velocity $u_t = x_1 - x_0$, and regress $v_\theta(x_t, t, c)$ against $u_t$ under the spatially-weighted MSE defined in \Cref{eq:wfm_loss}. On top of the optimizer setting reported in the main text, we apply gradient-norm clipping at $1.0$ and a $1000$-step linear warm-up before cosine decay to zero, and run training in fp32. The EMA model is evaluated every $20$ epochs by Euler-sampling $10$ steps from pure Gaussian noise, reporting MSE/PSNR in the normalized HST domain together with total-flux error in the original flux units.

\paragraph{\DIFadd{Inference cost and profiling protocol.}}
\DIFadd{All samplers share the same UNet and introduce no additional trainable network. FluxFlow has 13.12M trainable parameters at both scales, with 50.04\,MiB of FP32 parameter memory. Pixel-space inference is retained because of the VAE ablation in \Cref{tab:ablation}. We measure batch-size-one inference on one NVIDIA RTX 8000 GPU, using two warm-up images and the same 128-image subset, excluding data loading and synchronizing GPU execution. \Cref{tab:inference_cost} reports the cost of 10 Euler steps. Network FLOPs count 10 network forward passes and exclude the FFT correction.}

\begin{table}[!htbp] \centering \small \caption{\DIFadd{FluxFlow inference cost on one NVIDIA RTX 8000 GPU with batch size one and 10 Euler steps.}} \label{tab:inference_cost} \begin{tabular*}{\linewidth}{@{\extracolsep{\fill}}ccrrr} \toprule \DIFadd{Scale} & \DIFadd{HR shape} & \DIFadd{\shortstack{Time per image\\in ms}} & \DIFadd{\shortstack{Peak memory\\in MiB}} & \DIFadd{Network FLOPs} \\ \midrule \DIFadd{$\times 2$} & \DIFadd{$256\times256$} & \DIFadd{$60.80\pm0.90$} & \DIFadd{241.6} & \DIFadd{604.316 G} \\ \DIFadd{$\times 4$} & \DIFadd{$512\times512$} & \DIFadd{$219.94\pm1.83$} & \DIFadd{441.7} & \DIFadd{2.417 T} \\ \bottomrule \end{tabular*} \end{table} \par \begin{wraptable}[10]{r}{0.42\linewidth} \centering \small \setlength{\tabcolsep}{2pt} \caption{\DIFadd{Inference time in milliseconds per image on one NVIDIA RTX 8000 GPU.}} \label{tab:sampler_runtime} \begin{tabular*}{\linewidth}{@{\extracolsep{\fill}}lrr} \toprule \DIFadd{Method} & \DIFadd{$\times 2$} & \DIFadd{$\times 4$} \\ \midrule \DIFadd{DPS} & \DIFadd{114.41} & \DIFadd{405.01} \\ \DIFadd{$\Pi$GDM} & \DIFadd{153.47} & \DIFadd{441.76} \\ \DIFadd{FlowDPS} & \DIFadd{97.93} & \DIFadd{252.63} \\ \DIFadd{FLOWER} & \DIFadd{89.52} & \DIFadd{254.96} \\ \DIFadd{\textbf{Ours}} & \DIFadd{\textbf{60.80}} & \DIFadd{\textbf{219.94}} \\ \bottomrule \end{tabular*} \end{wraptable} \noindent \DIFadd{\Cref{tab:sampler_runtime} compares the inference times of these samplers under the same timing protocol. At $\times 2$, MC-FS is $1.47\times$ faster than FLOWER, without inference-time automatic differentiation or conjugate-gradient solves.}

\paragraph{Training Time.}
We report the end-to-end training time of FluxFlow and all baselines in \Cref{tab:training_time}, measured on a single NVIDIA A100-80GB GPU under an identical schedule budget. FluxFlow trains in 16h at $\times 2$ scale and 47h at $\times 4$ scale, substantially faster than the Transformer regression baseline HAT which needs over 540 hours and the bridge-based generative baseline AS-Bridge which needs 144 to 360 hours, while matching the photometric and detection metrics reported in \Cref{tab:experiment,tab:ablation}.

\begin{table}[!t]
\centering
\setlength{\tabcolsep}{8pt}
\caption{End-to-end training time on a single NVIDIA A100-80GB GPU at $\times 2$ and $\times 4$ scales.}
\label{tab:training_time}
\begin{tabular}{l|ccccccc|c}
\toprule
Method & SwinIR & HAT & FISR & cGAN & GD-Net & AS-Bridge & & \textbf{Ours} \\
\midrule
$\times 2$ & 94h & 552h & 94h & 8h & 5h & 144h & & \textbf{16h} \\
$\times 4$ & 98h & 549h & 176h & 16h & 184h & 360h & & \textbf{47h} \\
\bottomrule
\end{tabular}
\end{table}

\section{Analysis of Measurement-Consistent Flow Sampling}
\label{sup:mcfs}

This section  \DIFaddbegin \DIFadd{explains }\DIFaddend the  \DIFaddbegin \DIFadd{structure }\DIFaddend of  \DIFaddbegin \DIFadd{MC-FS }\DIFaddend and  \DIFaddbegin \DIFadd{analyzes simplified linear correction models}\DIFaddend . \Cref{sup:mcfs:why}  \DIFaddbegin \DIFadd{describes observable residuals and }\DIFaddend the  \DIFaddbegin \DIFadd{limitations caused }\DIFaddend by the  \DIFaddbegin \DIFadd{null space of the }\DIFaddend forward operator . \Cref{sup:mcfs:wiener_derivation} derives  \DIFaddbegin \DIFadd{a blur-domain }\DIFaddend Wiener  \DIFaddbegin \DIFadd{filter under stationary signal }\DIFaddend and  \DIFaddbegin \DIFadd{noise assumptions}\DIFaddend .  \DIFaddbegin \DIFadd{\Cref{sup:mcfs:adjoint_analysis,sup:mcfs:toy} study blur-only Fourier }\DIFaddend modes  \DIFaddbegin \DIFadd{and }\DIFaddend a  Gaussian-PSF toy  \DIFaddbegin \DIFadd{problem to illustrate finite-iteration conditioning. These calculations do not prove posterior correctness, }\DIFaddend convergence  \DIFaddbegin \DIFadd{of }\DIFaddend the  \DIFaddbegin \DIFadd{nonlinear conditional sampler, or removal }\DIFaddend of \DIFaddbegin \DIFadd{structures in }\DIFaddend the  \DIFaddbegin \DIFadd{measurement null space}\DIFaddend .

\subsection{Forward-Model Consistency for Ill-Posed Image Super-Resolution}
\label{sup:mcfs:why}

Astronomical super-resolution is a severely ill-posed inverse problem. Given the forward model:
\begin{equation}
    \mathbf{y} = \mathbf{A}(\mathbf{x}) + \boldsymbol{n}, \qquad \mathbf{A} = \mathbf{D}_s \circ \mathbf{H}_{\mathrm{PSF}},
\end{equation}
the operator $\mathbf{A}$ is rank-deficient: spatial downsampling $\mathbf{D}_s$ collapses an $H_{\mathrm{hr}} \times W_{\mathrm{hr}}$ image into an $H_{\mathrm{lr}} \times W_{\mathrm{lr}}$ observation, discarding $(s^2-1)/s^2$ of the spatial degrees of freedom, while PSF convolution $\mathbf{H}_{\mathrm{PSF}}$ further attenuates high-frequency content. Consequently, the set of high-resolution images $\mathbf{x}$ consistent with a given DESI observation $\mathbf{y}$ forms a high-dimensional \DIFaddbegin \DIFadd{consistency set that allows for noise. The set is defined as}\DIFaddend
\begin{equation}
    \mathcal{C}(\mathbf{y}) = \left\{ \mathbf{x} \in \mathbb{R}^{H_{\mathrm{hr}} \times W_{\mathrm{hr}}} : \|\mathbf{A}(\mathbf{x}) - \mathbf{y}\|^2 \leq \epsilon \right\},
    \label{eq:consistency_set}
\end{equation}
\DIFadd{In the noiseless limit, exact equality defines an affine set.} Here $\epsilon$ accounts for the noise level $\|\boldsymbol{n}\|^2$. The role of the flow matching prior $v_\theta$ is to select, from this enormous consistency set, samples that also lie on the data manifold of plausible HST-quality images.

 \DIFaddbegin \DIFadd{A residual }\DIFaddend can  \DIFaddbegin \DIFadd{expose structures whose forward projection disagrees with the observation }\DIFaddend under  \DIFaddbegin \DIFadd{the chosen model}\DIFaddend .  \DIFaddbegin \DIFadd{However}\DIFaddend ,  \DIFaddbegin \DIFadd{rank deficiency permits distinct HR structures }\DIFaddend with the  \DIFaddbegin \DIFadd{same LR projection, so consistency is not sufficient to exclude hallucinations. Noise }\DIFaddend and  \DIFaddbegin \DIFadd{forward-model mismatch can also cause genuine structure to have a nonzero residual}\DIFaddend .

Forward-model consistency correction directly addresses this failure mode by projecting the trajectory back toward $\mathcal{C}(\mathbf{y})$ at every integration step. Concretely, each ODE update is followed by a gradient step on the data-fidelity term:
\begin{equation}
    \mathcal{D}(\mathbf{x}) = \tfrac{1}{2}\|\mathbf{A}(\mathbf{x}) - \mathbf{y}\|^2,
    \label{eq:data_fidelity}
\end{equation}
whose gradient  is  \DIFaddbegin \DIFadd{$\nabla_{\mathbf{x}}\mathcal{D}(\mathbf{x})=\mathbf{A}^{\!\top}(\mathbf{A}(\mathbf{x})-\mathbf{y})$}\DIFaddend .  The \DIFaddbegin \DIFadd{Wiener variant replaces the PSF adjoint in this correction with the Wiener kernel rather than applying the exact gradient}\DIFaddend . \DIFaddbegin \DIFadd{Both depend on the approximate operator, and neither can resolve its null space without information from the learned conditional model.

}\DIFaddend 

\subsection{Derivation of the Wiener Correction Kernel}
\label{sup:mcfs:wiener_derivation}

We derive the Wiener correction kernel $W(\mathbf{k})$ of \Cref{eq:wiener_kernel} from the minimum mean-square error (MMSE) principle. Consider the deconvolution sub-problem at integration step $i$. Given an upsampled residual $\mathbf{r}^{(i)}_{\uparrow} = \mathbf{D}_s^{\!\top}(\mathbf{r}^{(i)})$ on the high-resolution grid, we seek a high-resolution correction signal $\boldsymbol{\rho}^{(i)}$ such that $\mathbf{H}_{\mathrm{PSF}}(\boldsymbol{\rho}^{(i)}) \approx \mathbf{r}^{(i)}_{\uparrow}$ in the presence of measurement noise. This order of operations matches the strict-adjoint update in \Cref{eq:wiener_correction}, with $H(\mathbf{k})$ and the residual it acts on both defined on the HR grid.

Working in the Fourier domain, let $H(\mathbf{k}) = \mathcal{F}[h]$ denote the optical transfer function of the PSF kernel $h$, and assume the correction signal admits a wide-sense stationary spectral representation with power spectral density $\Phi_\rho(\mathbf{k})$, while the noise has power $\sigma_n^2$. The Wiener filter is the linear estimator $\boldsymbol{\rho}^{(i)} = w * \mathbf{r}^{(i)}_{\uparrow}$ that minimizes the mean-square error:
\begin{equation}
    \mathrm{MSE}(w) = \mathbb{E}\!\left[\,\|\boldsymbol{\rho}^{(i)} - w * \mathbf{r}^{(i)}_{\uparrow}\|^2\,\right].
    \label{eq:wiener_mse}
\end{equation}
Setting the variation $\partial \mathrm{MSE}/\partial w = 0$ and solving in the Fourier domain~\cite{wiener1949extrapolation} yields the optimal filter:
\begin{equation}
    W(\mathbf{k}) \;=\; \frac{H^*(\mathbf{k})\, \Phi_\rho(\mathbf{k})}{|H(\mathbf{k})|^2\, \Phi_\rho(\mathbf{k}) + \sigma_n^2} \;=\; \frac{H^*(\mathbf{k})}{|H(\mathbf{k})|^2 + \lambda_{\mathrm{SNR}}^{-1}(\mathbf{k})},
    \label{eq:wiener_full}
\end{equation}
where $\lambda_{\mathrm{SNR}}(\mathbf{k}) = \Phi_\rho(\mathbf{k}) / \sigma_n^2$ is the frequency-dependent signal-to-noise ratio. In practice we replace $\lambda_{\mathrm{SNR}}(\mathbf{k})$ by a single global constant $\lambda_{\mathrm{SNR}}=50$, recovering the form in \Cref{eq:wiener_kernel}. This simplification avoids estimating the full power spectrum of the residual and treats the regularization parameter as a tunable hyperparameter.

\paragraph{Limiting Behaviors.} Two extreme regimes provide intuition. As $\lambda_{\mathrm{SNR}} \to \infty$, the noise term vanishes and \Cref{eq:wiener_full} reduces to the exact inverse filter:
\begin{equation}
    W(\mathbf{k}) \;\longrightarrow\; \frac{H^*(\mathbf{k})}{|H(\mathbf{k})|^2} \;=\; \frac{1}{H(\mathbf{k})},
\end{equation}
This regime maximally amplifies high frequencies but is unstable wherever $H(\mathbf{k}) \approx 0$. Conversely, as $\lambda_{\mathrm{SNR}} \to 0$, the regularization dominates the denominator and the filter magnitude vanishes
\begin{equation}
    W(\mathbf{k}) \;\longrightarrow\; \lambda_{\mathrm{SNR}}\,H^*(\mathbf{k}) \;\to\; 0,
\end{equation}
whose frequency-response shape remains proportional to the matched filter $H^*(\mathbf{k})$, equivalent up to scaling to the adjoint $\mathbf{H}_{\mathrm{PSF}}^{\!\top}$ of \Cref{eq:adjoint_correction}. The Wiener kernel thus interpolates continuously between the two extremes, and $\lambda_{\mathrm{SNR}}$ controls how aggressively the deconvolution restores high-frequency content suppressed by the PSF.

\DIFaddbegin \DIFadd{The observed robustness applies to the calibrated Gaussian setting.}\DIFaddend \subsection{Spectral Analysis of Adjoint Backprojection}
\label{sup:mcfs:adjoint_analysis}

The qualitative claim in \Cref{sec:mcfs} that naive adjoint back-projection accumulates PSF-scale smoothing across integration steps admits a precise frequency-domain characterization. For clarity we ignore the downsampling operator $\mathbf{D}_s$ in this analysis and focus on the PSF component $\mathbf{H}_{\mathrm{PSF}}$, which is responsible for the resolution loss. The downsampling step contributes additional aliasing but does not alter the qualitative conclusion.

 \DIFaddbegin \DIFadd{For this illustrative calculation, consider }\DIFaddend a \DIFaddbegin \DIFadd{blur-only, periodic-convolution operator, omit the learned velocity forcing, and analyze a }\DIFaddend single Fourier mode at spatial frequency $\mathbf{k}$. \DIFaddbegin \DIFadd{The full operator $\mathbf{A}=\mathbf{D}_s\mathbf{H}$ couples frequencies through decimation, so the following diagonal calculation is not an exact analysis of the implemented SR sampler. }\DIFaddend After the velocity update of \Cref{eq:euler_candidate}, the candidate state has Fourier amplitude $\tilde{X}^{(i+1)}(\mathbf{k})$. The adjoint correction of \Cref{eq:adjoint_correction} can be written in the Fourier domain as:
\begin{equation}
    X^{(i+1)}(\mathbf{k}) \;=\; \tilde{X}^{(i+1)}(\mathbf{k}) - \eta_i\, |H(\mathbf{k})|^2\, \tilde{X}^{(i+1)}(\mathbf{k}) + \eta_i\, H^*(\mathbf{k})\, Y(\mathbf{k}),
    \label{eq:adjoint_fourier}
\end{equation}
since  \DIFaddbegin \DIFadd{$\mathbf{H}^{\!\top}\mathbf{H}$ }\DIFaddend acts in  \DIFaddbegin \DIFadd{this blur-only calculation }\DIFaddend as multiplication by $|H(\mathbf{k})|^2$. Restricting attention to the homogeneous part gives the per-step transfer function:
\begin{equation}
    G_i(\mathbf{k}) \;=\; 1 - \eta_i\, |H(\mathbf{k})|^2,
\end{equation}
and after $N$ integration steps, the cumulative attenuation factor on mode $\mathbf{k}$ is:
\begin{equation}
    G_{\mathrm{total}}(\mathbf{k}) \;=\; \prod_{i=0}^{N-1} \left(1 - \eta_i\, |H(\mathbf{k})|^2\right).
    \label{eq:cumulative_attenuation}
\end{equation}
 \DIFaddbegin \DIFadd{For modes with $H(\mathbf{k})\neq0$}\DIFaddend ,  \DIFaddbegin \DIFadd{a constant-step blur-only iteration }\DIFaddend converges to  $H^*(\mathbf{k})Y(\mathbf{k})/|H(\mathbf{k})|^2$ \DIFaddbegin \DIFadd{when $|1-\eta|H(\mathbf{k})|^2|<1$}\DIFaddend .  \DIFaddbegin \DIFadd{The inverse }\DIFaddend filter amplifies  \DIFaddbegin \DIFadd{observation-noise power by }\DIFaddend $|H(\mathbf{k})|^{-2}$ \DIFaddbegin \DIFadd{. Convergence alone therefore does not imply denoising}\DIFaddend .  \DIFaddbegin \DIFadd{At frequencies with $H(\mathbf{k})=0$}\DIFaddend ,  \DIFaddbegin \DIFadd{this }\DIFaddend correction  \DIFaddbegin \DIFadd{leaves }\DIFaddend the  \DIFaddbegin \DIFadd{mode unchanged}\DIFaddend .  \DIFaddbegin \DIFadd{Finite iteration counts }\DIFaddend and  \DIFaddbegin \DIFadd{small $|H(\mathbf{k})|$ can slow the recovery of }\DIFaddend high-frequency  \DIFaddbegin \DIFadd{components in this simplified model}\DIFaddend .  \DIFaddbegin \DIFadd{These observations motivate }\DIFaddend the  \DIFaddbegin \DIFadd{toy comparison }\DIFaddend in  \DIFaddbegin \DIFadd{\Cref{sup:mcfs:toy}; they do not establish PSF broadening or }\DIFaddend convergence of  the  \DIFaddbegin \DIFadd{learned}\DIFaddend ,  \DIFaddbegin \DIFadd{downsampling-based sampler}\DIFaddend .

The  \DIFaddbegin \DIFadd{Wiener correction has bounded per-step gain in }\DIFaddend this  \DIFaddbegin \DIFadd{simplified model}\DIFaddend . Substituting $W(\mathbf{k})$ into the correction term replaces the multiplier $H^*(\mathbf{k})$ with $H^*(\mathbf{k}) / (|H(\mathbf{k})|^2 + \lambda_{\mathrm{SNR}}^{-1})$, which is uniformly bounded by $\tfrac{1}{2}\lambda_{\mathrm{SNR}}^{1/2}$. The Wiener kernel still tracks the inverse filter $1/H(\mathbf{k})$ deep within the passband, but stays finite in the transition band and  \DIFaddbegin \DIFadd{bounds }\DIFaddend the  \DIFaddbegin \DIFadd{gain of each correction}\DIFaddend .  \DIFaddbegin \DIFadd{This per-step bound does not guarantee asymptotic noise suppression or convergence of }\DIFaddend the  \DIFaddbegin \DIFadd{full nonlinear sampler; }\DIFaddend the  \DIFaddbegin \DIFadd{tradeoff with a finite number of iterations is examined in the toy model below}\DIFaddend .

\begin{figure}[!tp]
    \centering
    \includegraphics[width=\textwidth]{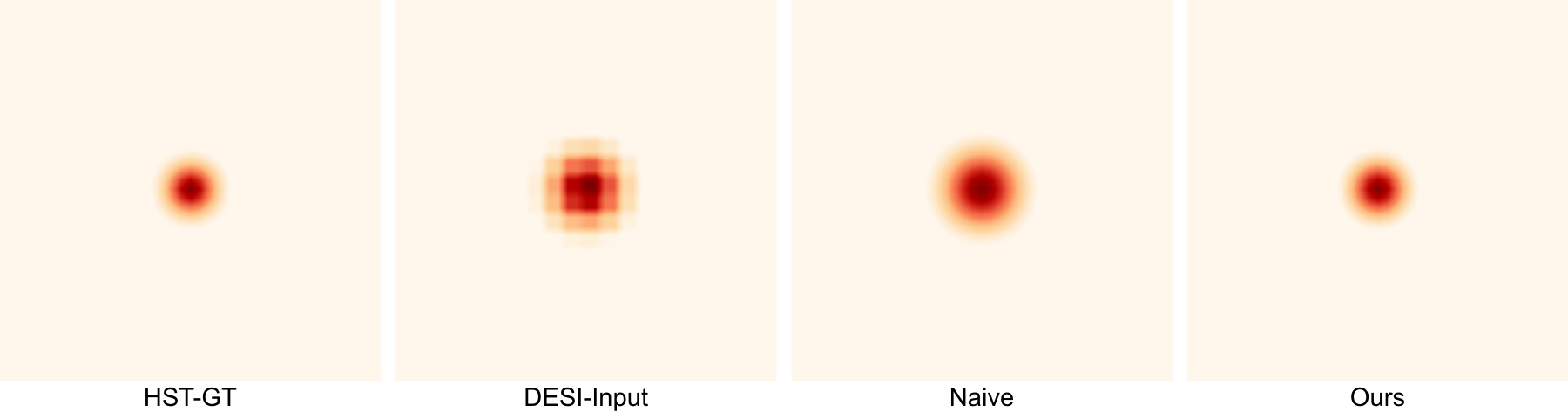}
    \caption{Toy reconstruction at $\sigma_n=0$ with $N=10$ Euler steps. The four panels show, from left to right, the HST-quality truth, the DESI observation, the naive adjoint back-projection, and the Wiener-corrected update. The naive recovery has fitted width $\sigma_{\mathrm{fit}}\approx 2.75$, visibly broadened relative to the truth despite the observation being noise-free. The Wiener-corrected update at the same $N$ has $\sigma_{\mathrm{fit}}\approx 2.03$, restored to the truth.}
    \label{fig:sup_toy_psf}
\end{figure}

\subsection{Closed-Form Toy Model}
\label{sup:mcfs:toy}

The qualitative argument above identifies two distinct failure modes of unregularized adjoint back-projection. The first is \emph{slow per-mode convergence} of high spatial frequencies, since the Gaussian PSF makes the per-step contraction rate $\eta|H(\mathbf{k})|^2$ exponentially small at high $|\mathbf{k}|$. The second is \emph{noise amplification} of the unregularized inverse filter at the iteration fixed point. These two failure modes operate on different timescales. The first dominates at finite iteration count $N$, while the second is a $N\!\to\!\infty$ pathology. Because our sampling pipeline runs for a finite $N$, the first is the dominant cause of the apparent PSF widening reported in our ablation. \DIFaddbegin \DIFadd{The following closed-form analysis shows that the Wiener kernel accelerates the convergence of individual Fourier modes within the passband for a finite number of iterations.}\DIFaddend It does not change the asymptotic fixed point of the iteration, since the Wiener kernel is strictly positive over the passband and any fixed point of $X^{(i+1)}=X^{(i)}-\eta W(\mathbf{k})(H(\mathbf{k})X^{(i)}-Y(\mathbf{k}))$ still satisfies $X^{(\infty)}(\mathbf{k})=Y(\mathbf{k})/H(\mathbf{k})$, but instead allows the iteration to be safely truncated at finite $N$ before noise amplification substantially affects the reconstruction.

\paragraph{Setup.}
We take the PSF to be a normalized $2$D Gaussian of width $\sigma_h$, the ground-truth PSF used to generate the simulated observation in the following toy analysis, so that
\begin{equation}
    h(\mathbf{r}) \;=\; \frac{1}{2\pi\sigma_h^2}\exp\Bigl(-\tfrac{|\mathbf{r}|^2}{2\sigma_h^2}\Bigr),
    \qquad
    H(\mathbf{k}) \;=\; \exp\Bigl(-\tfrac{1}{2}\sigma_h^2|\mathbf{k}|^2\Bigr).
\end{equation}
We probe the recovery operator with a single point source $x_{\mathrm{src}}(\mathbf{r}) = \Psi\,\delta(\mathbf{r})$ of amplitude $\Psi$, which acts as a clean impulse and isolates the recovery dynamics from any source structure. The observation noise is spatially white with power $\sigma_n^2$, so the forward observation in the Fourier domain becomes:
\begin{equation}
    Y(\mathbf{k}) \;=\; H(\mathbf{k})\,\Psi + \Gamma(\mathbf{k}),
    \qquad
    \langle\Gamma(\mathbf{k})\,\overline{\Gamma(\mathbf{k}')}\rangle \;=\; \sigma_n^2\,\delta(\mathbf{k}-\mathbf{k}').
\end{equation}
We restrict the analysis to the band $|\mathbf{k}|\leq k_{\max}$ supported by the LR observation. For $s\!\times\!s$ area downsampling we have $k_{\max}\approx \pi/s$ in HR frequency units. In contrast to a fixed-point analysis, our pipeline iterates \Cref{eq:adjoint_correction} for a finite number $N=10$ of Euler steps, with constant step size $\eta$ adopted purely as an analytical simplification for the closed-form toy-model derivation that follows. The actual sampling pipeline of \Cref{alg:mcfm} uses the linear-decay schedule $\eta_i=\eta_0(1-i/N)$ described in \Cref{sec:mcfm}, and \DIFaddbegin \DIFadd{extending }\DIFaddend the  \DIFaddbegin \DIFadd{closed-form expressions to }\DIFaddend that schedule \DIFaddbegin \DIFadd{requires the corresponding products of step-dependent factors}\DIFaddend . The closed-form analysis below is therefore parametric in $N$.

\paragraph{Per-mode Closed Form for Finite $N$.}
The adjoint update of \Cref{eq:adjoint_correction} acts independently on each Fourier mode. Writing the per-mode recursion $X^{(i+1)}(\mathbf{k}) = (1-\eta|H(\mathbf{k})|^2)X^{(i)}(\mathbf{k}) + \eta H^*(\mathbf{k})Y(\mathbf{k})$ with $X^{(0)}=0$ and summing the resulting geometric series yields the per-mode closed form:
\begin{equation}
    X^{(N)}_{\mathrm{naive}}(\mathbf{k}) \;=\; g_N^{\mathrm{naive}}(\mathbf{k})\,\Bigl[\,\Psi \,+\, \frac{\Gamma(\mathbf{k})}{H(\mathbf{k})}\,\Bigr],
    \qquad
    g_N^{\mathrm{naive}}(\mathbf{k}) \;\equiv\; 1-\bigl(1-\eta|H(\mathbf{k})|^2\bigr)^{N}.
    \label{eq:toy_finite_T_naive}
\end{equation}
The factor $g_N^{\mathrm{naive}}(\mathbf{k})$ is the per-mode signal-recovery factor, gating both the signal term $\Psi$ and the noise term $\Gamma/H$, and approaches unity only as $N\!\to\!\infty$. Per step, mode $\mathbf{k}$ contracts toward its fixed point at rate $\eta|H(\mathbf{k})|^2$, which is exponentially small at high $|\mathbf{k}|$ for a Gaussian PSF. For our setting $\sigma_h=2$ and $k_{\max}=\pi/4$, $|H(k_{\max})|^2 = \exp(-\sigma_h^2 k_{\max}^2) = \exp(-\pi^2/4)\approx 0.085$, more than an order of magnitude below the deep-passband value $|H(\mathbf{0})|^2=1$, so the per-step contraction at the band edge is correspondingly slow. At any practical finite $N$ the band edge therefore remains underconverged with $g_N^{\mathrm{naive}}(k_{\max})$ bounded well below unity, while the deep passband saturates at $g_N^{\mathrm{naive}}\!\approx\!1$ within a few iterations.

\begin{figure}[!tp]
    \centering
    \begin{subfigure}{0.325\textwidth}
        \centering
        \includegraphics[width=\linewidth]{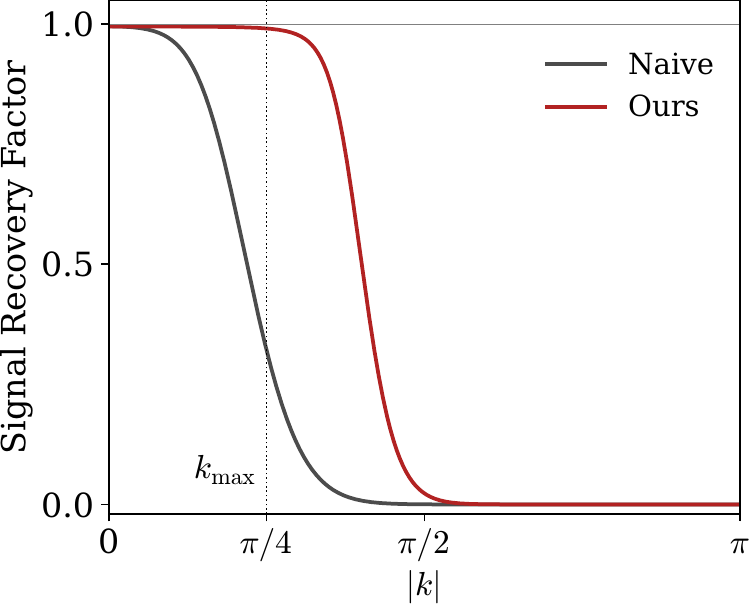}
        \caption{Per-mode signal recovery $g_N(\mathbf{k})$ at $N=10$. Naive drops sharply within the LR band $|\mathbf{k}|\leq k_{\max}$, recovering only $\sim\!40\%$ at the band edge, whereas ours persists.}
        \label{fig:sup_toy_recovery}
    \end{subfigure}
    \hfill
    \begin{subfigure}{0.325\textwidth}
        \centering
        \includegraphics[width=\linewidth]{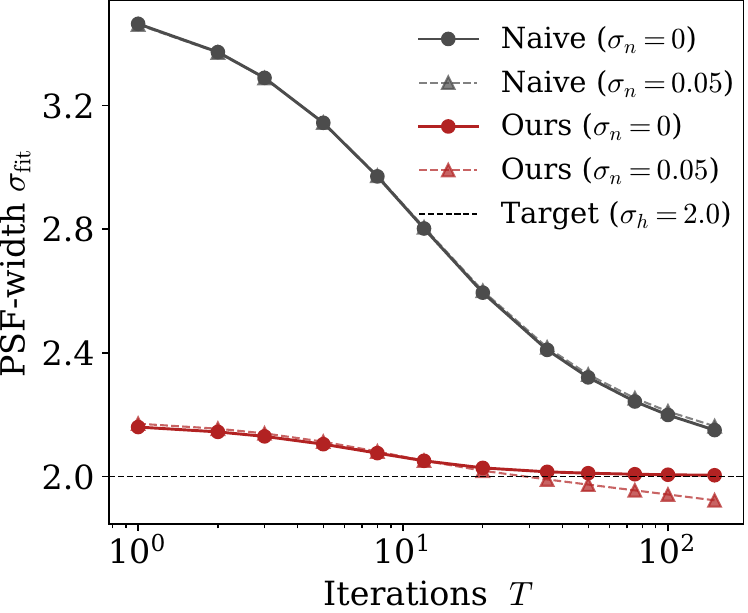}
        \caption{Fitted $\sigma_{\mathrm{fit}}$ vs.\ iteration $N$ at $\sigma_n=0$ and $\sigma_n=0.05$. The two noise levels overlap, showing that the widening is a low-pass effect rather than noise amplification.}
        \label{fig:sup_toy_sigma}
    \end{subfigure}
    \hfill
    \begin{subfigure}{0.325\textwidth}
        \centering
        \includegraphics[width=\linewidth]{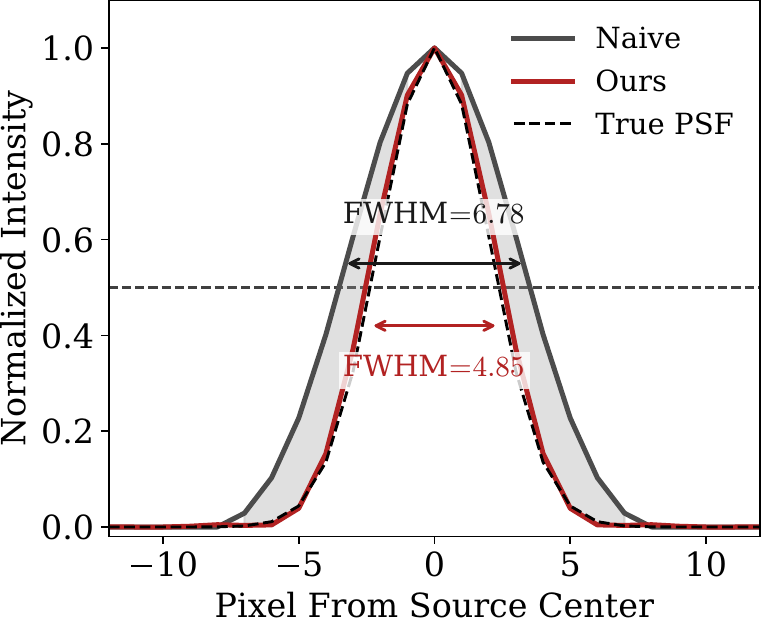}
        \caption{$1$D radial profile cuts at $N=10$. Naive is visibly broader than the true PSF, with FWHM enlarged from $4.71$ to $6.78$ HR pixels, while Wiener has FWHM $\approx 4.85$.}
        \label{fig:sup_toy_intensity}
    \end{subfigure}
    \caption{Numerical verification of the closed-form analysis at the same parameters as \Cref{fig:sup_toy_psf}. Panels~(a)--(c) directly visualize \Cref{eq:toy_finite_T_naive,eq:toy_widening_low_pass,eq:toy_gT_wiener}.}
    \label{fig:sup_toy_panels}
\end{figure}

\paragraph{Apparent PSF Widening at Finite $N$.}
\Cref{eq:toy_finite_T_naive} attenuates both signal and noise by the same factor $g_N^{\mathrm{naive}}$. The deterministic signal term is therefore a low-passed version of the true delta whenever $g_N^{\mathrm{naive}}<1$ at the band edge. Fitting a Gaussian profile to the recovered impulse returns a width:
\begin{equation}
    \sigma_{\mathrm{fit}}^{\mathrm{naive}} \;>\; \sigma_{\infty} \qquad \text{for any finite } N,
    \label{eq:toy_widening_low_pass}
\end{equation}
where $\sigma_{\infty}$ denotes the band-limited resolution floor set by $k_{\max}$, i.e.\ the spatial width of the inverse Fourier transform of the indicator of $|\mathbf{k}|\leq k_{\max}$ (a 2D sinc-like profile), which is independent of the forward blur $\sigma_h$. As $N\!\to\!\infty$, $g_N^{\mathrm{naive}}(\mathbf{k})\to 1$ within the LR-supported band and $\sigma_{\mathrm{fit}}\!\to\!\sigma_{\infty}$ rather than to $\sigma_h$. The apparent numerical agreement with $\sigma_h\approx 2$ in \Cref{fig:sup_toy_sigma} reflects the specific choice of $\sigma_h$ and $k_{\max}$ used in this toy setup, not a genuine convergence to the optical blur scale. \Cref{fig:sup_toy_psf} shows the qualitative outcome, with the naive recovery visibly broader than the truth even at the noise-free condition of $\sigma_n=0$.

\paragraph{Wiener Correction.}
Substituting the Wiener kernel $W(\mathbf{k}) = H^*(\mathbf{k})/(|H(\mathbf{k})|^2+\lambda_{\mathrm{SNR}}^{-1})$ for $H^*(\mathbf{k})$ in the correction replaces the per-mode contraction rate $\eta|H(\mathbf{k})|^2$ with
\begin{equation}
    \eta\,\frac{|H(\mathbf{k})|^2}{|H(\mathbf{k})|^2 + \lambda_{\mathrm{SNR}}^{-1}}.
    \label{eq:toy_wiener_rate}
\end{equation}
Within the LR-supported band, $|H(\mathbf{k})|^2\!\gg\!\lambda_{\mathrm{SNR}}^{-1}$ at our default $\lambda_{\mathrm{SNR}}\!=\!50$, so this ratio is close to unity rather than to $|H(\mathbf{k})|^2$. The per-step contraction rate is therefore approximately $\eta$ uniformly across the band, an order of magnitude faster than the naive update at the band edge. The corresponding signal-recovery factor
\begin{equation}
    g_N^{\mathrm{Wiener}}(\mathbf{k}) \;=\; 1-\Bigl(1-\eta\,\tfrac{|H(\mathbf{k})|^2}{|H(\mathbf{k})|^2+\lambda_{\mathrm{SNR}}^{-1}}\Bigr)^{N}
    \label{eq:toy_gT_wiener}
\end{equation}
saturates to unity across the entire band $|\mathbf{k}|\leq k_{\max}$ within a small $N$ (\Cref{fig:sup_toy_recovery}). The Wiener-corrected reconstruction is therefore essentially band-converged at modest $N$, and \Cref{eq:toy_widening_low_pass} no longer applies. \Cref{fig:sup_toy_sigma} shows $\sigma_{\mathrm{fit}}^{\mathrm{Wiener}}\!\to\!\sigma_h$ within a few iterations, while the corresponding $1$D profile cut (\Cref{fig:sup_toy_intensity}) tracks the true PSF within line width.

\paragraph{Fixed-point Noise Amplification.}
For completeness, we record the $N\!\to\!\infty$ behaviour. In that limit, \Cref{eq:toy_finite_T_naive} collapses to $X^{(\infty)}(\mathbf{k}) = \Psi + \Gamma(\mathbf{k})/H(\mathbf{k})$, whose noise spectrum $\Phi_{\mathrm{rec}}(\mathbf{k}) = \sigma_n^2\,\exp(\sigma_h^2|\mathbf{k}|^2)$ diverges exponentially at high $|\mathbf{k}|$. Bandlimiting and integrating with the continuous angular-frequency Parseval factor $1/(2\pi)^2$ yields the closed-form per-pixel variance:
\begin{equation}
    \sigma_{\mathrm{rec}}^2 \;=\; \frac{\sigma_n^2}{4\pi\,\sigma_h^2}\bigl(\exp(\sigma_h^2 k_{\max}^2) - 1\bigr),
    \label{eq:toy_variance}
\end{equation}
which evaluates to roughly $0.21\,\sigma_n^2$ at the DESI--HST setting ($\sigma_h=2$, $k_{\max}=\pi/4$). While modest in absolute terms, this corresponds to a substantial amplification of the noise power that lives within the narrow LR-supported band $|\mathbf{k}|\leq k_{\max}$, and the prefactor $\exp(\sigma_h^2 k_{\max}^2)-1$ grows exponentially as $k_{\max}$ approaches the Nyquist limit, so this term remains a real concern for any iteration run to convergence. The practical $N$ used in our pipeline keeps the recovery short of this fixed point. In our regime the dominant pathology is therefore the finite-$N$ low-pass effect of \Cref{eq:toy_widening_low_pass} rather than the asymptotic noise amplification of \Cref{eq:toy_variance}.

\section{Additional Quantitative Results}
\label{sec:sup:experiments}

We further evaluate FluxFlow on two existing astronomical super-resolution benchmarks. The AstroSR dataset~\cite{miao2024astrosr} pairs galaxy cutouts across two different ground-based optical surveys, drawing low-resolution images from the Sloan Digital Sky Survey (SDSS)~\cite{york2000sloan} and high-resolution targets from the Subaru Hyper Suprime-Cam survey (HSC)~\cite{aihara2018hyper}, and is distributed as three-channel RGB composites rather than flux-calibrated single-band photometry. The STAR dataset~\cite{wu2025star} consists of space-based HST star-field cutouts in the F814W band~\cite{sirianni2005photometric}, whose low-resolution counterparts are produced by synthetic downsampling of the same HST images. Neither benchmark exposes the severe atmospheric PSF broadening that characterizes genuine ground-to-space observation pairs, as also summarized in \Cref{tab:astro_sr_datasets}, nor a calibrated atmospheric PSF kernel for the LR to HR transformation, so the LR to HR gap is much milder than the seeing-limited regime our full pipeline targets. We therefore set the PSF kernel $h$ in the forward model of \Cref{eq:forward_model} to the identity $\mathbf{H}_{\mathrm{PSF}}\!=\!\mathbf{I}$ for these experiments, so that the inference-time MC-FS correction reduces to a residual back-projection through the area-downsampling adjoint $\mathbf{D}_s^{\!\top}$ alone and no Wiener deconvolution is applied. In addition, neither benchmark provides per-pixel inverse-variance weight maps or co-registered HR catalogs from which to derive source masks, so we omit both the inverse-variance term $\widetilde{\mathbf{W}}$ and the source-importance term $\mathbf{S}$ from the WFM objective in ~\Cref{eq:wfm_loss}, recovering the unweighted CFM loss of \Cref{eq:cfm_loss}. All other architectural and optimization choices follow \Cref{sec:experiment}.

\begin{table}[!tp]
\centering
\small
\setlength{\tabcolsep}{7pt}
\caption{Quantitative comparison at $\times2$ and $\times4$ on STAR dataset~\cite{wu2025star}.}
\label{tab:star_results}
\makebox[\textwidth][c]{%
\begin{tabular}{ll@{\hspace{0.5em}}|@{\hspace{0.5em}}cccccccc}
\toprule
Scale & Metric & Bicubic & EDSR & SRGAN & RCAN & SwinIR & HAT & FISR & \textbf{Ours} \\
\midrule
& PSNR$\uparrow$ & 28.71 & 33.50 & 32.25 & 33.89 & 34.05 & 34.38\thd & 34.40\sed & 34.66\fst \\
$\times2$ & SSIM$\uparrow$ & 0.6945 & 0.7729 & 0.7501 & 0.7808 & 0.7812 & 0.7836\thd & 0.7859\sed & 0.7918\fst \\
& Flux-L1$\downarrow$ & 125.4 & 48.78 & 51.35 & 43.44 & 42.41 & 41.27\sed & 41.05\fst & 41.84\thd \\
\midrule
& PSNR$\uparrow$ & 26.21 & 30.73 & 30.33 & 32.13 & 32.24 & 32.67\fst & 32.51\sed & 32.44\thd \\
$\times4$ & SSIM$\uparrow$ & 0.5993 & 0.6549 & 0.6522 & 0.6838 & 0.6881\thd & 0.6936\fst & 0.6891\sed & 0.6867 \\
& Flux-L1$\downarrow$ & 239.2 & 67.47 & 74.72 & 60.52 & 55.35\thd & 48.49\fst & 50.94\sed & 57.20 \\
\bottomrule
\end{tabular}%
}
\end{table}

\begin{table}[!tp]
\centering
\small
\caption{Quantitative comparison at $\times2$ on AstroSR dataset~\cite{miao2024astrosr}.}
\label{tab:astrosr_results}
\setlength{\tabcolsep}{5.5pt}
\makebox[\textwidth][c]{%
\begin{tabular}{ll@{\hspace{0.5em}}|@{\hspace{0.5em}}ccccccccc}
\toprule
Scale & Metric & Bicubic & EDSR & SRGAN & RCAN & ENLCA & SwinIR & HAT & FISR & \textbf{Ours} \\
\midrule
& PSNR$\uparrow$ & 17.75 & 23.71 & 23.46 & 23.72 & 23.70 & 23.74 & 23.81\thd & 23.84\sed & 23.87\fst \\
$\times2$ & SSIM$\uparrow$ & 0.171 & 0.350 & 0.348 & 0.351 & 0.350 & 0.352\thd & 0.352\thd & 0.354\fst & 0.352\thd \\
& Flux-L1$\downarrow$ & 1916 & 213.6 & 165.9\fst & 226.3 & 216.1 & 223.1 & 193.3\thd & 192.5\sed & 198.7 \\
\bottomrule
\end{tabular}%
}
\end{table}

The Flux-L1 metric in this section is defined consistently with \Cref{tab:experiment} in the main text. We deliberately report the Flux-L1, defined as \Cref{eq:flux_l1_appendix}, rather than the per-source averaged Flux Error (FE) defined in STAR~\cite{wu2025star} for two reasons. First, our DESI to HST mapping operates in calibrated nanomaggie units whose per-pixel values are extremely small. The sum-based formulation therefore preserves a meaningful range and remains consistent with the definition adopted in \Cref{tab:experiment}, while dividing by the number of recovered sources would compress the metric into a numerically uninformative regime. Second, individual sources span a very wide range of extent, brightness, and morphology, from compact point-like profiles to extended galaxies. Under such heterogeneity, a per-source mean weights a faint compact detection and a bright extended structure equally and obscures where the photometric error actually concentrates. Aggregating the residual as a sum over all pixels instead reflects the cumulative photometric fidelity of the field as a whole.

We further note an important difference in the evaluation protocol. The STAR benchmark~\cite{wu2025star} adopts a per-image min-max normalization that depends on the ground-truth extrema of each test image, which is not reproducible at inference time when the ground truth is unavailable. To keep the evaluation usable in a deployment setting and consistent with the rest of the paper, we instead adopt the same global normalization protocol used for our DESI--HST experiments described in Appendix~\ref{sup:implementation}, namely a $0.01\%$ tail clip followed by global min-max rescaling fitted on the training split, and we report PSNR and SSIM in the normalized domain and Flux-L1 in the denormalized native units. For the AstroSR benchmark~\cite{miao2024astrosr}, the data are distributed as 8-bit RGB composites with consistent scaling across SDSS and HSC, so no further normalization is required and we evaluate directly on the released values.

\DIFaddbegin \DIFadd{We evaluate the FluxFlow network on STAR and AstroSR, with results reported in \Cref{tab:star_results,tab:astrosr_results}. As specified above}\DIFaddend , \DIFaddbegin \DIFadd{these configurations use an identity PSF }\DIFaddend and  \DIFaddbegin \DIFadd{omit Wiener deconvolution}\DIFaddend ,  \DIFaddbegin \DIFadd{inverse-variance weighting, and source weighting}\DIFaddend . On  STAR  at $\times 2$,  the  \DIFaddbegin \DIFadd{reported configuration leads in }\DIFaddend PSNR and SSIM , \DIFaddbegin \DIFadd{while HAT }\DIFaddend and  \DIFaddbegin \DIFadd{FISR have lower Flux-L1. }\DIFaddend It  \DIFaddbegin \DIFadd{does not lead }\DIFaddend the  \DIFaddbegin \DIFadd{$\times 4$ table}\DIFaddend . On  AstroSR  at $\times 2$,  \DIFaddbegin \DIFadd{it leads in }\DIFaddend PSNR  \DIFaddbegin \DIFadd{but not }\DIFaddend SSIM \DIFaddbegin \DIFadd{or Flux-L1}\DIFaddend .  AstroSR \DIFaddbegin \DIFadd{'s }\DIFaddend RGB composites  \DIFaddbegin \DIFadd{also limit }\DIFaddend the  \DIFaddbegin \DIFadd{interpretation of }\DIFaddend Flux-L1  \DIFaddbegin \DIFadd{as }\DIFaddend physical  \DIFaddbegin \DIFadd{photometry. These metric-dependent results support transfer of the conditional backbone to synthetic space-to-space }\DIFaddend and  \DIFaddbegin \DIFadd{cross-instrument ground-to-ground settings}\DIFaddend .  \DIFaddbegin \DIFadd{They do not establish zero-shot transfer of }\DIFaddend the  \DIFaddbegin \DIFadd{full method or superiority across }\DIFaddend degradation  \DIFaddbegin \DIFadd{types. The main results concern real ground-to-space reconstruction under atmospheric seeing. Comparing these different datasets alone does not isolate }\DIFaddend the  \DIFaddbegin \DIFadd{cause }\DIFaddend of the  \DIFaddbegin \DIFadd{performance differences}\DIFaddend .

\section{Additional Ablation Study}
\label{sec:sup:ablation}

This section presents additional ablations beyond the main study of \Cref{tab:ablation}. We first examine two alternatives to pixel-space OT-CFM, a latent-space flow-matching variant in a frozen SD-VAE~\cite{sdvae2022} feature space and a conditional-bridge variant that replaces the Gaussian source with a noisy bicubic-upsampled DESI input. We then sweep the three inference-time hyperparameters of MC-FS, the forward Gaussian PSF width $\sigma_{\mathrm{PSF}}$, the Wiener regularization $\lambda_{\mathrm{SNR}}$, and the correction step size $\eta_0$. We close with detection-overlay visualisations on six DESI--HST patches showing how MC-FS suppresses spurious sources relative to naive OT-CFM sampling.

\begin{figure}[!t]
    \centering
    \includegraphics[width=\textwidth]{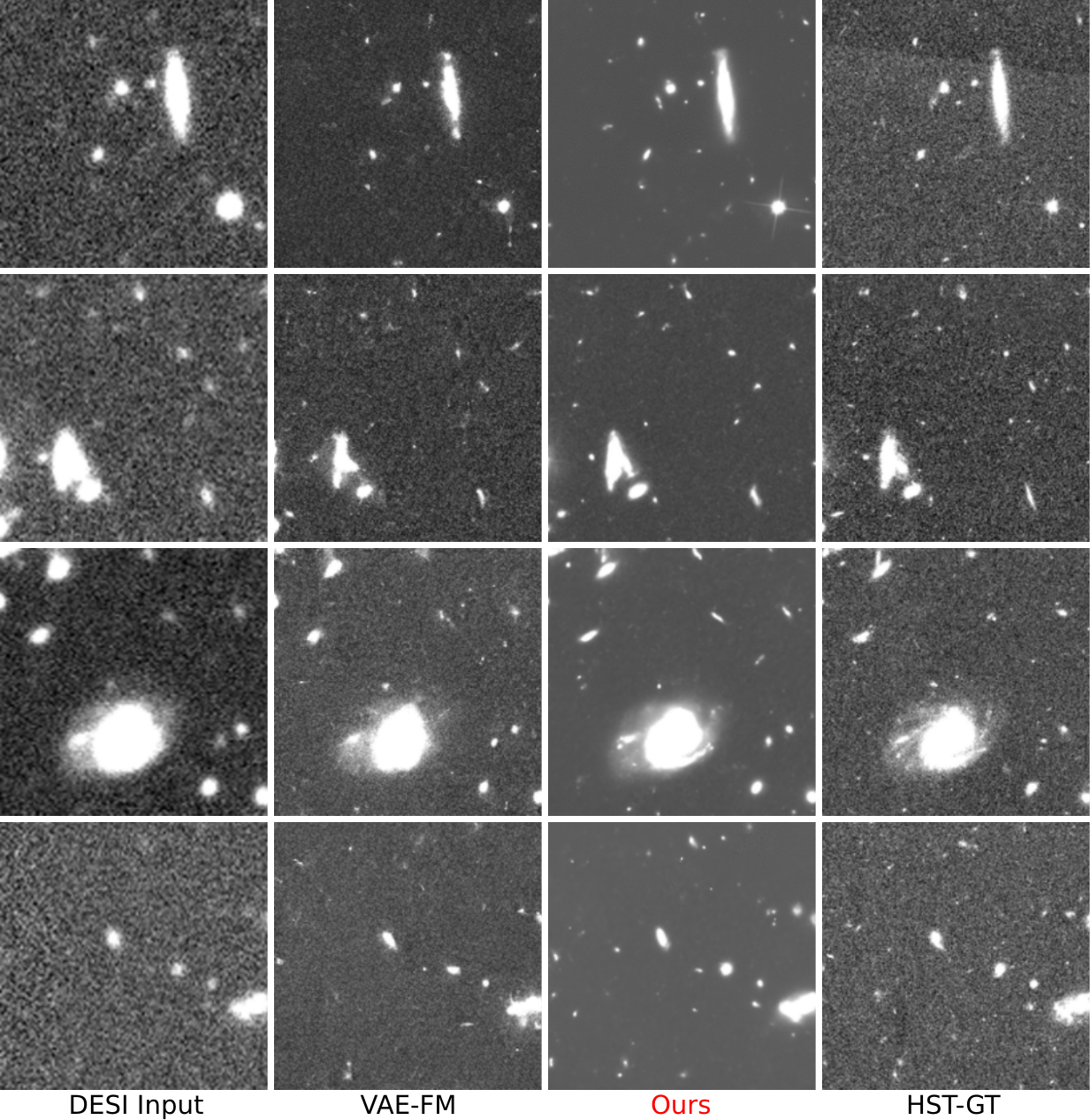}
    \caption{Qualitative comparison of the latent-space flow-matching ablation. Four representative DESI--HST test patches are shown from left to right as the DESI input, the latent-space FM variant operating on SD-VAE features, FluxFlow operating directly in pixel space, and the HST  \DIFaddbeginFL \DIFaddFL{observations}\DIFaddendFL . The VAE-FM blurs faint compact sources and oversmooths the galaxy profiles, while pixel-space FluxFlow recovers source morphology and resolved structure that closely match the HST target.}
    \label{fig:appendix_fm_vae}
\end{figure}

\paragraph{Latent Space Flow-Matching.}
\label{sup:ablation:latent}
Many recent image generative models operate in the latent space of a pretrained VAE~\cite{diederik2019introduction}, which compresses input images and turns the diffusion process into a much lower-dimensional regression problem~\cite{rombach2022high}. We test whether the same recipe transfers to astronomical super-resolution using SD-VAE~\cite{sdvae2022}. The pixel-normalization protocol of \Cref{sup:implementation} maps every cutout to $[0,1]$ before encoding, so the SD-VAE sees inputs in its native operating range. The \emph{w/ VAE} row of \Cref{tab:ablation} reports the result. All photometric and detection metrics degrade substantially relative to pixel-space FluxFlow, with detection F1 collapsing primarily through a sharp drop in recall, indicating that compact sources are systematically washed out by the encoder rather than misplaced. We attribute this collapse to the VAE bottleneck smoothing out the compact, high-contrast point sources that dominate the HST signal. \Cref{fig:appendix_fm_vae} shows the visual signature. The latent-space variant blurs faint compact sources and oversmooths galaxy profiles, recovering only the low-frequency structure that survives the VAE bottleneck, while pixel-space FluxFlow recovers morphology and resolved structure that closely match the HST target. We therefore retain pixel-space training in the main model. For reproducibility, the variant encodes the bicubically upsampled DESI input and the HST target with the frozen SD-VAE encoder into latent codes $z_x$ and $z_y$, then trains a conditional U-Net along the OT-CFM interpolant $z_t = (1-t)\,z_0 + t\,z_y$ from a Gaussian prior $z_0 \sim \mathcal{N}(0, I)$, decoding back to pixel space at inference and averaging the three channels into a single-channel HR estimate. The latent U-Net uses 8 input and 4 output channels with $\text{base\_dim}=64$ and $\text{time\_dim}=256$, trained with AdamW for 300 epochs at batch size 88, learning rate $10^{-4}$, weight decay $10^{-4}$, 1000 warmup steps, gradient clipping at 1.0, and no EMA.

\paragraph{Conditional-Bridge Flow-Matching.}
The Gaussian source distribution of OT-CFM places the inference-time trajectory at pure noise, forcing the velocity network to recover the entire HR signal across certain Euler steps. A natural alternative is to start from a coarse condition-aligned image and refine the residual. We test this by replacing the Gaussian source with a noisy bicubic upsample of the DESI observation, $\mathbf{x}_0 = \mathrm{BicubicUp}(\mathbf{y}) + \boldsymbol{\epsilon}$ with $\boldsymbol{\epsilon}\!\sim\!\mathcal{N}(\mathbf{0}, 0.4^2\mathbf{I})$, where the noise scale $0.4$ was tuned on a small validation grid. The \emph{w/ Bridge} row of \Cref{tab:ablation} is intermediate. All photometric and detection metrics degrade modestly relative to the pixel-space OT-CFM model, with smaller gaps than the latent-space variant but consistent in direction across PSNR, Flux-L1, and detection F1. The interpretation is that the bridge starting point indeed helps the velocity field locally but couples the prior too tightly to the bicubic mean, suppressing the high-frequency content that pixel-space OT-CFM recovers under MC-FS guidance from a noise prior. We therefore retain the Gaussian source in the main model.

\begin{figure}[!t]
    \centering
    \begin{subfigure}{\textwidth}
        \centering
        \includegraphics[width=\linewidth]{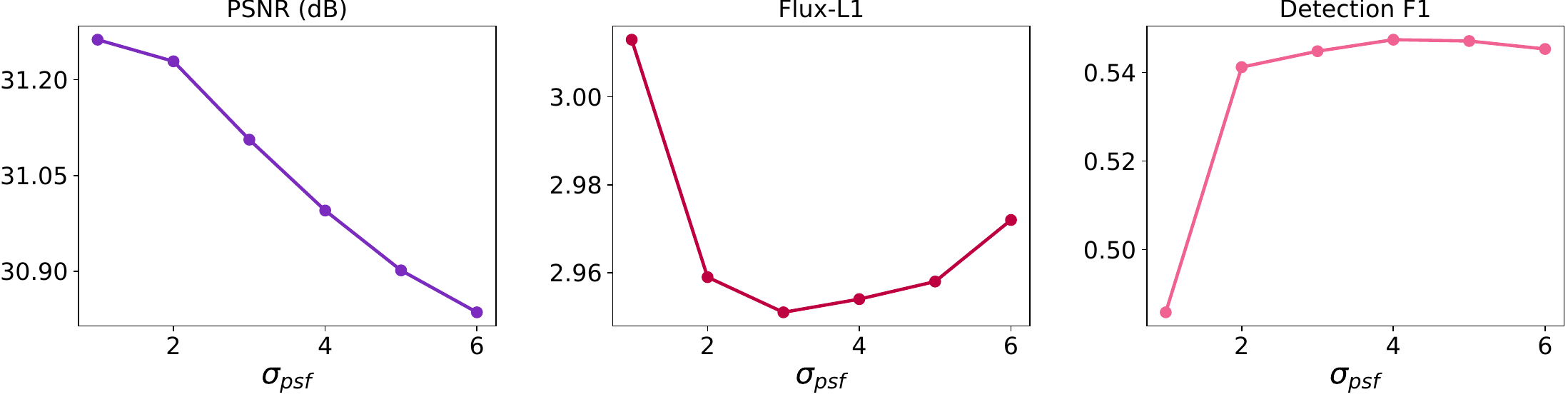}
        \caption{$\times 2$ scale.}
    \end{subfigure}
    \begin{subfigure}{\textwidth}
        \centering
        \includegraphics[width=\linewidth]{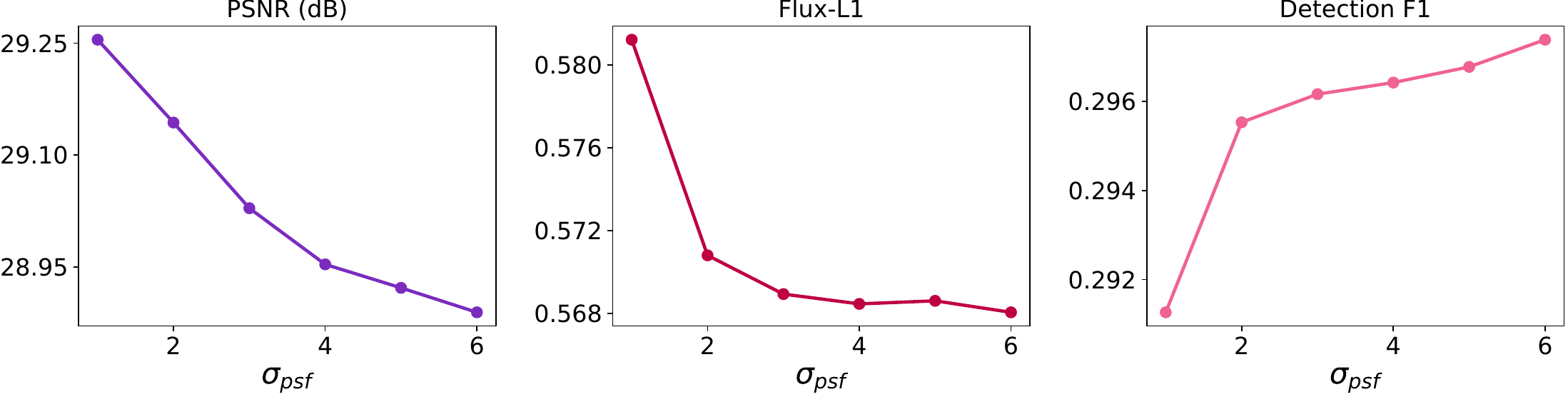}
        \caption{$\times 4$ scale.}
    \end{subfigure}
    \caption{Effect of the forward PSF width $\sigma_{\mathrm{PSF}}$ used inside MC-FS at $\times 2$ (top) and $\times 4$ (bottom). PSNR decreases monotonically with $\sigma_{\mathrm{PSF}}$, while Flux-L1 and Detection F1 form broad plateaus over an intermediate range. We set $\sigma_{\mathrm{PSF}}=2$ HR pixels at both scales in our experiments.}
    \label{fig:abl_psf_sigma}
\end{figure}

\paragraph{Effect of Forward PSF Kernel $\sigma_{\mathrm{PSF}}$.}
The forward operator $\mathbf{A}=\mathbf{D}_s\circ\mathbf{H}_{\mathrm{PSF}}$ used inside MC-FS approximates the DESI seeing-limited PSF by a single isotropic Gaussian of width $\sigma_{\mathrm{PSF}}$, which controls how aggressively the Wiener correction back-projects the LR residual onto the HR grid. \Cref{fig:abl_psf_sigma} sweeps $\sigma_{\mathrm{PSF}}$ from $1$ to $6$ HR pixels at both scales while the rest of the pipeline is held fixed. PSNR decreases monotonically with $\sigma_{\mathrm{PSF}}$ at both scales because a wider forward model relaxes the high-frequency agreement with the DESI observation. Flux-L1 and Detection F1 follow a different trade. At $\times 2$, Flux-L1 reaches its minimum near $\sigma_{\mathrm{PSF}}\!\in\![3,5]$ ($\sim\!2.95$) and Detection F1 forms a broad plateau across $\sigma_{\mathrm{PSF}}\!\in\![2,6]$ (between $0.541$ and $0.547$), dropping sharply only at the very narrow setting $\sigma_{\mathrm{PSF}}=1$ ($0.486$) where over-aggressive deconvolution destroys faint sources. The $\times 4$ trend is qualitatively similar but flatter: PSNR decreases smoothly from $29.25$\,dB at $\sigma_{\mathrm{PSF}}=1$ to $28.89$\,dB at $\sigma_{\mathrm{PSF}}=6$, while Flux-L1 ($\sim\!0.57$) and Detection F1 ($\sim\!0.296$) sit on near-flat plateaus across $\sigma_{\mathrm{PSF}}\!\geq\!2$. Our default $\sigma_{\mathrm{PSF}}=2$ at both scales preserves the high-PSNR end while staying inside the Flux-L1 and F1 plateaus, which we found gave the most consistent reconstructions across patches with widely varying DESI seeing.

\begin{figure}[!t]
    \centering
    \includegraphics[width=\textwidth]{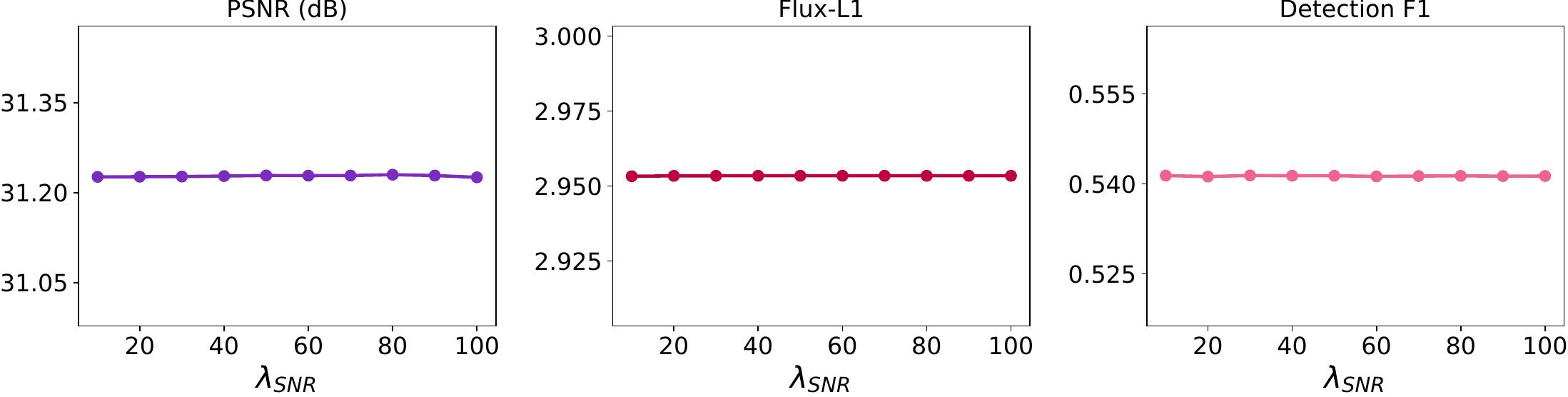}
    \caption{Effect of the Wiener regularization $\lambda_{\mathrm{SNR}}$ at $\times 2$. Sweeping $\lambda_{\mathrm{SNR}}$ from $10$ to $100$ leaves PSNR, Flux-L1, and Detection F1 essentially unchanged,  \DIFaddbeginFL \DIFaddFL{under }\DIFaddendFL the  \DIFaddbeginFL \DIFaddFL{calibrated Gaussian forward model}\DIFaddendFL . Our default $\lambda_{\mathrm{SNR}}=50$ is the value used throughout \Cref{tab:experiment}.}
    \label{fig:abl_lambda_snr}
\end{figure}

\paragraph{Effect of Wiener Regularization $\lambda_{\mathrm{SNR}}$.}
The  \DIFaddbegin \DIFadd{parameter }\DIFaddend $\lambda_{\mathrm{SNR}}$ in  \Cref{eq:wiener_kernel}  \DIFaddbegin \DIFadd{controls regularization }\DIFaddend of the  \DIFaddbegin \DIFadd{correction. For small values }\DIFaddend the  filter \DIFaddbegin \DIFadd{vanishes in magnitude while remaining proportional to }\DIFaddend $H^*(\mathbf{k})$\DIFaddbegin \DIFadd{. }\DIFaddend At  large  \DIFaddbegin \DIFadd{values it approaches an inverse where $H(\mathbf{k})\neq0$}\DIFaddend . \Cref{fig:abl_lambda_snr}  \DIFaddbegin \DIFadd{reports a sweep }\DIFaddend from $10$ to $100$ at  \DIFaddbegin \DIFadd{$\times2$ }\DIFaddend with  \DIFaddbegin \DIFadd{the }\DIFaddend other settings fixed.  PSNR , Flux-L1 , and Detection F1  \DIFaddbegin \DIFadd{vary little over this tested range, which contains the default $50$}\DIFaddend . This  \DIFaddbegin \DIFadd{is empirical robustness under }\DIFaddend the \DIFaddbegin \DIFadd{calibrated Gaussian model, not a guarantee outside the tested interval or under PSF misspecification. The }\DIFaddend spectral \DIFaddbegin \DIFadd{interpretation depends on frequency. For frequencies satisfying $|H(\mathbf{k})|^2\gg\lambda_{\mathrm{SNR}}^{-1}$, changing the regularizer has little effect on the filter denominator.}\DIFaddend  \DIFaddbegin \DIFadd{This explains why a }\DIFaddend plateau \DIFaddbegin \DIFadd{can occur without making the parameter irrelevant. It is not a claim that every spatial frequency satisfies that inequality. Anisotropic or spatially varying PSFs can have directional attenuation and position-dependent spectra}\DIFaddend , \DIFaddbegin \DIFadd{so one global Fourier-domain filter }\DIFaddend and the  \DIFaddbegin \DIFadd{observed robustness need not carry over }\DIFaddend to  \DIFaddbegin \DIFadd{those settings}\DIFaddend .

\begin{figure}[!t]
    \centering
    \includegraphics[width=\textwidth]{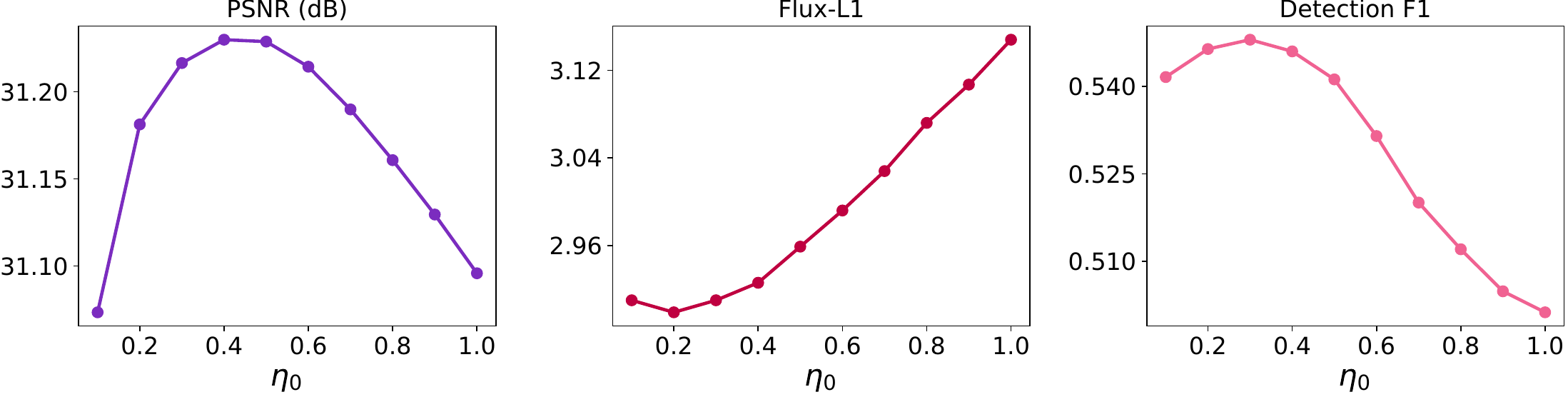}
    \caption{Effect of the base Wiener correction step size $\eta_0$ at $\times 2$, with the per-step decay $\eta_i = \eta_0(1-i/N)$ otherwise unchanged. PSNR is concave with a peak near $\eta_0\!\approx\!0.4$, Flux-L1  \DIFaddbeginFL \DIFaddFL{reaches its minimum near $\eta_0=0.2$ and then rises}\DIFaddendFL , and Detection F1 peaks near $\eta_0\!\approx\!0.3$ before falling. Our default $\eta_0=0.5$ is the value used throughout \Cref{tab:experiment} and sits just to the right of the joint optimum, biasing toward a slightly more conservative correction.}
    \label{fig:abl_eta}
\end{figure}

\paragraph{Effect of Wiener Correction Step Size $\eta_0$.}
The base step size $\eta_0$ controls how aggressively each MC-FS iteration projects the candidate state toward the consistency set $\mathcal{C}(\mathbf{y})$, with the schedule $\eta_i = \eta_0(1 - i/N)$  \DIFaddbegin \DIFadd{giving a final applied coefficient $\eta_{N-1}=\eta_0/N$. }\DIFaddend The  \DIFaddbegin \DIFadd{continuous schedule reaches }\DIFaddend zero \DIFaddbegin \DIFadd{only }\DIFaddend at  \DIFaddbegin \DIFadd{$t=1$}\DIFaddend . \Cref{fig:abl_eta} sweeps $\eta_0$ from $0.1$ to $1.0$ at $\times 2$ with all other settings fixed. PSNR is concave, peaking at $\eta_0=0.4$ with $31.230$\,dB and degrading to $31.073$ and $31.096$\,dB at $\eta_0=0.1$ and $\eta_0=1.0$. Too small a step under-corrects the residual and leaves hallucinated structure in place\DIFadd{. T}oo large a step over-shoots into a non-physical region of $\mathcal{C}(\mathbf{y})$ and injects artificial flux. Flux-L1 confirms this over-shoot, rising from its minimum $2.899$ at $\eta_0=0.2$ to $3.148$ at $\eta_0=1.0$. Detection F1 peaks at $\eta_0=0.3$ ($0.548$) and falls steadily for $\eta_0\geq0.5$ as over-correction begins to suppress real low-SNR sources alongside hallucinations. Our default $\eta_0=0.5$ sits just to the right of the joint optimum,  \DIFaddbegin \DIFadd{retaining the original reconstruction setting; this sweep does not establish robustness to arbitrary }\DIFaddend seeing \DIFaddbegin \DIFadd{variation}\DIFaddend .

\begin{figure}[!t]
    \centering
    \includegraphics[width=\textwidth]{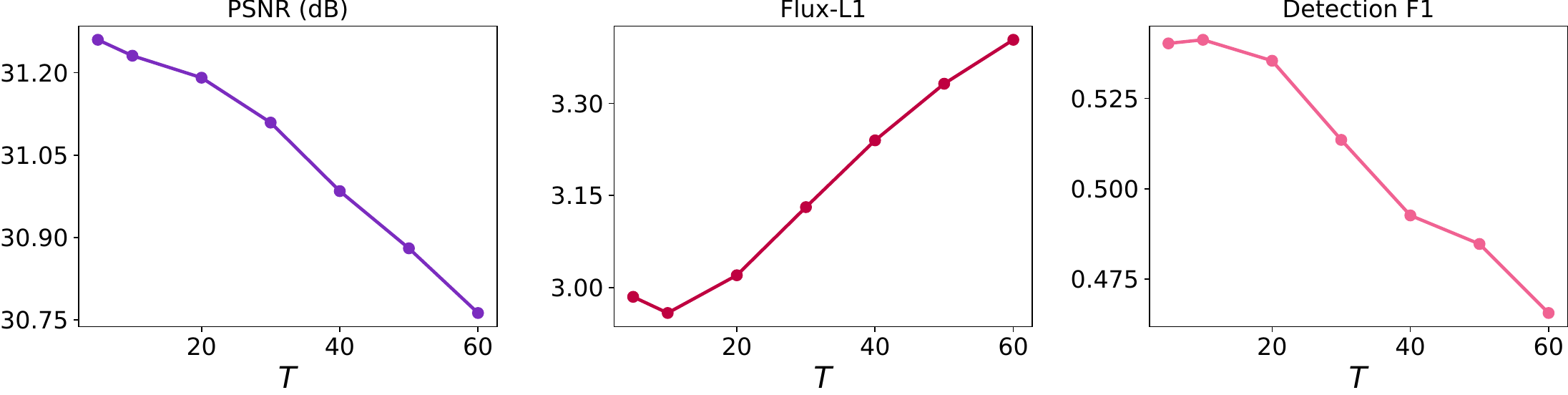}
    \caption{Effect of the number of sampling steps $T$ at $\times 2$, with all other MC-FS hyperparameters held at their defaults. PSNR decreases monotonically with $T$\DIFaddFL{.} Flux-L1 and Detection F1 attain their joint optimum at $T=10$ before degrading at larger $T$  \DIFaddbeginFL \DIFaddFL{in this experiment}\DIFaddendFL . Our default $T=10$ is the value used throughout \Cref{tab:experiment}.}
    \label{fig:abl_num_steps}
\end{figure}

\paragraph{Effect of Number of Sampling Steps $T$.}
The number of sampling steps $T$ (denoted $N$ in \Cref{alg:mcfm}) sets the discretization of the OT-CFM ODE and, jointly with the linear decay $\eta_i = \eta_0(1-i/T)$, the per-step magnitude of the Wiener correction. \Cref{fig:abl_num_steps} sweeps $T$ from $5$ to $60$ at $\times 2$ with all other settings fixed. PSNR is highest at $T=5$ ($31.26$\,dB) and decreases monotonically to $30.76$\,dB at $T=60$, while Flux-L1 attains its minimum at $T=10$ ($2.96$) and rises steadily to $3.40$ at $T=60$. Detection F1 peaks at $T=10$ ($0.541$) and falls to $0.466$ at $T=60$.  \DIFaddbegin \DIFadd{Changing }\DIFaddend $T$  \DIFaddbegin \DIFadd{alters both ODE discretization }\DIFaddend and  the  \DIFaddbegin \DIFadd{number of corrections. At a fixed relative time $i/T$}\DIFaddend ,  the  \DIFaddbegin \DIFadd{stated schedule has no $1/T$ factor; these measurements do not isolate the cause of the trend}\DIFaddend . Our default $T=10$ sits at the joint optimum of Flux-L1 and Detection F1 with PSNR within a small margin of its peak \Cref{tab:experiment}.

\begin{figure}[!ht]
    \centering
    \includegraphics[width=\textwidth]{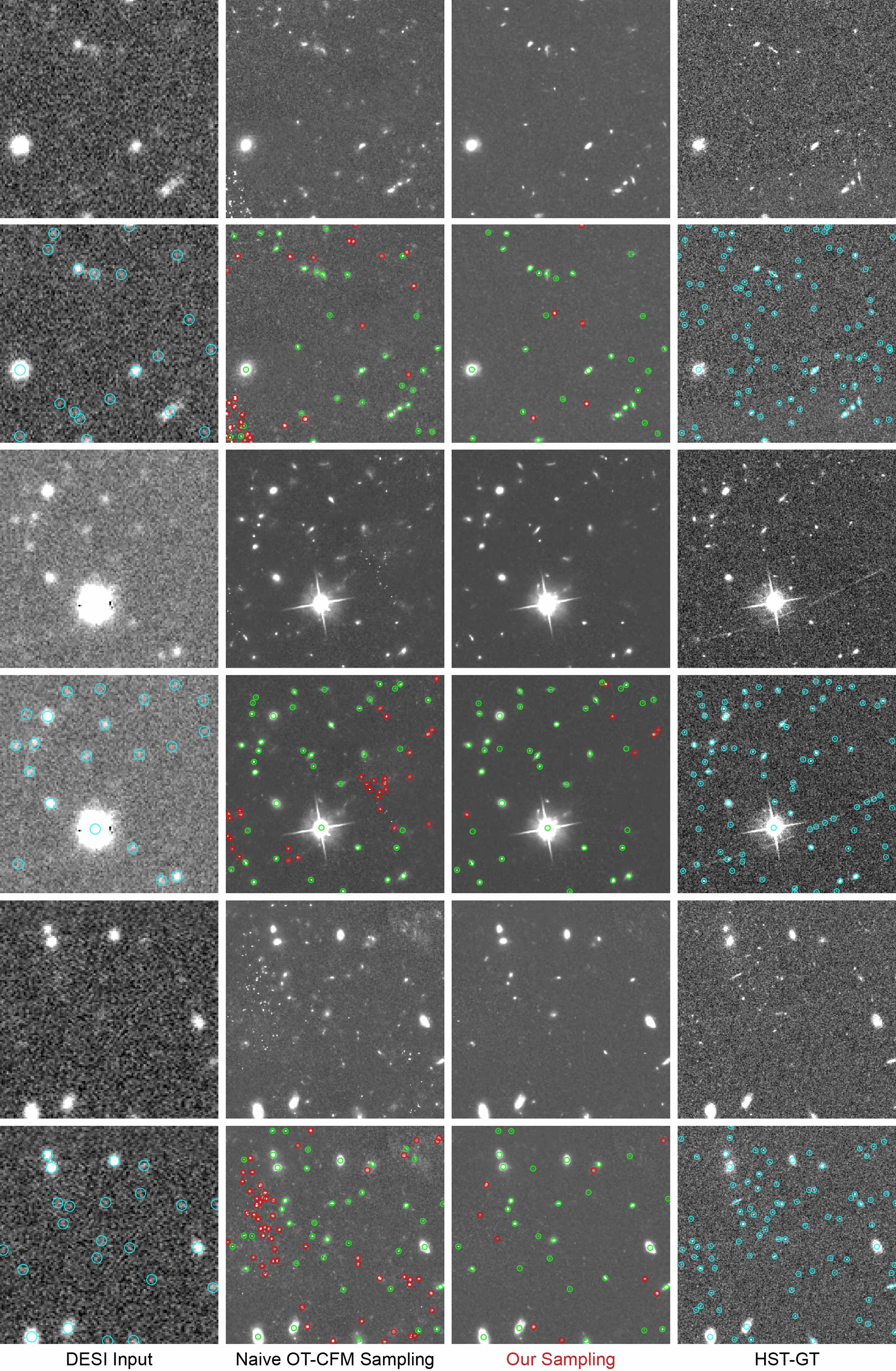}
    \caption{MC-FS hallucination suppression on three DESI--HST examples. Columns show the DESI input, naive OT-CFM, MC-FS, and HST  . The second row overlays SExtractor detections, with \textcolor{cyan}{Cyan} for ground-truth sources, \textcolor{green}{Green} for true positives, and \textcolor{red}{Red} for false positives.}
    \label{fig:sup_sampling_2}
\end{figure}

\begin{figure}[!ht]
    \centering
    \includegraphics[width=\textwidth]{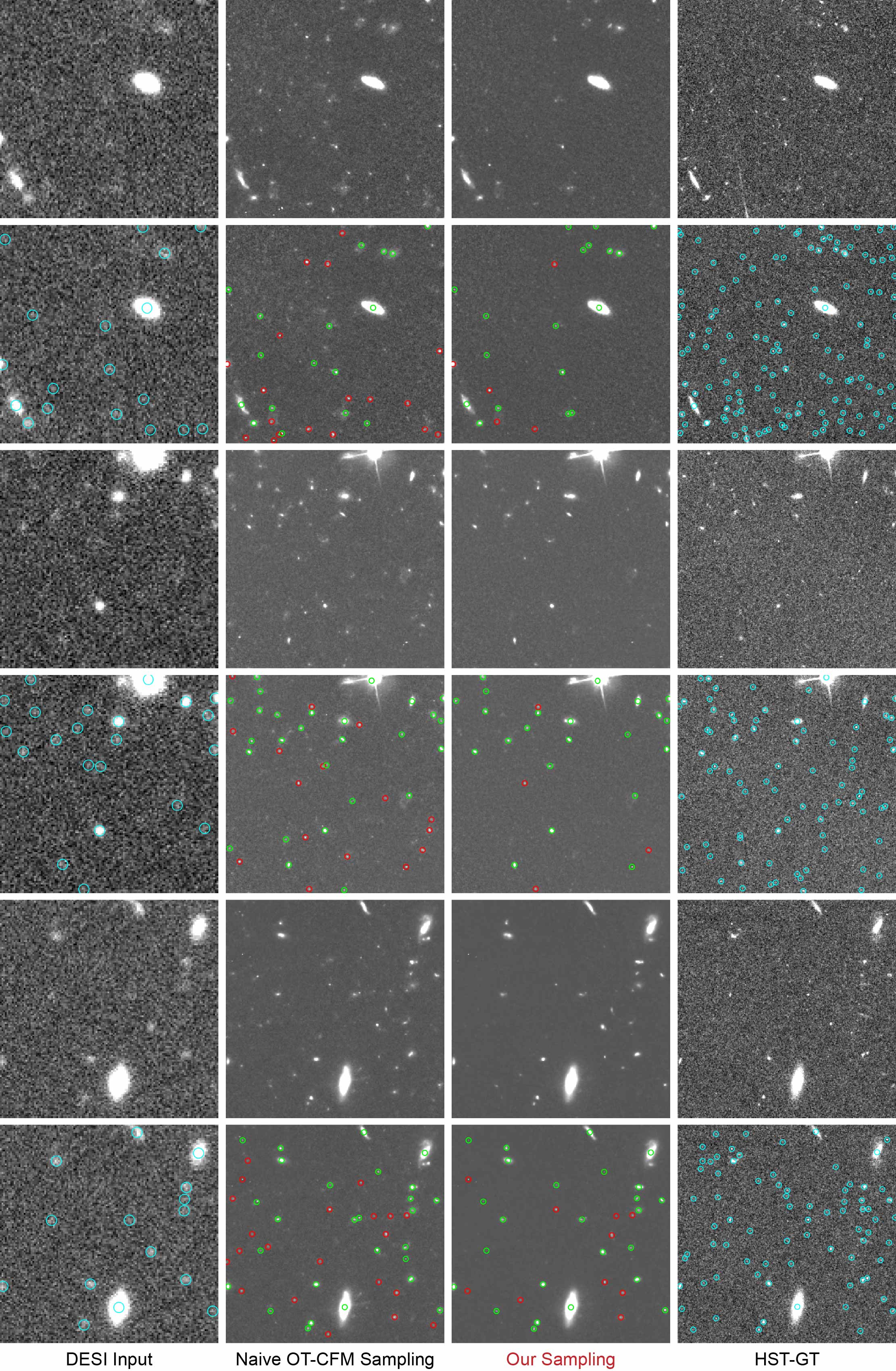}
    \caption{Additional MC-FS detection comparisons under the protocol of \Cref{fig:sup_sampling_2}, covering a faint background field, a saturated bright star with diffraction spikes, and a moderately crowded region. \textcolor{cyan}{Cyan} for ground-truth sources, \textcolor{green}{Green} for true positives, and \textcolor{red}{Red} for false positives.}
    \label{fig:sup_sampling_1}
\end{figure}

\paragraph{Visualization of Source Detection.}
\Cref{fig:sup_sampling_2,fig:sup_sampling_1} visualize how MC-FS suppresses hallucination sources across several representative DESI--HST test images. Each example uses four columns showing the DESI input, naive OT-CFM sampling without MC-FS, our MC-FS sampling, and the HST ground-truth, with the top row giving the reconstruction and the bottom row overlaying the SExtractor catalogue measured on the same image. Cyan circles mark ground-truth sources, green circles mark true-positive detections, and red dots mark spurious detections without a counterpart in the ground truth. The naive column is densely populated with red dots in low-SNR background regions and around the wings of bright stars, the regimes in which the velocity field most readily sharpens noise into compact source-like morphology with no preimage in the DESI observation. The MC-FS column eliminates most of these red dots while preserving the green true-positive population, recovering precision without collapsing recall. The suppression is most pronounced on the saturated-star examples in \Cref{fig:sup_sampling_2}, where strong diffraction spikes induce many spurious sources under naive sampling but are correctly recognized as instrumental structure once the residual against the DESI observation is enforced. \Cref{fig:sup_sampling_1} shows the same pattern on a bright extended galaxy and a faint background field, where naive sampling fragments the extended profile and hallucinates compact peaks in the surrounding noise. These visualizations are the per-image counterpart of the precision and F1 gains of the full model over the \emph{w/o MCFS} ablation in \Cref{tab:ablation}.

\section{Downstream Scientific Application}
We complement the field-aggregated metrics in \Cref{tab:experiment} with a per-source evaluation that converts each reconstruction into the quantities that actually enter downstream catalog products. The relevant quantities are AB magnitudes for photometric calibration, a luminosity-proxy stellar mass for galaxy population statistics, and second-moment ellipticities for weak-lensing shape catalogs. Sources are read from a precomputed SEP~\cite{bertin1996sextractor,barbary2016sep} segmentation map of the HST  , with labels of fewer than $10$ pixels discarded. Every reconstruction is measured under the same segmentation footprint $\mathcal{R}_i$ as the ground truth, so each prediction inherits a strictly one-to-one source correspondence with the GT and detector-side variability does not enter the comparison.

\paragraph{Local Background Subtraction.} Before any per-source measurement, we subtract a local background mesh from both the ground truth and each reconstruction. The mesh is computed with SEP's \texttt{Background} routine on a $32\times 32$ pixel grid with mesh-size parameters $\text{bw}=\text{bh}=32$ and mesh-filtering parameters $\text{fw}=\text{fh}=1$ that disable additional smoothing. The per-pixel weight map $\mathbf{W}$ acts as a validity mask in which pixels at $\mathbf{W}_p = 0$ are excluded from both the background fit and the subsequent measurement. The HST   has already had its background removed during SEP preprocessing, but the residual background bias of each reconstruction is method-dependent and systematic, negative for bicubic upsampling and positive for several generative baselines, and would otherwise translate into a global flux offset that pushes a non-trivial fraction of $\hat{F}_i$ negative and out of the magnitude domain. All formulas below operate on the background-subtracted image.

\paragraph{Bright-Source Protocol.} For the scientific tasks considered here the relevant population is the bright, resolved end where photometric, structural, and shear estimates are stable rather than dominated by sub-detection-limit fluctuations. We therefore retain only sources whose ground-truth integrated flux satisfies
\begin{equation}
    F_i^{\mathrm{GT}} \;=\; \sum_{p \in \mathcal{R}_i} x_p^{\mathrm{GT}} \;\geq\; F_{\mathrm{thr}},
    \label{eq:bright_threshold}
\end{equation}
on the background-subtracted GT, with $\mathcal{R}_i$ the SEP segmentation footprint of source $i$. We use the same threshold $F_{\mathrm{thr}} = 0.5$\,nMgy at both $\times 2$ and $\times 4$. Because the integrated flux of a fixed physical source is conserved across the two scales by construction of the dataset, the same threshold selects the same physical population, yielding $N_{\mathrm{src}} = 7437$ matched bright sources at both scales. All metrics below are reported as both the mean and the median across this set, the median being the relevant robust statistic for catalog products in which a small number of saturated or hallucinated cores would otherwise dominate the mean.

\paragraph{Magnitude Error.} For each source the integrated flux on the reconstruction is $\hat{F}_i = \sum_{p\in\mathcal{R}_i}\hat{x}_p$, measured on the background-subtracted image and the same footprint as $F_i^{\mathrm{GT}}$. The AB-style magnitude is computed with a calibrated zero point $Z_{\mathrm{p}}=25$,
\begin{equation}
    m_i \;=\; -2.5\,\log_{10}(F_i) + Z_{\mathrm{p}},
    \label{eq:mag_def}
\end{equation}
and the per-source magnitude residual is
\begin{equation}
    \Delta m_i \;=\; \bigl| \hat{m}_i - m_i^{\mathrm{GT}} \bigr| \;=\; 2.5 \,\bigl| \log_{10}\hat{F}_i - \log_{10} F_i^{\mathrm{GT}} \bigr|,
    \label{eq:mag_err}
\end{equation}
evaluated only on sources with both $\hat{F}_i > 0$ and $F_i^{\mathrm{GT}} > 0$. This is the standard photometric residual used for source-catalog calibration and for color-magnitude diagrams. The logarithmic scaling makes $\Delta m$ sensitive to multiplicative photometric bias, so a method that systematically biases integrated flux high or low is penalized regardless of the absolute scale.

\paragraph{Stellar-Mass Error.} A bright resolved source contributes to galaxy stellar-mass functions and SFR-mass diagrams~\cite{conselice2014evolution} primarily through its luminosity. We adopt a single-band luminosity-proxy stellar mass with a fiducial mass-to-light ratio $\Theta_\star$ held common across reconstructions and ground truth,
\begin{equation}
    \hat{M}_{\star,i} \;=\; \Theta_\star \cdot \hat{F}_i, \qquad \Theta_\star = 3,
    \label{eq:mass_def}
\end{equation}
on the same nanomaggie-scale flux as \Cref{eq:mag_def}. The mean absolute error is
\begin{equation}
    \mathrm{Mass\,MAE} \;=\; \frac{1}{N_{\mathrm{src}}}\sum_{i=1}^{N_{\mathrm{src}}} \bigl|\, \hat{M}_{\star,i} - M_{\star,i}^{\mathrm{GT}} \,\bigr|,
    \label{eq:mass_mae}
\end{equation}
reported in the same unit as $\Theta_\star \cdot F$, namely $\Theta_\star$-scaled nanomaggies. Unlike $\Delta m$, the linear form weights bright sources more heavily than faint ones, exposing methods that smear or hallucinate the high-luminosity end of the population, the regime that dominates galaxy stellar-mass-function inference.

\begin{table}[!tp]
\centering
\footnotesize
\caption{Per-source downstream-task accuracy at $\times 2$ on the matched bright-source set. Magnitude and shear are dimensionless, while the stellar-mass MAE is reported in $\Theta_\star$-scaled nanomaggies as defined in \Cref{eq:mass_def}. Best, second, and third are shaded with \textcolor{fst}{$\blacksquare$}, \textcolor{sec}{$\blacksquare$}, \textcolor{thd}{$\blacksquare$}.}
\label{tab:downstream_x2}
\setlength{\tabcolsep}{3pt}
\renewcommand{\arraystretch}{1}
{\fontsize{8pt}{10pt}\selectfont
\begin{tabularx}{\textwidth}{l|>{\centering\arraybackslash}X >{\centering\arraybackslash}X|>{\centering\arraybackslash}X >{\centering\arraybackslash}X|>{\centering\arraybackslash}X >{\centering\arraybackslash}X}
\hline
& \multicolumn{2}{c|}{$\Delta m\ \downarrow$} & \multicolumn{2}{c|}{Mass\,MAE\ $\downarrow$} & \multicolumn{2}{c}{$\Delta\gamma\ \downarrow$} \\
\cmidrule(lr){2-3}\cmidrule(lr){4-5}\cmidrule(lr){6-7}
Method & mean & median & mean & median & mean & median \\
\hline
Bicubic~\cite{keys1981cubic} & 0.758 & 0.637 & 7.767\thd & 1.537 & 0.148\fst & 0.122\thd \\
SwinIR~\cite{liang2021swinir} & 0.649 & 0.300 & 8.197 & 0.762 & 0.149\sed & 0.116\sed \\
HAT~\cite{chen2025hat} & 0.542\sed & 0.192\sed & 7.962 & 0.524\sed & 0.151\thd & 0.112\fst \\
FISR~\cite{wu2025star} & 0.561 & 0.230 & 8.014 & 0.608 & 0.165 & 0.133 \\
cGAN~\cite{rai2025generative} & 0.548\thd & 0.215\thd & 7.996 & 0.586\thd & 0.159 & 0.131 \\
GD-Net~\cite{shan2025galaxy} & 0.664 & 0.393 & 7.659\sed & 1.025 & 0.166 & 0.131 \\
\textbf{Ours} & 0.516\fst & 0.187\fst & 7.632\fst & 0.508\fst & 0.161 & 0.126 \\
\hline
\end{tabularx}
}
\end{table}

\begin{table}[!tp]
\centering
\footnotesize
\caption{Per-source downstream-task accuracy at $\times 4$ on the same matched bright-source set as \Cref{tab:downstream_x2}. Stellar-mass MAE is in $\Theta_\star$-scaled nanomaggies.}
\label{tab:downstream_x4}
\setlength{\tabcolsep}{3pt}
\renewcommand{\arraystretch}{1}
{\fontsize{8pt}{10pt}\selectfont
\begin{tabularx}{\textwidth}{l|>{\centering\arraybackslash}X >{\centering\arraybackslash}X|>{\centering\arraybackslash}X >{\centering\arraybackslash}X|>{\centering\arraybackslash}X >{\centering\arraybackslash}X}
\hline
& \multicolumn{2}{c|}{$\Delta m\ \downarrow$} & \multicolumn{2}{c|}{Mass\,MAE\ $\downarrow$} & \multicolumn{2}{c}{$\Delta\gamma\ \downarrow$} \\
\cmidrule(lr){2-3}\cmidrule(lr){4-5}\cmidrule(lr){6-7}
Method & mean & median & mean & median & mean & median \\
\hline
Bicubic~\cite{keys1981cubic} & 1.067 & 0.863 & 16.90 & 3.648 & 0.144 & 0.115 \\
SwinIR~\cite{liang2021swinir} & 0.969 & 0.461 & 16.51 & 2.43 & 0.118\thd & 0.090\sed \\
HAT~\cite{chen2025hat} & 0.886\thd & 0.397\sed & 16.19\sed & 2.207\sed & 0.108\fst & 0.081\fst \\
FISR~\cite{wu2025star} & 0.876\sed & 0.417\thd & 16.25\thd & 2.246\thd & 0.124 & 0.097 \\
cGAN~\cite{rai2025generative} & 0.902 & 0.460 & 16.42 & 2.440 & 0.135 & 0.109 \\
GD-Net~\cite{shan2025galaxy} & 1.509 & 1.308 & 17.58 & 4.630 & 0.170 & 0.139 \\
\textbf{Ours} & 0.851\fst & 0.389\fst & 16.07\fst & 2.131\fst & 0.117\sed & 0.091\thd \\
\hline
\end{tabularx}
}
\end{table}

\paragraph{Weak-Lensing Shear Error.} Cosmic-shear analyses depend on the recovered shape of background galaxies~\cite{bernstein2002shapes,mandelbaum2015great3}. For each source we compute flux-weighted second-order brightness moments on the background-subtracted reconstruction, with positivity-clipped weights to prevent residual negative-noise pixels from biasing the centroid,
\begin{equation}
    \omega_p \;=\; \max(\hat{x}_p, 0), \qquad \bar{p}_{a,i} \;=\; \frac{\sum_{p\in\mathcal{R}_i} \omega_p\, p_a}{\sum_{p\in\mathcal{R}_i} \omega_p}, \qquad a \in \{x,y\}.
    \label{eq:centroid}
\end{equation}
The second-moment tensor is then
\begin{equation}
    Q_{ab,i} \;=\; \frac{\sum_{p\in\mathcal{R}_i} \omega_p\,(p_a - \bar{p}_{a,i})(p_b - \bar{p}_{b,i})}{\sum_{p\in\mathcal{R}_i} \omega_p}, \qquad a,b \in \{x,y\},
    \label{eq:quadrupole}
\end{equation}
falling back to a uniform weight whenever $\sum_p \omega_p = 0$. The two ellipticity components~\cite{bernstein2002shapes} are
\begin{equation}
    \hat{e}_{1,i} \;=\; \frac{Q_{xx,i} - Q_{yy,i}}{Q_{xx,i} + Q_{yy,i}}, \qquad
    \hat{e}_{2,i} \;=\; \frac{2\, Q_{xy,i}}{Q_{xx,i} + Q_{yy,i}},
    \label{eq:ellipticity}
\end{equation}
and the per-source shear error is the magnitude of the residual ellipticity vector
\begin{equation}
    \Delta\gamma_i \;=\; \sqrt{\bigl(\hat{e}_{1,i} - e_{1,i}^{\mathrm{GT}}\bigr)^2 + \bigl(\hat{e}_{2,i} - e_{2,i}^{\mathrm{GT}}\bigr)^2},
    \label{eq:shear_err}
\end{equation}
where $e_{a,i}^{\mathrm{GT}}$ is the same statistic measured on the HST  \DIFaddbegin \DIFadd{observations}\DIFaddend . The median of $\{\Delta\gamma_i\}$ is the relevant statistic for cosmic-shear pipelines, where bulk biases on the bright end propagate directly into the inferred lensing power spectrum. Measuring the moments on the GT segmentation footprint, rather than re-running source detection on the reconstruction, isolates the per-source shape error from the orthogonal failure mode of catalog-level miscentering.

\paragraph{Discussion.} At both $\times 2$ and $\times 4$, FluxFlow obtains the best mean and the best median on four of the six metrics, dominating the magnitude error and the luminosity-proxy stellar-mass MAE on the same matched bright-source set. The remaining shear advantage belongs to HAT, SwinIR, and bicubic upsampling, whose conservative mean-image behaviour preserves elliptical shapes through smoothing rather than recovering high-frequency structure. FluxFlow nevertheless places second on shear mean at $\times 4$ at $0.117$ against the HAT best of $0.108$ and third on shear median at $0.091$ against the HAT best of $0.081$, and the corresponding $\times 2$ gap is similarly small at roughly $8\%$ of the best value. The systematic magnitude-error and stellar-mass advantages indicate that the inverse-variance and source-region weighting of \Cref{eq:wfm_loss} together with the Wiener-corrected MC-FS sampling translate directly into the per-source flux fidelity that downstream catalog products require, while costing only a small amount of shape bias relative to the smoothest regression baselines.

\section{Additional Visualization}
\label{sup:sec:visualization_sample}

\begin{figure}[p]
    \centering
    \includegraphics[width=\textwidth]{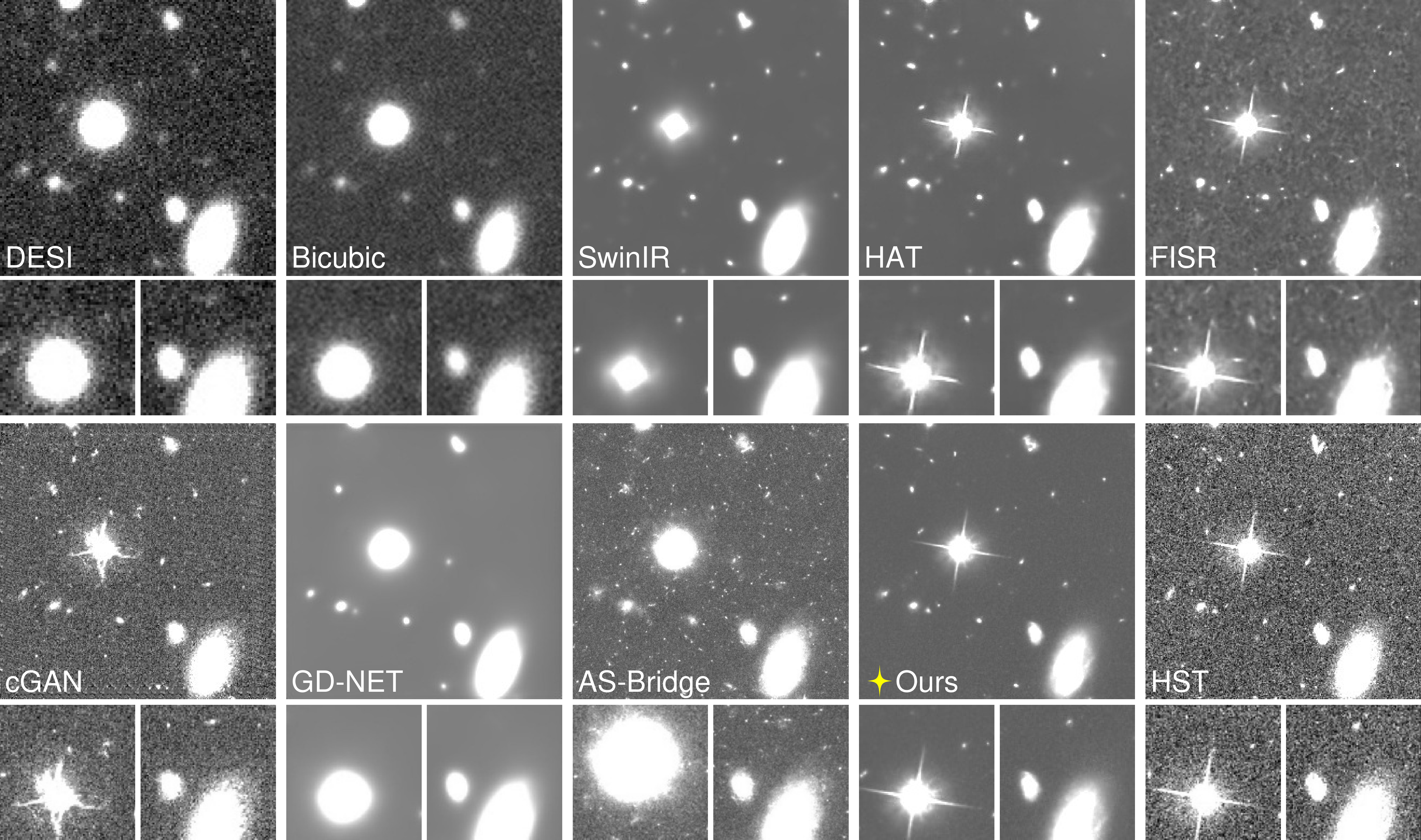}
    \caption{Qualitative DESI--HST $\times 2$ comparison of all evaluated methods, with two zoom-in panels per method. FluxFlow recovers the diffraction spikes of the central star and the morphology of the extended source while avoiding the speckled background of AS-Bridge.}
    \label{fig:appendix_ours_1}
\end{figure}

\begin{figure}[!ht]
    \centering
    \includegraphics[width=\textwidth]{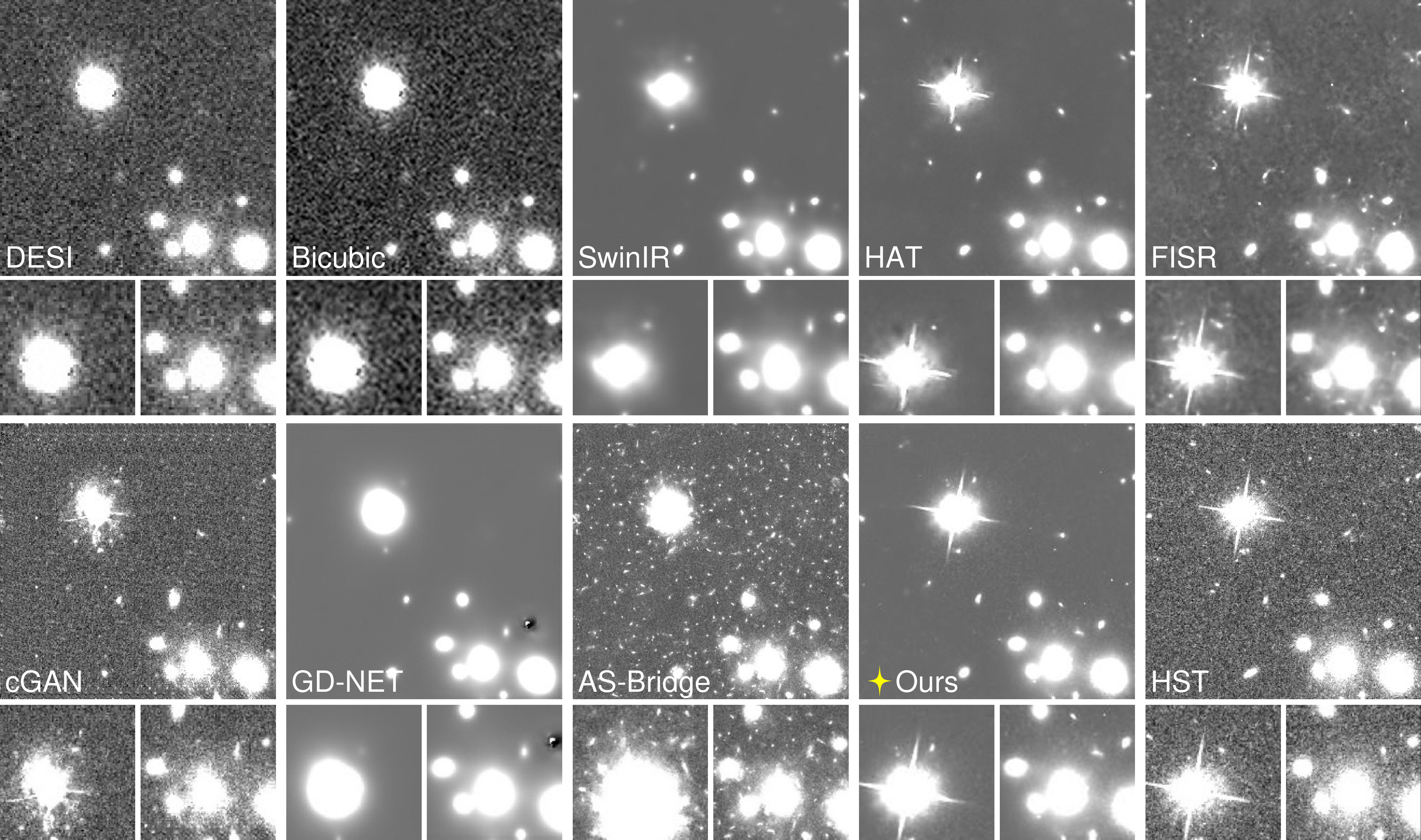}
    \caption{Additional DESI--HST $\times 2$ comparison, on a bright central star surrounded by closely packed compact sources. FluxFlow reproduces the spike pattern and resolves the neighboring sources without the merging or smearing seen in the baselines.}
    \label{fig:appendix_ours_2}
\end{figure}

\begin{figure}[!ht]
    \centering
    \includegraphics[width=\textwidth]{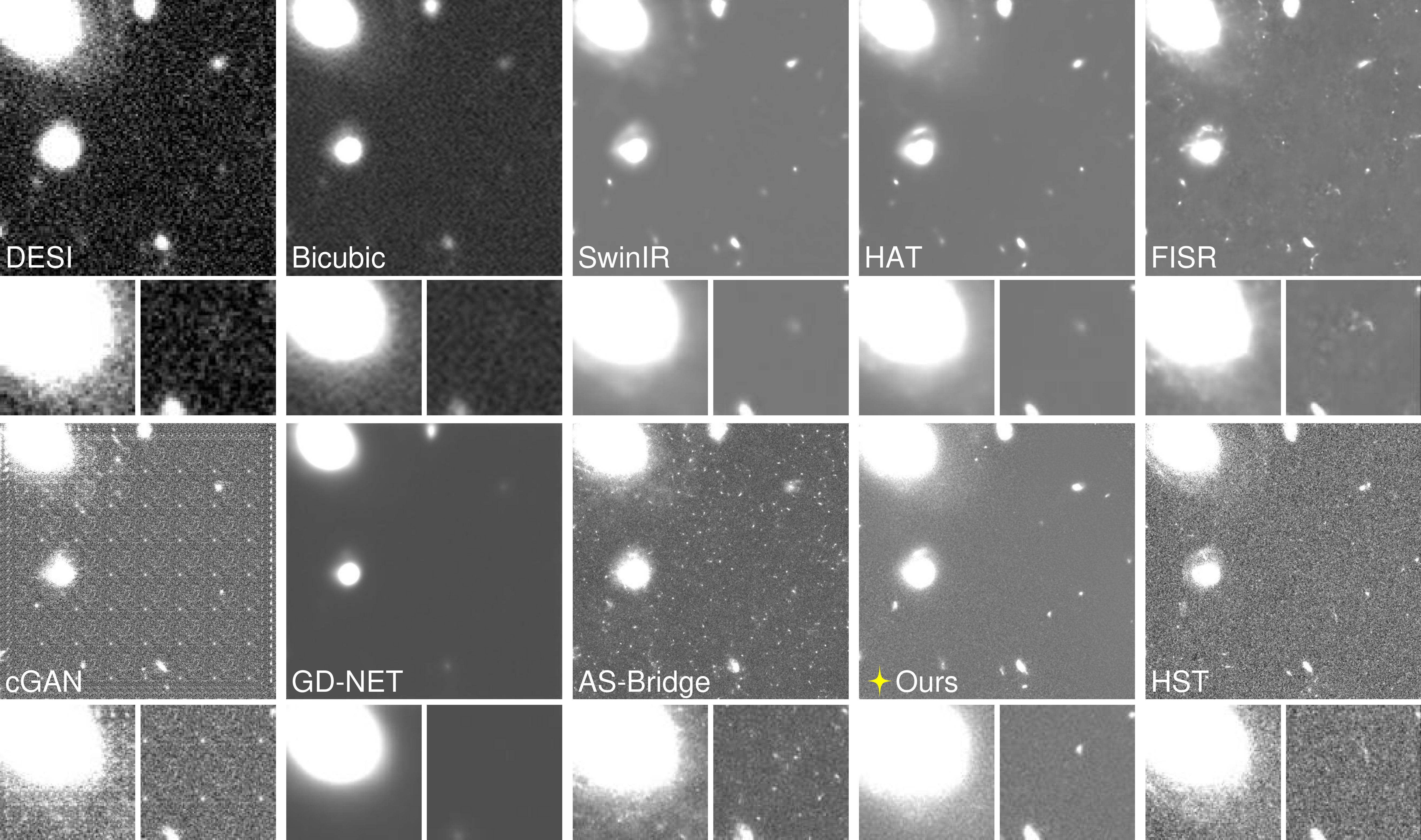}
    \caption{DESI--HST $\times 4$ comparison on a field with a saturated source on the upper-left corner and several low-SNR background galaxies. FluxFlow preserves faint background galaxies that Bicubic and GD-Net erase while reconstructing the saturated halo more cleanly than the speckled output of AS-Bridge.}
    \label{fig:appendix_ours_3}
\end{figure}

\begin{figure}[!ht]
    \centering
    \includegraphics[width=\textwidth]{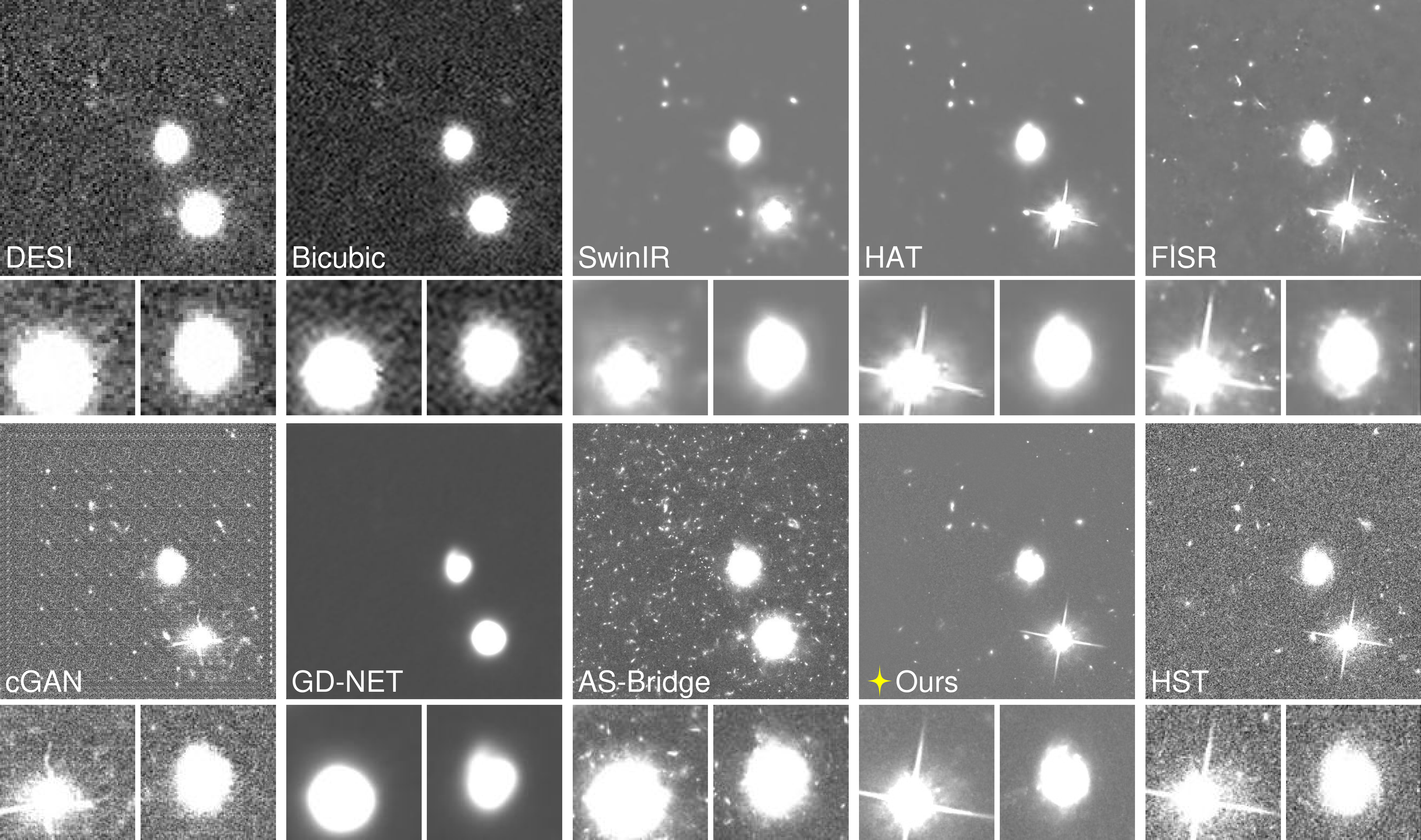}
    \caption{DESI--HST $\times 4$ comparison on a binary-star field. FluxFlow reproduces the four-pointed diffraction pattern on the lower companion together with surrounding faint sources, whereas regression baselines suppress the spike pattern and the generative baselines either oversmooth or oversharpen the background.}
    \label{fig:appendix_ours_4}
\end{figure}

We provide additional qualitative comparisons against all evaluated baselines on the DESI--HST test set. \Cref{fig:appendix_ours_1,fig:appendix_ours_2} present $\times 2$ super-resolution results, the former on a bright central star with an extended companion and the latter on a saturated star surrounded by closely packed compact sources. \Cref{fig:appendix_ours_3,fig:appendix_ours_4} extend the comparison to $\times 4$, covering a faint background-galaxy field with a saturated upper-left source and a binary-star field with resolved diffraction spikes. Across all four cases, FluxFlow recovers diffraction patterns, faint background structures, and source morphology that regression baselines oversmooth and that generative baselines either speckle or oversharpen, consistent with the trends reported in \Cref{tab:experiment}.

\clearpage
\section*{NeurIPS Paper Checklist}

\begin{enumerate}

\item {\bf Claims}
    \item[] Question: Do the main claims made in the abstract and introduction accurately reflect the paper's contributions and scope?
    \item[] Answer: \answerYes{}
        \item[] Justification:  \DIFaddbegin \DIFadd{The abstract }\DIFaddend and  \DIFaddbegin \DIFadd{introduction restrict the claims to the evaluated benchmark and HST-referenced metrics; limitations distinguish these from independent sky truth}\DIFaddend .

\item {\bf Limitations}
    \item[] Question: Does the paper discuss the limitations of the work performed by the authors?
    \item[] Answer: \answerYes{}
        \item[] Justification:  \DIFaddbegin \DIFadd{The conclusion }\DIFaddend and  \DIFaddbegin \DIFadd{Appendix~\ref{sec:reference_scope} discuss PSF misspecification, HST artifacts, bandpass differences, transfer, and uncertainty}\DIFaddend .

\item {\bf Theory assumptions and proofs}
    \item[] Question: For each theoretical result, does the paper provide the full set of assumptions and a complete (and correct) proof?
    \item[] Answer:  \DIFaddbegin \answerNo{}
    \DIFaddend \item[] Justification:  \DIFaddbegin \DIFadd{Appendix~\ref{sup:mcfs} provides derivations for the Wiener correction and simplified spectral models; a general convergence proof for the full sampler is not provided}\DIFaddend .

    \item {\bf Experimental result reproducibility}
    \item[] Question: Does the paper fully disclose all the information needed to reproduce the main experimental results of the paper to the extent that it affects the main claims and/or conclusions of the paper (regardless of whether the code and data are provided or not)?
    \item[] Answer: \answerYes{}
        \item[] Justification: Implementation details included.

\item {\bf Open access to data and code}
    \item[] Question: Does the paper provide open access to the data and code, with sufficient instructions to faithfully reproduce the main experimental results, as described in supplemental material?
    \item[] Answer:  \DIFaddbegin \answerNo{}
    \DIFaddend \item[] Justification:  \DIFaddbegin \DIFadd{Code }\DIFaddend and  \DIFaddbegin \DIFadd{usage instructions }\DIFaddend are  \DIFaddbegin \DIFadd{available at }\url{https://github.com/I2WM/FluxFlow}\DIFaddend .  \DIFaddbegin \DIFadd{The full reproduction package is not yet publicly available}\DIFaddend .

\item {\bf Experimental setting/details}
    \item[] Question: Does the paper specify all the training and test details (e.g., data splits, hyperparameters, how they were chosen, type of optimizer) necessary to understand the results?
    \item[] Answer: \answerYes{}
        \item[] Justification: Included.

\item {\bf Experiment statistical significance}
    \item[] Question: Does the paper report error bars suitably and correctly defined or other appropriate information about the statistical significance of the experiments?
    \item[] Answer:  \DIFaddbegin \answerNo{}
    \DIFaddend \item[] Justification:  \DIFaddbegin \DIFadd{The reconstruction comparisons do not report across-seed confidence intervals or statistical significance tests}\DIFaddend . \DIFaddbegin \DIFadd{Point estimates should not be interpreted as statistically established superiority.

}\DIFaddend

\item {\bf Experiments compute resources}
    \item[] Question: For each experiment, does the paper provide sufficient information on the computer resources (type of compute workers, memory, time of execution) needed to reproduce the experiments?
    \item[] Answer: \answerYes{}
        \item[] Justification: Provided.

\item {\bf Code of ethics}
    \item[] Question: Does the research conducted in the paper conform, in every respect, with the NeurIPS Code of Ethics \url{https://neurips.cc/public/EthicsGuidelines}?
    \item[] Answer: \answerYes{}
        \item[] Justification: \DIFaddbegin \DIFadd{The research uses astronomical survey images and does not involve human subjects or personal data.

}\DIFaddend 

\item {\bf Broader impacts}
    \item[] Question: Does the paper discuss both potential positive societal impacts and negative societal impacts of the work performed?
    \item[] Answer:  \DIFaddbegin \answerYes{}
    \DIFaddend \item[] Justification:  \DIFaddbegin \DIFadd{Appendix~\ref{sec:reference_scope} discusses access to survey analysis and the risk of unsupported scientific interpretation of generated sources, together with the need for independent validation}\DIFaddend .

\item {\bf Safeguards}
    \item[] Question: Does the paper describe safeguards that have been put in place for responsible release of data or models that have a high risk for misuse (e.g., pre-trained language models, image generators, or scraped datasets)?
    \item[] Answer: \answerNA{}
        \item[] Justification: No risk

\item {\bf Licenses for existing assets}
    \item[] Question: Are the creators or original owners of assets (e.g., code, data, models), used in the paper, properly credited and are the license and terms of use explicitly mentioned and properly respected?
    \item[] Answer: \answerYes{}
        \item[] Justification: cited and follow the license.

\item {\bf New assets}
    \item[] Question: Are new assets introduced in the paper well documented and is the documentation provided alongside the assets?
    \item[] Answer: \answerYes{}
        \item[] Justification: Provided

\item {\bf Crowdsourcing and research with human subjects}
    \item[] Question: For crowdsourcing experiments and research with human subjects, does the paper include the full text of instructions given to participants and screenshots, if applicable, as well as details about compensation (if any)? 
    \item[] Answer: \answerNA{}
        \item[] Justification: No crowdsourcing experiment.

\item {\bf Institutional review board (IRB) approvals or equivalent for research with human subjects}
    \item[] Question: Does the paper describe potential risks incurred by study participants, whether such risks were disclosed to the subjects, and whether Institutional Review Board (IRB) approvals (or an equivalent approval/review based on the requirements of your country or institution) were obtained?
    \item[] Answer: \answerNA{}
        \item[] Justification: No risk.

\item {\bf Declaration of LLM usage}
    \item[] Question: Does the paper describe the usage of LLMs if it is an important, original, or non-standard component of the core methods in this research? Note that if the LLM is used only for writing, editing, or formatting purposes and does \emph{not} impact the core methodology, scientific rigor, or originality of the research, declaration is not required.
    \item[] Answer: \answerNA{}
        \item[] Justification: LLM  \DIFaddbegin \DIFadd{assistance is }\DIFaddend used for  \DIFaddbegin \DIFadd{writing, editing, and formatting this revision, not as a component of the proposed core method}\DIFaddend .

\end{enumerate}

\end{document}